\documentclass[twocolumn,10pt]{asme2ej}

\usepackage{epsfig} 
\usepackage{lipsum}

\usepackage{epsfig}
\usepackage{amsmath}
\usepackage{amsfonts}
\usepackage{amssymb}
\usepackage{mathdots}
\usepackage{mathdots}
\usepackage[utf8]{inputenc}
\usepackage[titletoc,title]{appendix}
\usepackage{titletoc}

\usepackage{enumerate}
\usepackage{amsmath}
\usepackage{amssymb}
\usepackage{centernot}
\usepackage[mathscr]{euscript}
\usepackage{url}
\usepackage{hyphenat}
\usepackage{bm}
\usepackage{verbatim}
\usepackage{accents}
\usepackage{graphicx}
\usepackage{subcaption}
\usepackage{listings}
\usepackage{xcolor}
\usepackage{grffile}
\usepackage{rotating}
\usepackage{tikz}

\usepackage{grffile}
\usepackage{algorithm,algorithmicx}
\usepackage{algpseudocode}
\algnewcommand\algorithmicinput{\textbf{Input: }}
\algnewcommand\algorithmicoutput{\textbf{Output: }}
\usepackage{afterpage}
\usepackage{epstopdf}
\usepackage{widetext}
\usepackage{cuted}
\usepackage{multicol}

\everymath{\displaystyle}
\DeclareMathOperator*{\argmin}{argmin}
\DeclareMathOperator*{\argmax}{argmax}

\newcommand\black[1]{\textcolor{black}{#1}}

\usepackage{lineno,hyperref}
\modulolinenumbers[5]

\title{srMO-BO-3GP: A sequential regularized multi-objective Bayesian optimization for constrained design applications using an uncertain Pareto classifier}

\author{Anh Tran\thanks{Corresponding author: Anh Tran (anhtran@sandia.gov)} \ , Mike Eldred
    \affiliation{
    Optimization and Uncertainty Quantification \\ 
    Sandia National Laboratories \\
    Albuquerque, NM 87123 \\
    Email: \{anhtran, mseldre\}@sandia.gov\\ 
    }
}

\author{Scott McCann \\
    Silicon Technology \\
    Xilinx Inc. \\
    San Jose, CA 95124 \\
    Email: smccann@xilinx.com \\
}

\author{Yan Wang\\
    Woodruff School of Mechanical Engineering \\
    Georgia Institute of Technology \\
    Atlanta, GA 30332 \\
    Email: yan.wang@me.gatech.edu \\
}

\begin{document}

\maketitle  
\makeatletter
\let\ps@oldempty\ps@empty 
\renewcommand\ps@empty\ps@plain
\makeatother


\begin{abstract}
    
Bayesian optimization (BO) is an efficient and flexible global optimization framework that is applicable to a very wide range of engineering applications. 
To leverage the capability of the classical BO, many extensions, including multi-objective, multi-fidelity, parallelization, and latent-variable modeling, have been proposed to address the limitations of the classical BO framework. 
In this work, we propose a novel multi-objective BO formalism, called srMO-BO-3GP, to solve multi-objective optimization problems in a sequential setting. 
Three different Gaussian processes (GPs) are stacked together, where each of the GPs is assigned with a different task. The first GP is used to approximate a single-objective computed from the multi-objective definition, the second GP is used to learn the unknown constraints, and the third one is used to learn the uncertain Pareto frontier. 
At each iteration, a multi-objective augmented Tchebycheff function is adopted to convert multi-objective to single-objective, where the regularization with a regularized ridge term is also introduced to smooth the single-objective function. 
Finally, we couple the third GP along with the classical BO framework to explore the convergence and diversity of the Pareto frontier by the acquisition function for exploitation and exploration. 
The proposed framework is demonstrated using several numerical benchmark functions, as well as a thermomechanical finite element model for flip-chip package design optimization. 
\end{abstract}

\section{Introduction}

Global optimization is a common problem that appears in many contexts and particularly important for engineering design. 
The most used global optimization methods are population based searching algorithms, such as evolutionary algorithms. 
In contrast, surrogate based searching methods rely on the guidance of surrogate models to perform sequential sampling. Bayesian optimization (BO) method is such an example. BO is a derivative-free surrogate-based global optimization technique that has started being used in engineering domains. A surrogate of the objective function is constructed and continuously updated during the search. The sequential sampling is based on the maximization of an acquisition function, also constructed in the same searching space. Different acquisition functions such as expected improvement (EI), probability of improvement (PI), upper confidence bound (UCB), and entropy-based approaches have been used.  
The classical BO methods were developed for single-objective optimization, whereas multi-objective optimization problems are of interest in practice. 
In real-world applications, objectives are often found to conflict with each other. Design decision making requires us to enumerate options and make trade-offs between these objectives. 
Thus, it is important to obtain the set of optimal solutions in the Pareto sense so that the relationships among objectives can be fully explored.

Some efforts ~\cite{allmendinger2017surrogate,rojas2019survey} have been made to develop multi-objective Bayesian optimization (MOBO) methods. 
Jeong and Obayashi~\cite{jeong2005efficient} proposed an EI-based criteria and employed genetic algorithm (GA) to search for potential sampling points on the Pareto frontier. 
Knowles~\cite{knowles2006parego} introduced the ParEGO framework with the augmented Tchebycheff function to scalarize the multi-objective functions, which was later extended by Davins-Valldaura et al.~\cite{davins2017parego} by including the probability of Pareto frontier with another GP. 
D\"{a}chert et al. \cite{dachert2012augmented} studied the importance of the augmented weighted Tchebycheff regularization parameter between $\ell^{\infty}$ and $\ell^1$ norms. 
Zhang et al.~\cite{zhang2009expensive} proposed the MOEA/D framework with the classical Tchebycheff scalarization function and decomposed a multi-objective (MO) problem into a number of single-objective optimization subproblems. 
Li et al.~\cite{li2008kriging,li2009improving} improved the MO GA methods with the K-MOGA framework, where the posterior mean and posterior variance are considered to assist GA methods. 
Feliot et al.~\cite{feliot2017bayesian} proposed a multi-objective EI acquisition function considering constraints based on the posterior mean and posterior variance. 
Yang et al.~\cite{yang2019multi} proposed an expected hypervolume improvement gradient focusing on bi-objective problems and further accelerated the computation of hypervolume using an analytical derivation of gradients. Valladares and Tovar~\cite{valladares2020simple} compared MO EI (or MEI) criteria against Tchebycheff in $\ell^1$ and $\ell^{\infty}$ and concluded that randomized weights indeed help diversifying the Pareto frontier.

Gupta et al.~\cite{gupta2018exploiting} proposed a batch MOBO framework with a weighted linear average objective and demonstrated by optimizing the heat treatment process of an Al-Sc alloy. 
Shu et al.~\cite{shu2020new} developed a composite acquisition function for MOBO to efficiently sample with simultaneous improvement of convergence and diversity in constructing the Pareto frontier. 
Hern{\'a}ndez-Lobato et al.~\cite{hernandez2016predictive,garrido2019predictive} proposed a predictive entropy search acquisition function for MO problems considering constraints. 
Abdolshah et al.~\cite{abdolshah2019cost} proposed a MOBO framework where various computational efforts are taken into account when exploring the input domain. 
Gaudrie et al.~\cite{gaudrie2019targeting} proposed a sequential and batch extension for MOBO method by maximizing the expected hypervolume improvement. 
Palar et al.~\cite{palar2020impact} compared the impact of different covariance functions in MOBO and concluded that Mat\'{e}rn-3/2 is the most robust kernel for design optimization applications. 
Golovin et al~\cite{golovin2020random} proposed a hypervolume scalarization algorithm and proved the regret bound scales as $\mathcal{O}(\sqrt{T})$. 
Han and Zheng~\cite{han2020kriging} proposed an EIR2 indicator as a minimax extension for EI acquisition function in the context of MO optimization problems. 
\black{While hypervolume-based approaches (also known as $\mathcal{S}$-metric or Lebesgue measure) aim at both convergence and diversity, it also comes with a computational cost to compute the hypervolume. Let $s$ be the number of objectives and $n$ be the number of data points. The grid-based hypervolume algorithm scales at $\mathcal{O}(m \cdot n^m)$, whereas the Walking Fish Group algorithm has a complexity of $\mathcal{O}(s \cdot  2^n)$~\cite{zhao2018fast}. Beume et al.~\cite{beume2009complexity} proved the lower bound of $\mathcal{O}(n \log n)$ for any number of objectives $s>1$ and presented an algorithm that scales particularly so for $s=3$. }
We propose an improved acquisition function that can \black{avoid the cost of computing hypervolume for arbitrarily many objective functions}. 


In this paper,  
\black{we propose a new framework, called  srMO-BO-3GP,} 
to stack three GPs to solve the MO constrained optimization problems. 
Furthermore, we explore the possibility of introducing regularization in the Tchebycheff scalarization method. 
The approach is unique in three different perspectives. 
First, we employ GP as a machine learning technique to probabilistically search for the Pareto frontier, by stacking this classifier over the objective GP. 
The acquisition function in the classical BO method is employed to balance the exploitation and the exploration of this Pareto GP, which enhances its richness, diversity, and convergence in the Pareto frontier. 
Second, we combine the multi-objective functions to a single-objective function with a randomized weight vector and regularize this single-objective function by a ridge regularization to smoothen the single-objective function. 
The philosophy of our approach is intrinsically similar to that of classical BO methods. 
Compared to other approach in literature, the heuristic proposed in this research is greatly simplified, yet its efficiency is comparable to other methods. 
Our approach is more general in the sense that it does not restrict to a particular form of the acquisition function, and leaves the choice of acquisition functions up to the users. 
The idea is extended based on the work of Davins-Valldaura et al.~\cite{davins2017parego}; however, our work differs from their work in treating the Pareto GP, as well as the acquisition function. 
In our approach, we opt to include both exploitation and exploration of this Pareto GP by considering uncertainty, whereas in their approach, only the probability of Pareto frontier (i.e. exploitation) is considered. 
Another novelty in our approach is the ridge regularized term in the acquisition to smooth the acquisition function. 

Two significant advantages of the proposed srMO-BO-3GP algorithms are highlighted as follows. 
First, the regularization terms in the acquisition can significantly mitigate the effects of noise on observations. 
Second, our approach is not limited by the number of objective functions, as in other approaches. For example, if hypervolume-based approach is considered, the number of objectives would have a scalability effect on the computation of the hypervolume. 
Our proposed approach avoids limitation by converting the identification of Pareto frontier into an uncertain classification problems in machine learning context, with a flavor of uncertainty quantification. 
Thus, as the Pareto GP classifier gains more accuracy, it converges to the true Pareto frontier through the mean of classical Bayesian optimization approach.

For the remaining of this paper, Section~\ref{sec:Methodology} describes the proposed srMO-BO-3GP framework. 
Section~\ref{sec:NumericalResults} provides the numerical results for several numerical analytical functions, as well as an engineering thermomechanical finite element model (FEM) using the proposed approach. 
Section~\ref{sec:Discussion} provides discussions, and Section~\ref{sec:Conclusion} concludes the paper. 
Compared to our previous preliminary work in ~\cite{tran2020srmobo3gp}, in this paper, we provide a comprehensive benchmark for the proposed srMO-BO-3GP framework and update the weighted parameters balancing $\ell^\infty$ and $\ell^0$ norms in the augmented Tchebycheff scalarization function.

\section{Methodology}
\label{sec:Methodology}

We follow the formulation of Lin et al.~\cite{lin2018batched} in defining the MOBO problem. 
For the sake of clarity, we denote $\mathbf{x} = \{x_i\}_{i=1}^d \in \mathcal{X} \subseteq \mathbb{R}^d$ as the input of $d$ continuous variables in $d$-dimensional space, whereas $\mathbf{y} = \{y_j\}_{j=1}^s$ as $s$ outputs. 
Traditionally, BO solves the single-objective optimization problem
\begin{equation}
x^* = \argmin_{\mathbf{x} \in \mathcal{X}} f(\mathbf{x}),
\end{equation}
subject to the constraints $c(\mathbf{x}) \leq 0$. 
In this paper, we consider the scenario of MO optimization problem, 
\begin{equation}
\label{eq:ProblemStatement}
x^* = \argmin_{\mathbf{x} \in \mathcal{X}} (f_1(\mathbf{x}), \dots, f_s(\mathbf{x})),
\end{equation}
subject to the known constraints $\mathbf{g}(\mathbf{x}) \leq 0$. 

\subsection{Gaussian process}
\label{subsec:GP}


Assume that $f$ is a function of $\mathbf{x}$, where $\mathbf{x} \in \mathcal{X}$ is a $d$-dimensional input, and $y$ is the observation. 
Let the dataset $\mathcal{D} = (\mathbf{x}_i, y_i)_{i=1}^n$, where $n$ is the number of observations. 
A GP regression assumes that $\mathbf{f} = f_{1:n}$ is jointly Gaussian, and the observation $y$ is normally distributed given $f$, 
\begin{equation}
\mathbf{f} | \mathbf{X} \sim \mathcal{N}(\mathbf{m}, \mathbf{K}),
\end{equation} 
\begin{equation}
\mathbf{y} | \mathbf{f},\sigma^2 \sim \mathcal{N}(\mathbf{f}, \sigma^2 \mathbf{I}),
\end{equation} 
where $m_i:=\mu(\mathbf{x}_i)$ and $K_{i,j}:= k(\mathbf{x}_i, \mathbf{x}_j)$. 

The covariance kernel $\mathbf{K}$ is a choice of modeling covariance between inputs. At an unknown sampling location $\mathbf{x}$, the predicted response is described by a posterior Gaussian distribution, where the posterior mean is
\begin{equation}
\label{eq:mean}
\mu_n(\mathbf{x}) = \mu_0(\mathbf{x}) + \mathbf{k}(\mathbf{x})^T (\mathbf{K} + \sigma^2 \mathbf{I})^{-1} (\mathbf{y} - \mathbf{m}),
\end{equation}
and the posterior variance is
\begin{equation}
\label{eq:variance}
\sigma_n^2 = \mathbf{k}(\mathbf{x}) - \mathbf{k}(\mathbf{x})^T (\mathbf{K}  + \sigma^2 \mathbf{I})^{-1} \mathbf{k}(\mathbf{x}),
\end{equation}
where $k(\mathbf{x})$ is the covariance vector between the query point $\mathbf{x}$ and $\mathbf{x}_{1:n}$. 
The classical GP formulation assumes stationary covariance matrix, which only depends on the distance $r = \lVert \mathbf{x} - \mathbf{x'} \rVert$. 
Several most common kernels for GP include~\cite{shahriari2016taking}
\begin{itemize}
\item $k_{\text{Mat{\'e}rn}1} (\mathbf{x}, \mathbf{x'}) = \theta_0^2 \exp{(-r)}$, 
\item $k_{\text{Mat{\'e}rn}3} (\mathbf{x}, \mathbf{x'}) = \theta_0^2 \exp{(-\sqrt{3}r)} (1+\sqrt{3} r)$, 
\item $k_{\text{Mat{\'e}rn}5} (\mathbf{x}, \mathbf{x'}) = \theta_0^2 \exp{(-\sqrt{5}r)} \left( 1 + \sqrt{5}r + \frac{5}{3}r^2 \right)$, 
\item $k_{\text{sq-exp}} (\mathbf{x}, \mathbf{x'}) = \theta_0^2 \exp{\left(-\frac{1}{2}r^2 \right)}$.
\end{itemize}
The log-likelihood function can be written as
\begin{equation}
\begin{array}{lll}
\log{p(\mathbf{y} | \mathbf{x}_{1:n}, \theta )} &=& - \frac{n}{2} \log{(2\pi)} - \frac{1}{2} \log{| \mathbf{K}^{\theta} + \sigma^2 \mathbf{I} |}  \\
& & - \frac{1}{2} (\mathbf{y} - \mathbf{m}_{\theta})^T (\mathbf{K}^{\theta} + \sigma^2 \mathbf{I} )^{-1} (\mathbf{y} - \mathbf{m}_{\theta}).
\end{array}
\end{equation}
Optimizing the log-likelihood function yields the hyper-parameter $\theta$ at the computational cost of $\mathcal{O}(n^3)$ due to the cost to compute the inverse of the covariance matrix.

\subsection{Multi-objective function}

We adopt the definition of Pareto-dominant from Rojas-Gonzalez et al.~\cite{rojas2019survey} to define the Pareto frontier. 

\begin{enumerate}[\text{Definition}-1: ]

\item $\mathbf{x}_1$ is said to dominate $\mathbf{x}_2$, denoted as $\mathbf{x}_1 \preceq \mathbf{x}_2$, if and only if $\forall j: 1\leq j \leq s$, such that $y_j(\mathbf{x}_1) \leq y_j(\mathbf{x}_2)$, and $\exists j: 1 \leq j \leq s$, such that $y_j(\mathbf{x}_1) < y_j(\mathbf{x}_2)$. 

\item $\mathbf{x}_1$ is said to strictly dominate $\mathbf{x}_2$, denoted as $\mathbf{x}_1 \prec \mathbf{x}_2$, if and only if $\forall j: 1\leq j \leq s$, such that $y_j(\mathbf{x}_1) < y_j(\mathbf{x}_2)$.
\end{enumerate}

A MO problem can be solved by converting it to a single-objective problem, where the single-objective function is a weighted sum of multiple objectives~\cite{rojas2019survey}, such as,
\begin{enumerate}
\item weighted Tchebycheff scalarization function $y = \max_{1 \leq i \leq s} w_i ( y_i(\mathbf{x}) - z_i^* )$, 
\item the weighted sum scalarization function: $y = \sum_{i=1}^s w_i y_i(\mathbf{x})$, 
\item and the augmented (weighted) Tchebycheff scalarization function $y = \max_{1 \leq i \leq s} w_i ( y_i(\mathbf{x}) - z_i^* ) + \rho \sum_{i=1}^s w_i y_i(\mathbf{x}) $, 
\end{enumerate}
where $z_i^*$ denotes the ideal value for the $i$-th objective, the weights $0 \leq w_i \leq 1$, $\sum_{i=1}^m w_i =1$, $\rho$ is a positive constant
Typically, $\rho$ is set as 0.05 in~\cite{knowles2006parego,davins2017parego}. 
However, we follow \cite{ishibuchi2010use} and set $\rho = 0.65$ due to its superior performance; the idea of tuning $\rho$ adaptively remains open for future research.

Following the idea of Davins-Valldaura et al.~\cite{davins2017parego} in converting multiple objectives to a single-objective function, 
we propose a new objective function based on the augmented Tchebycheff scalarization function, with the addition of a regularized term for the single-objective function, which could be interpreted in terms of ridge regression,
\begin{equation}
\label{eq:ridgeTchebycheff}
\begin{array}{lll}
y &=& \max_{1 \leq j \leq s} w_j y_j(\mathbf{x})  + \rho \sum_{j=1}^s w_j y_j(\mathbf{x}) + \lambda \lVert \mathbf{x} \rVert_2, \\
\end{array}
\end{equation}
where $\lambda$ is a regularization parameter. 
The first term can be thought of as an $\ell^\infty$-norm, whereas the second term can be thought of as a $\ell^1$-norm \cite{dachert2012augmented}. 
Various treatments for the Tchebycheff decomposition exist in literature \cite{trivedi2016survey}. 
The regularization term is used to smoothen the objective function and mitigate the singularity effect when Tchebycheff decomposition is introduced. 
More specifically, the term $ \max_{1 \leq i \leq s} w_j  y_j(\mathbf{x})  $ does not have a continuous first derivative, while the assumption for fitting GP is that the underlying function needs to be smooth. 
In that sense, it is necessary to introduce a small ridge regularization term in $\ell^2$-norm.

\subsection{Constraints}
\label{subsec:Constraints}

We consider the known constraints and hidden constraints, where the known constraints are known before the functional evaluation, whereas the hidden constraints must be learn indirectly through the functional evaluation. 
\black{For real-world engineering design applications, some constraints are imposed to the design performance which does not have a simple analytical form to evaluate. Simulations such as finite element model and computational fluid dynamics need to be run to check if the constraints are satisfied. Those are called unknown constraints, which typically correspond to unforeseeable and reproducible failure of the applications with low performance for a certain region of inputs. It could also be due to numerical issues in simulations, e.g. mesh failure, numerical instability, and numerical divergence. Such errors must be accounted for, but they are only known after the functional evaluation of $f(\cdot)$ (in Equation \ref{eq:ProblemStatement}) is invoked with simulations. We refer to the constraints that are related to the functional evaluation $f(\cdot)$ are unknown constraints. On the contrary, the known constraints $c(\mathbf{x})$ are those that can be evaluated or invoked separately based on the available analytical forms without calling the functional evaluation $f(\cdot)$ by simulation.}

\subsubsection{Known constraints}

Known constraints can be represented as a set of inequalities $\mathbf{g}(\mathbf{x}) \leq \mathbf{0}$, where $\mathbf{g}$ is a relatively cheap set of functions to evaluate, compared to the real objective function $f$. 
The known constraints can be easily implemented by directly penalizing the acquisition by setting it to zero if known constraints are violated, while leaving unviolated sampling points as is.

In practice, the penalty scheme is implemented by multiplying the acquisition with another indicator function $\mathcal{I}(\mathbf{x})$, 
\begin{equation}
\label{eq:indicatorFunction}
\mathcal{I}(\mathbf{x}) = 
\begin{cases}
1, \quad \text{ if } \forall k: g_k(\mathbf{x}) \leq 0,  \\ 
0, \quad \text{ if } \exists k: g_k(\mathbf{x}) > 0,
\end{cases}
\end{equation}
where $g_k$ denotes the $k$-th constraint in the set of known constraints.

\subsubsection{Unknown/hidden constraints}

We adopt our previous strategy by employing a probabilistic binary classifier to learn the unknown or hidden constraints. 
As a result, feasible and infeasible regions are separated in the input domain $\mathcal{X}$. 
These two regions are mutually exclusive, i.e. disjoint, because a sampling point cannot be both feasible and infeasible at the same time. 
The labels of feasible/infeasible for sampling points are fixed, in the sense that as the optimization process advances, the labels do not change.

Denote the feasibility dataset as $\{\mathbf{x}_i, c_i\}_{i=1}^n$, where $n$ is the number of data points and $c$ is associated with unknown constraints. 
We assign $c_i = 1$ if $\mathbf{x}_i$ is feasible, and $c_i=0$ if $\mathbf{x}_i$ is infeasible. At an unknown $\mathbf{x}$, the feasibility classifier provides a probability mass function, with $\text{Pr}(\mathbf{x} | c(\mathbf{x}) = 1)$ as the predicted probability of satisfying the hidden constraints, and $\text{Pr}(\mathbf{x} | c(\mathbf{x}) = 0)$ as the predicted probability of failing the hidden constraints. 
It is noteworthy that their sum adds up $\text{Pr}(\mathbf{x} | c(\mathbf{x}) = 0) + \text{Pr}(\mathbf{x} | c(\mathbf{x}) = 1)$, as they are mutually exclusive and there are only two possibilities. 
The probability of passing the unknown constraints will be used to condition on the acquisition, resulting in the multiplication of $\text{Pr}(\mathbf{x} | c(\mathbf{x}) = 1)$ in the conditioned acquisition. 

Even though there are many available probabilistic binary classifier in the context of machine learning, for example, $k$-NN~\cite{bentley1975multidimensional}, AdaBoost~\cite{hastie2009multi}, RandomForest~\cite{breiman2001random}, support vector machine~\cite{hearst1998support} (SVM), least squares support vector machine (LSSVM)~\cite{suykens1999least}, and convolutional neural network~\cite{lecun2015deep}, in this work, we restrict our methodology to GP as a binary probabilistic classifier. 
Labeling feasibility as described above, the posterior mean of feasibility GP can be used to predict the probability of satisfying unknown constraints, i.e. $\mu_{\text{feasible}}(\mathbf{x}) = \text{Pr}(\mathbf{x} | c(\mathbf{x}) = 1)$.

\subsection{Pareto frontier with an uncertain GP classifier}
\label{subsec:ParetoClassif}

At each iteration, we construct the current Pareto frontier, which is subjected to change as the optimization process advances. 
If a sampling point is currently Pareto-dominant, the point is labeled as 1, and if the sampling point is not Pareto-dominant (i.e. it is dominated by another point in the dataset), the sampling point is labeled as 0. 
The classification process is thus uncertain, in the sense that the labels change from one iteration to another. 
This is to contrast with the constraint feasibility classifier $\mu_{\text{feasible}}(\mathbf{x})$, where the labels are fixed and do not change, as the optimization process advances. 
The uncertainty in the Pareto frontier classification gradually decreases, as the number of sampling data points increases.

We explore the possibility of using GP as an uncertain Pareto frontier classifier. 
Surprisingly, GP is one of a few well-established machine learning techniques that considers uncertainty in prediction through its posterior variance function. 
Because of this particular reason, GP is employed as an uncertain classifier to construct the Pareto frontier, where the uncertainty is quantified by the GP posterior variance function (Eq~\ref{eq:variance}). 

The main idea is to force the Pareto GP classifier to balance its learning by exploitation and exploration, especially when the Pareto frontier prediction is not accurate by focusing on the unknown region at the beginning of the optimization process. 
As the optimization advances, if the Pareto frontier can be classified with high accuracy, the BO framework should be exploited to promote convergence and explored to promote the diversity in the MO optimization settings. 
This is consistent with the philosophy of the classical BO approach, which strikes for the balance of exploration and exploitation. 
Furthermore, the convergence and diversity, which are the two keys measure of MO optimization problems~\cite{rojas2019survey}, are promoted by employing common acquisition functions on the Pareto frontier GP. 
As a result, the acquisition of Pareto GP classifier is included as another in the main acquisition function for the classical BO method (Equation~\ref{eq:FullAcquisition}). 

It is noteworthy to point out that the Pareto classification problem is also mutually exclusive, in the sense that a sampling point cannot be both Pareto-dominant and Pareto-non-dominant at the same time. 
For an unknown location $\mathbf{x}$, the Pareto GP classifier provides both the probability of being Pareto-dominant $\mu_{\text{Pareto}}(\mathbf{x})$, which is bounded between 0 and 1, as well as the uncertainty associated with the probability as $\sigma^2_{\text{Pareto}}(\mathbf{x})$. 
\black{Instead of employing logistic logit function which maps from $(-\infty,+\infty)$ to $[0,1]$ for binary classification problem~\cite{rasmussen2004gaussian}, we simply use the regression formulation for GP and clip the prediction if the responses are beyond the range of $[0,1]$.}

\subsection{Acquisition function}
\label{subsec:acqFunc}

We propose an extended criteria in learning the Pareto frontier considering uncertainty, as opposed to directly couple the Pareto-dominant probability into the EI acquisition function in Davins-Valldaura et al~\cite{davins2017parego}. 
The composite acquisition function is defined as
\begin{equation}
a(\mathbf{x}) = \underbrace{a_{\text{obj}}(\mathbf{x})}_\text{objective GP\ } 
\cdot \underbrace{a_{\text{Pareto}}(\mathbf{x})}_\text{uncertain Pareto\ } 
\cdot \underbrace{\text{Pr}(\mathbf{x} | c(\mathbf{x}) = 1)}_\text{unknown cstr.\ } 
\cdot \underbrace{\mathcal{I}(\mathbf{x})}_\text{known cstr.\ }.
\label{eq:FullAcquisition}
\end{equation}
Equation~\ref{eq:FullAcquisition} is explained as follows. 
The first term, $a_{\text{obj}}(\mathbf{x};\  \mu_{\text{obj}}(\mathbf{x})$, $\sigma^2_{\text{obj}}(\mathbf{x}))$, is the regularized augmented Tchebycheff single-objective function in Equation~\ref{eq:ridgeTchebycheff}. 
At each time step, a random weight vector $\mathbf{w} = (w_1, \dots, w_s)$ is sampled to combine multiple objectives $\{y_j\}_{j=1}^s$ to a single-objective function $y$. 
A GP model is fitted using the dataset of $n$ data points $\{\mathbf{x}_i, y_i\}_{i=1}^n$. 
An acquisition function $a_{\text{obj}}(\mathbf{x})$ is formed, as a function of posterior mean $\mu_{\text{obj}}(\mathbf{x})$ and posterior variance $\sigma^2_{\text{obj}}(\mathbf{x})$. 
The second term, $a_{\text{Pareto}}(\mathbf{x};\  \mu_{\text{Pareto}}(\mathbf{x})$, $\sigma^2_{\text{Pareto}}(\mathbf{x}))$, is the acquisition function based on the Pareto frontier GP classifier. 
The third term, $\text{Pr}(\mathbf{x} | c(\mathbf{x}) = 1) =  \mu_{\text{feasible}}(\mathbf{x})$, is the probability of passing unknown constraints, provided by posterior mean of the second GP. 
The fourth term, $\mathcal{I}(\mathbf{x})$, is the indicator function \black{describing known constraints as} in Equation~\ref{eq:indicatorFunction}. The next sampling point $\mathbf{x}^*$ is obtained by maximizing the acquisition function described in Equation~\ref{eq:FullAcquisition}, i.e.
\begin{equation}
\label{eq:NextSamplingPoint}
\mathbf{x}^* = \argmax_{\mathbf{x} \in \mathcal{X}} a(\mathbf{x})
\end{equation}

\subsection{Summary}

The srMO-BO-3GP framework is summarized in Algorithm~\ref{alg:srmo-bo-3gp}. 
CMA-ES~\cite{hansen2003reducing,hansen2004evaluating} is used as the sub-optimizer to obtain the next sampling point in Equation~\ref{eq:NextSamplingPoint}, where the default settings are retained. 
An interface between the srMO-BO-3GP optimizer and the engineering applications is constructed by MATLAB, Python, and Shell scripts. 
Also, since the maximization is set as a default setting, the Tchebycheff function is algebraically manipulated to conform with the default setting.



\begin{algorithm*}
\caption{srMO-BO-3GP algorithm.}
\label{alg:srmo-bo-3gp}
\algorithmicinput dataset $\mathcal{D}_n$ consisting of inputs, observation, feasibility $(\mathbf{x}, \mathbf{y}, c)_{i=1}^n$, multi-objective $(\mathbf{x}_i, \mathbf{y}_i)_{i=1}^n$, constraint GP $(\mathbf{x}, c_i)_{i=1}^n$, 
\begin{algorithmic}[1]

\For{$n=1,2,\dots,$}
\State randomize a weight vector $\mathbf{w}$
\State combine $\{y_j\}_{j=1}^s$ to $y$ \Comment{multi- to single-objective - Eq. ~\ref{eq:ridgeTchebycheff})}
\State construct single-objective GP \Comment{GP \#1: $\mu_{\text{obj}}(\mathbf{x})$, $\sigma^2_{\text{obj}}(\mathbf{x})$}
\State construct Pareto front \Comment{GP \#2: $\mu_{\text{Pareto}}(\mathbf{x})$, $\sigma^2_{\text{Pareto}}(\mathbf{x})$}
\State \quad find current Pareto front
\State \quad construct Pareto classifier GP
\State construct constraints classifier GP \Comment{GP \#3: $\mu_{\text{feasible}}(\mathbf{x}), \sigma^2_{\text{feasible}}(\mathbf{x})$}
\State locate the next sampling point $\mathbf{x}_{n+1}$ \Comment{(Eq.~\ref{eq:FullAcquisition})}
\State query objectives $\mathbf{y}_{n+1} \gets \{y_j\}_{j=1}^s$ and feasibility $c_{n+1}$ 
\State augment dataset $\mathcal{D}_{n+1} \gets \{ \mathcal{D}_n \cup (\mathbf{x}_{n+1}, \mathbf{y}_{n+1}, c_{n+1}) \}$
\EndFor
\end{algorithmic}
\end{algorithm*}

\section{Numerical benchmarks}
\label{sec:NumericalResults}

In this section, we follow the standard test suite~\cite{davins2017parego,huband2006review} to benchmark and investigate the effectiveness and the efficiency of the proposed srMO-BO-3GP algorithm by ZDT~\cite{zitzler2000comparison} and DTLZ~\cite{deb2005scalable} test suites, where the geometry of Pareto frontiers are well visualized by Jin et al.~\cite{jin2019multi} as well as Ishibuchi et al.~\cite{ishibuchi2015pareto}; analytical solutions of Pareto frontiers are provided by the work of Deb et al~\cite{deb2005scalable}. 
The regularization $\lambda$ parameter is set to 0.01, even though its effect requires a further benchmark study. 
We compare the numerical performance between variants of srMO-BO-3GP, particularly for
\begin{itemize}
\item - variations of acquisition function for the objective GP: PI, EI, and UCB;
\item - variations of acquisition function for the Pareto GP: PI, EI, and UCB;
\item - regularized vs. non-regularized Tchebycheff objective function.
\end{itemize}
This results in 18 different variants of the srMO-BO-3GP. 
The name convention for these variants is Reg/NoReg-\{PI,EI,UCB\}-\{PI,EI,UCB\}, which corresponds the option of regularization, the objective GP, and the Pareto GP. 
Note that in the DTLZ benchmark, $d = M+k-1$, and the function $g(\mathbf{x})$ is constructed based on $k = d - M +1 $ variables, $(x_M, \dots, x_d)$.
The benchmark is repeated 5 times for each variant BO method and each benchmark function, resulting in almost 1,000 runs in total. 
\black{In all the optimization runs, 5 initial sampling points are randomly chosen using Monte Carlo sampling. The maximum number of iteration is 1500. For ZDT and DTLZ benchmarks, the dimensions are 3 and 4, respectively. For ZDT benchmark, the number of objectives is 2, whereas for DTLZ benchmark, the number of objectives is 3.}


\subsection{ZDT test suite}
\subsubsection{ZDT1}

For a $d$-dimensional input $\mathbf{x} \in [0,1]^d$, a more general ZDT1 function can be introduced \cite{zitzler2000comparison}
\begin{equation}
f_1(\mathbf{x}) = x_1, \quad f_2(\mathbf{x}) = g h,
\end{equation}
where
\begin{equation}
g = 1 + 9 \sum_{i=2}^{d} \frac{x_i}{d-1}, \quad h = 1 - \sqrt{ \frac{f_1}{g} }.
\end{equation}

\subsubsection{ZDT2}

For a $d$-dimensional input $\mathbf{x} \in [0,1]^d$, the ZDT2 function can be described as \cite{zitzler2000comparison}
\begin{equation}
f_1(\mathbf{x}) = x_1, \quad f_2(\mathbf{x}) = g h,
\end{equation}
where
\begin{equation}
g = 1 + 9 \sum_{i=2}^{d} \frac{x_i}{d-1}, \quad h = 1 - \left( \frac{f_1}{g} \right)^2.
\end{equation}

\subsubsection{ZDT3}

For a $d$-dimensional input $\mathbf{x} \in [0,1]^d$, the ZDT3 function can be described as \cite{zitzler2000comparison}
\begin{equation}
f_1(\mathbf{x}) = x_1, \quad f_2(\mathbf{x}) = g h,
\end{equation}
where
\begin{equation}
g = 1 + 9 \sum_{i=2}^{d} \frac{x_i}{d-1}, \quad h = 1 - \sqrt{\frac{f_1}{g}} - \left( \frac{f_1}{g} \sin(10\pi f_1) \right).
\end{equation}

\subsubsection{ZDT4}

For a $d$-dimensional input $\mathbf{x} \in [0,1]^1 \times [-5,5]^{d-1}$, the ZDT4 function can be described as \cite{zitzler2000comparison}
\begin{equation}
f_1(\mathbf{x}) = x_1, \quad f_2(\mathbf{x}) = g h,
\end{equation}
where
\begin{equation}
g = 1 + 10 (d - 1) + \sum_{i=2}^{d} [x_i^2 - 10\cos(4\pi x_i)], \quad h = 1 - \sqrt{ \frac{f_1}{g} }.
\end{equation}
The Pareto frontier are with $g(\mathbf{x}) = 1$.

\subsubsection{ZDT6}

For a $d$-dimensional input $\mathbf{x} \in [0,1]^d$, the ZDT6 function can be described as \cite{zitzler2000comparison}
\begin{equation}
f_1(\mathbf{x}) = 1 - \exp(-4 x_1) \sin^6(6\pi x_1), \quad f_2(\mathbf{x}) = g h,
\end{equation}
where
\begin{equation}
g = 1 + 9 \left( \sum_{i=2}^{d} \frac{x_i}{d-1} \right)^{0.25}, \quad h = 1 - \left( \frac{f_1}{g} \right)^2.
\end{equation}

\subsection{DTLZ test suite}

\subsubsection{DTLZ1}

A generalized DTLZ1 function for a $d$-dimensional input $\mathbf{x} \in [0,1]^d$ with $M$ objectives, $f_{1:M}$, is defined as \cite{deb2005scalable}
\footnotesize
\begin{eqnarray}
f_1 (\mathbf{x}) = 0.5 (1 + g) \prod_{i=1}^{M-1} x_i \\
f_{k=2:M-1} (\mathbf{x}) = 0.5 (1 + g) \left[ \prod_{i=1}^{M-k} x_i \right] (1 - x_{M - k + 1}), \\
f_M(\mathbf{x}) = 0.5 (1 + g) (1 - x_1),
\end{eqnarray}
\normalsize
where $g = 100 \left( d + \sum_{i=1}^d \left\{ (x_i - 0.5)^2 - \cos\left[20\pi (x_i - 0.5) \right] \right\} \right)$.

\subsubsection{DTLZ2}

A generalized DTLZ2 function for a $d$-dimensional input $\mathbf{x} \in [0,1]^d$ with $M$ objectives, $f_{1:M}$, is defined as \cite{deb2005scalable}
\footnotesize
\begin{eqnarray}
f_1 (\mathbf{x}) = 0.5 (1 + g) \prod_{i=1}^{M-1} \cos\left( \frac{x_i\pi}{2} \right) \\
f_{k=2:M-1} (\mathbf{x}) = 0.5 (1 + g) \left[ \prod_{i=1}^{M-k} \cos\left( \frac{x_i\pi}{2} \right) \right] \sin\left( \frac{x_{M - k + 1}\pi}{2} \right), \\
f_M(\mathbf{x}) = 0.5 (1 + g) \sin\left( \frac{x_1 \pi}{2} \right),
\end{eqnarray}
\normalsize
where $g = \sum_{i=M}^d \left( x_i - 0.5 \right)^2$.

\subsubsection{DTLZ3}

A generalized DTLZ2 function for a $d$-dimensional input $\mathbf{x} \in [0,1]^d$ with $M$ objectives, $f_{1:M}$, is defined as \cite{deb2005scalable}
\footnotesize
\begin{eqnarray}
f_1 (\mathbf{x}) = 0.5 (1 + g) \prod_{i=1}^{M-1} \cos\left( \frac{x_i\pi}{2} \right) \\
f_{k=2:M-1} (\mathbf{x}) = 0.5 (1 + g) \left[ \prod_{i=1}^{M-k} \cos\left( \frac{x_i\pi}{2} \right) \right] \sin\left( \frac{x_{M - k + 1}\pi}{2} \right), \\
f_M(\mathbf{x}) = 0.5 (1 + g) \sin\left( \frac{x_1 \pi}{2} \right),
\end{eqnarray}
\normalsize
where $g = 100 \left( k + \sum_{i=M}^d \left\{ (x_i - 0.5)^2 - \cos\left[20\pi (x_i - 0.5) \right] \right\} \right)$.

\subsubsection{DTLZ4}

A generalized DTLZ4 function for a $d$-dimensional input $\mathbf{x} \in [0,1]^d$ with $M$ objectives, $f_{1:M}$, is defined as \cite{deb2005scalable}
\footnotesize
\begin{eqnarray}
f_1 (\mathbf{x}) = 0.5 (1 + g) \prod_{i=1}^{M-1} \cos\left( \frac{x_i^{\alpha}\pi}{2} \right) \\
f_{k=2:M-1} (\mathbf{x}) = 0.5 (1 + g) \left[ \prod_{i=1}^{M-k} \cos\left( \frac{x_i^{\alpha}\pi}{2} \right) \right] \sin\left( \frac{x_{M - k + 1}\pi}{2} \right), \\
f_M(\mathbf{x}) = 0.5 (1 + g) \sin\left( \frac{x_1 \pi}{2} \right),
\end{eqnarray}
\normalsize
where $g = \sum_{i=M}^d \left( x_i - 0.5 \right)^2$.

\subsubsection{DTLZ5}

A generalized DTLZ5 function for a $d$-dimensional input $\mathbf{x} \in [0,1]^d$ with $M$ objectives, $f_{1:M}$, is defined as \cite{deb2005scalable}
\footnotesize
\begin{eqnarray}
f_1 (\mathbf{x}) = 0.5 (1 + g) \prod_{i=1}^{M-1} \cos\left( \frac{\theta_i\pi}{2} \right) \\
f_{k=2:M-1} (\mathbf{x}) = 0.5 (1 + g) \left[ \prod_{i=1}^{M-k} \cos\left( \frac{\theta_i\pi}{2} \right) \right] \sin\left( \frac{x_{M - k + 1}\pi}{2} \right), \\
f_M(\mathbf{x}) = 0.5 (1 + g) \sin\left( \frac{x_1 \pi}{2} \right),
\end{eqnarray}
\normalsize
where 
$\theta_i = \frac{\pi}{4 (1 + g)} (1 + 2 g x_i)$, for $2 \leq i \leq M$, 
$g = \sum_{i=M}^d \left( x_i - 0.5 \right)^2$.

\subsubsection{DTLZ6}

A generalized DTLZ6 function for a $d$-dimensional input $\mathbf{x} \in [0,1]^d$ with $M$ objectives, $f_{1:M}$, is defined as \cite{deb2005scalable}
\footnotesize
\begin{eqnarray}
f_1 (\mathbf{x}) = 0.5 (1 + g) \prod_{i=1}^{M-1} \cos\left( \frac{\theta_i\pi}{2} \right) \\
f_{k=2:M-1} (\mathbf{x}) = 0.5 (1 + g) \left[ \prod_{i=1}^{M-k} \cos\left( \frac{\theta_i\pi}{2} \right) \right] \sin\left( \frac{x_{M - k + 1}\pi}{2} \right), \\
f_M(\mathbf{x}) = 0.5 (1 + g) \sin\left( \frac{x_1 \pi}{2} \right),
\end{eqnarray}
\normalsize
where 
$\theta_i = \frac{\pi}{4 (1 + g)} (1 + 2 g x_i$ for $2 \leq i \leq M$, 
$g = \sum_{i=M}^d x_i^{0.1} $.

\subsection{Performance metrics}

We followed and adopted the notation from Jin et al.~\cite{jin2019multi} and Hern{\'a}ndez-Lobato et al.~\cite{hernandez2016predictive}, to measure numerical performance for our proposed method, namely the generational distance (GD), the inverted generational distance (IGD), and the hypervolume (HV). 
These metrics also require prior knowledge of the true Pareto frontier. 
For the sake of clarity, we denote $\text{PF}^*$ as a set of the discretized points on the true Pareto frontier of an arbitrary benchmark function, and $\text{PF}$ as the set of obtained Pareto frontier.

\subsubsection{Generational distance (GD)}

The generational distance between PF$^*$ and PF, denoted as GD, is computed as 
\begin{equation}
\text{GD}(\text{PF}, \text{PF}^*) = \frac{\sqrt{\sum_{u\in\text{PF}} d(u, \text{PF}^*) }}{|\# \text{PF}|},
\end{equation}
where $d$ is the minimum Euclidean distance between an obtained Pareto-dominant point and the true Pareto frontier. In other word, $d(u, \text{PF}^*) = \argmin_{v \in \text{PF}^*} d(u,v)$. 
The denominator, $|\# \text{PF}|$, is the number of obtained solution in PF. 
A smaller value of GD indicates a better convergence to the true Pareto frontier.

\subsubsection{Inverted generational distance (IGD)}

\begin{equation}
\text{IGD}(\text{PF}, \text{PF}^*) = \frac{\sqrt{\sum_{v\in\text{PF}^*} d(\text{PF}, v) }}{|\# \text{PF}^*|},
\end{equation}
where $d(\text{PF}, v) = \argmin_{v\in\text{PF}^*} d(u,v)$ is the minimum Euclidean distance between $v$ on the true Pareto frontier with the obtained Pareto frontier. 
The denominator, $|\# \text{PF}^*|$, is the number of discretization Pareto-dominant point. 
As pointed out by Jin et al.~\cite{jin2019multi}, if $|\#\text{PF}^*|$ is sufficiently large, i.e. when the true Pareto frontier is sufficiently discretized, then IGD can measure both convergence and diversity. 
A smaller of IGD indicates a better convergence and diversity to the true Pareto frontier.

\subsubsection{Log of relative hypervolume difference (LRHD)}

The hypervolume metric, which is a strictly monotonic measure, also known as Lesbegue measure or $\mathcal{S}$-metric, is computed using Walking Fish Group algorithms used~\cite{cox2016improving,while2005new}. 
We adopt the log of relative hypervolume difference from Hern{\'a}ndez-Lobato et al.~\cite{hernandez2016predictive} to measure the performance. 
The log of relative hypervolume difference is computed as
\begin{equation}
\text{LRHD} = \log{( | \text{HV} - \text{HV}_{\text{ideal}} | )},
\end{equation}
where HV$_\text{ideal}$ is the ideal hypervolume and $\text{HV} > \text{HV}_\text{ideal}$ is the computed hypervolume of the obtained Pareto frontier. 
The origin is chosen as the reference point to compute the hypervolume. 
A smaller of LRHD indicates a better convergence and diversity to the true Pareto frontier. 

\subsection{Comparison between variants}

\begin{figure*}[!t]

        \begin{subfigure}[b]{0.33\textwidth}
        \centering
        \includegraphics[width=1.0\textwidth, keepaspectratio]{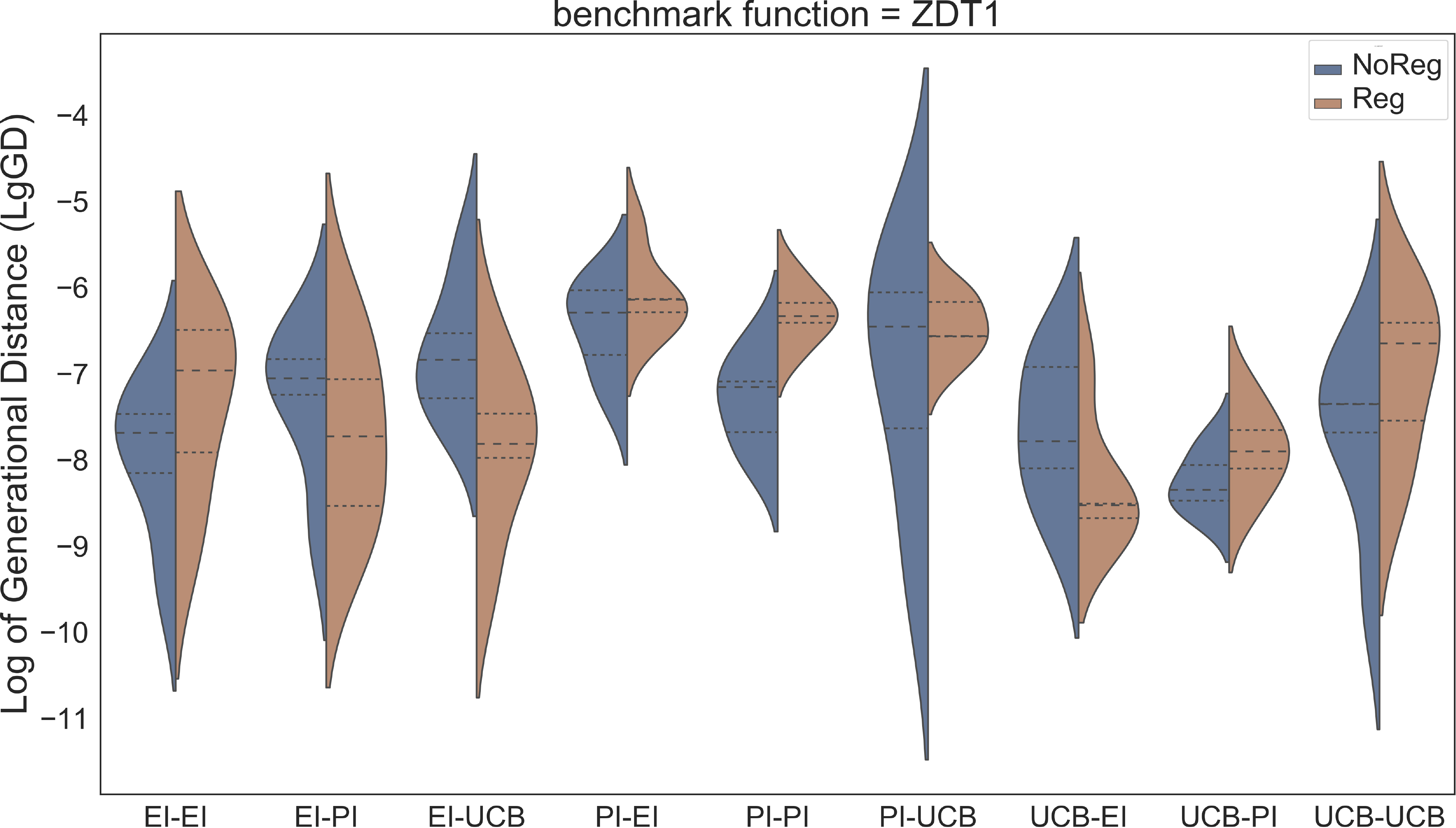}
        \end{subfigure}
        \hfill
        \begin{subfigure}[b]{0.33\textwidth}
        \centering
        \includegraphics[width=1.0\textwidth, keepaspectratio]{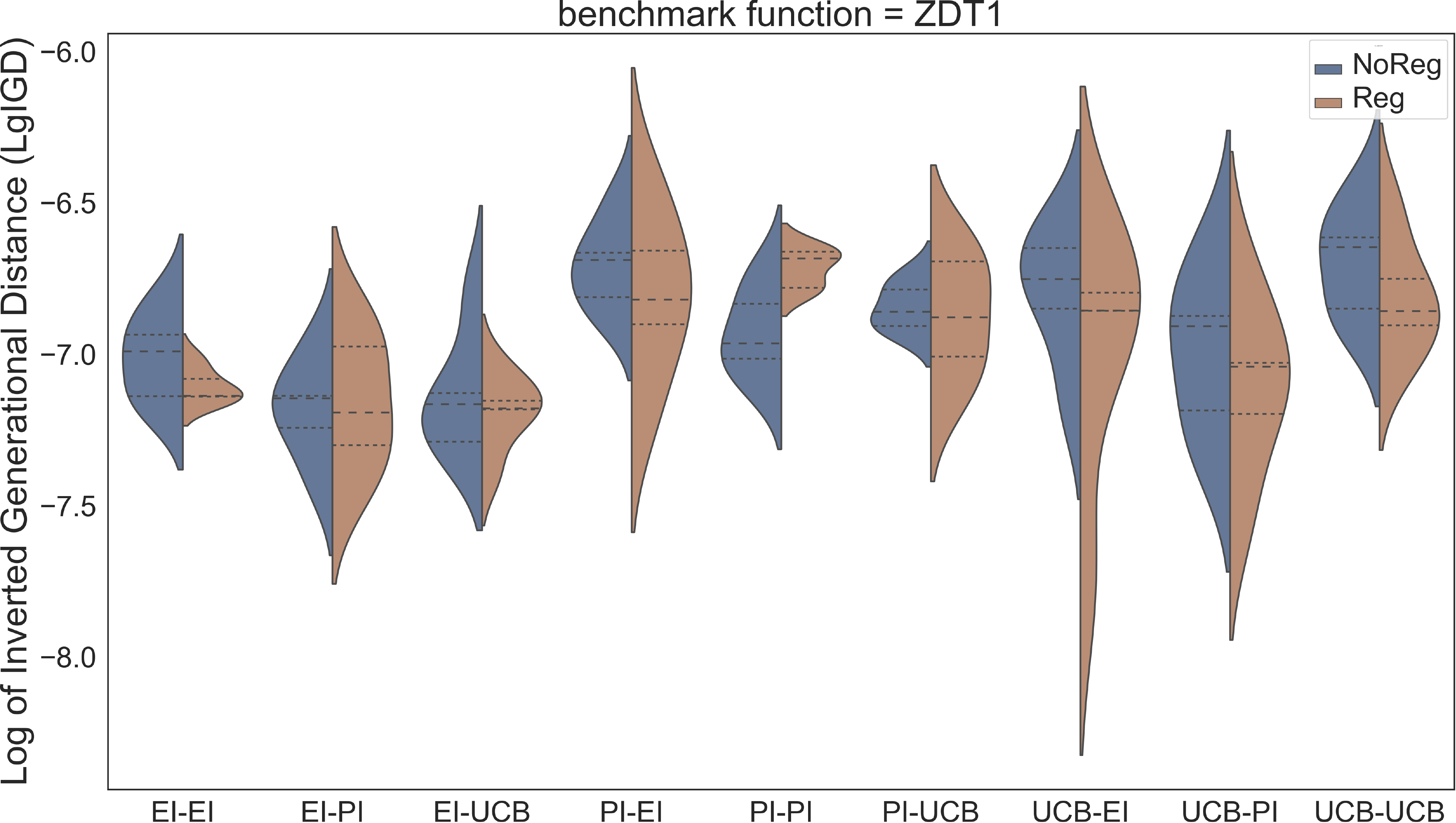}
        \end{subfigure}
        \hfill
        \begin{subfigure}[b]{0.33\textwidth}
        \centering
        \includegraphics[width=1.0\textwidth, keepaspectratio]{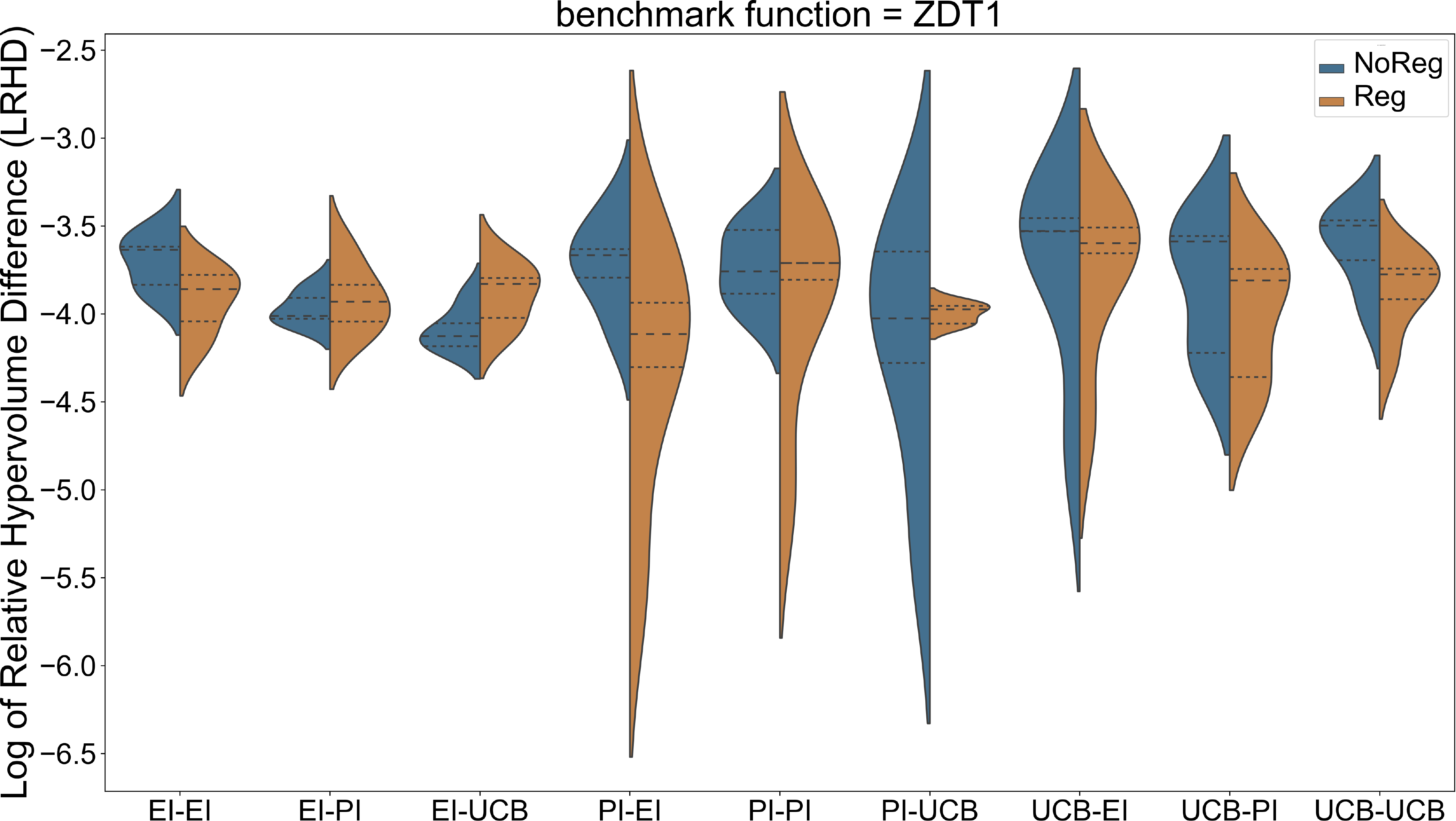}
        \end{subfigure}

        \vfill

        \begin{subfigure}[b]{0.33\textwidth}
        \centering
        \includegraphics[width=1.0\textwidth, keepaspectratio]{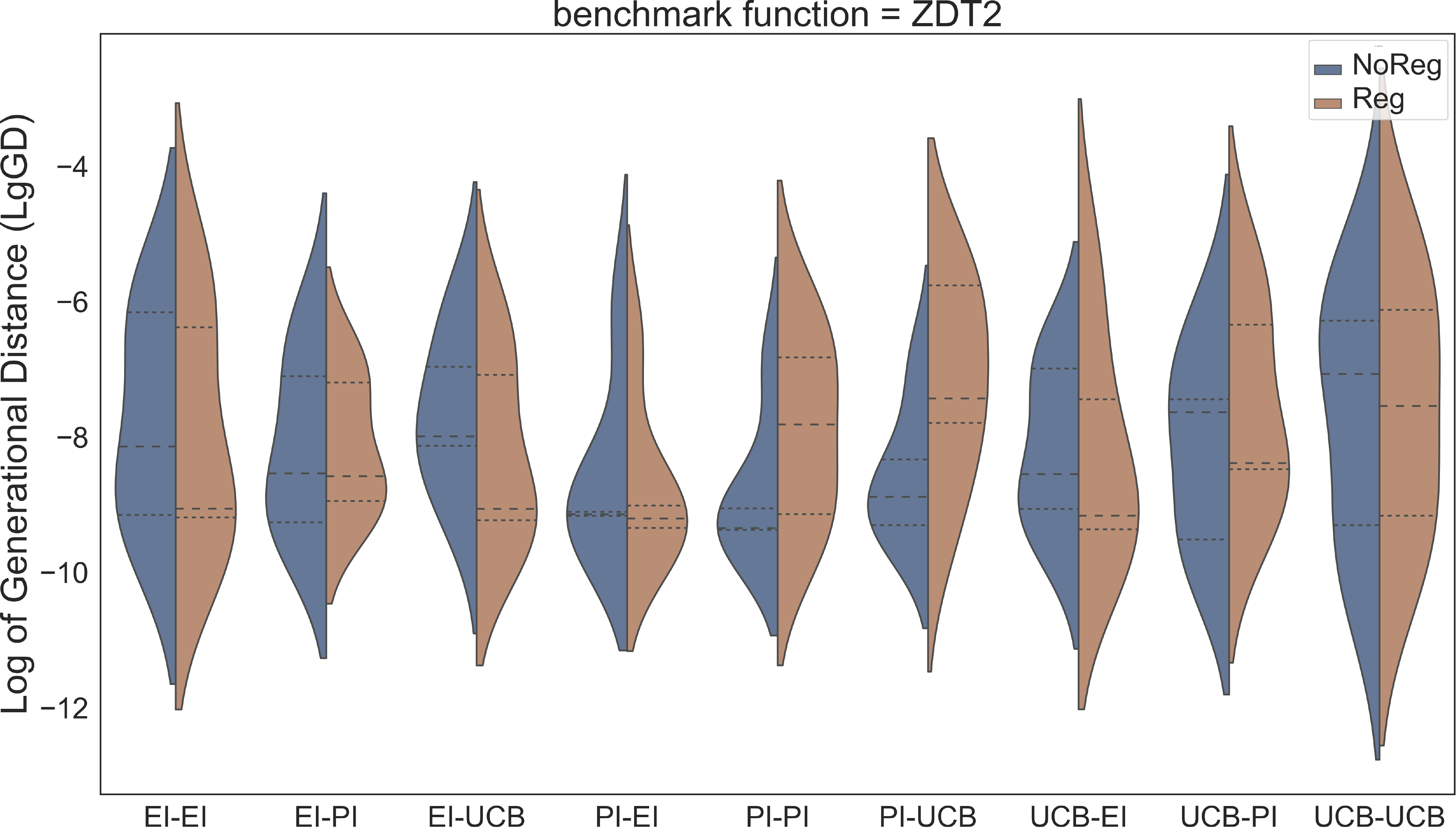}
        \end{subfigure}
        \hfill
        \begin{subfigure}[b]{0.33\textwidth}
        \centering
        \includegraphics[width=1.0\textwidth, keepaspectratio]{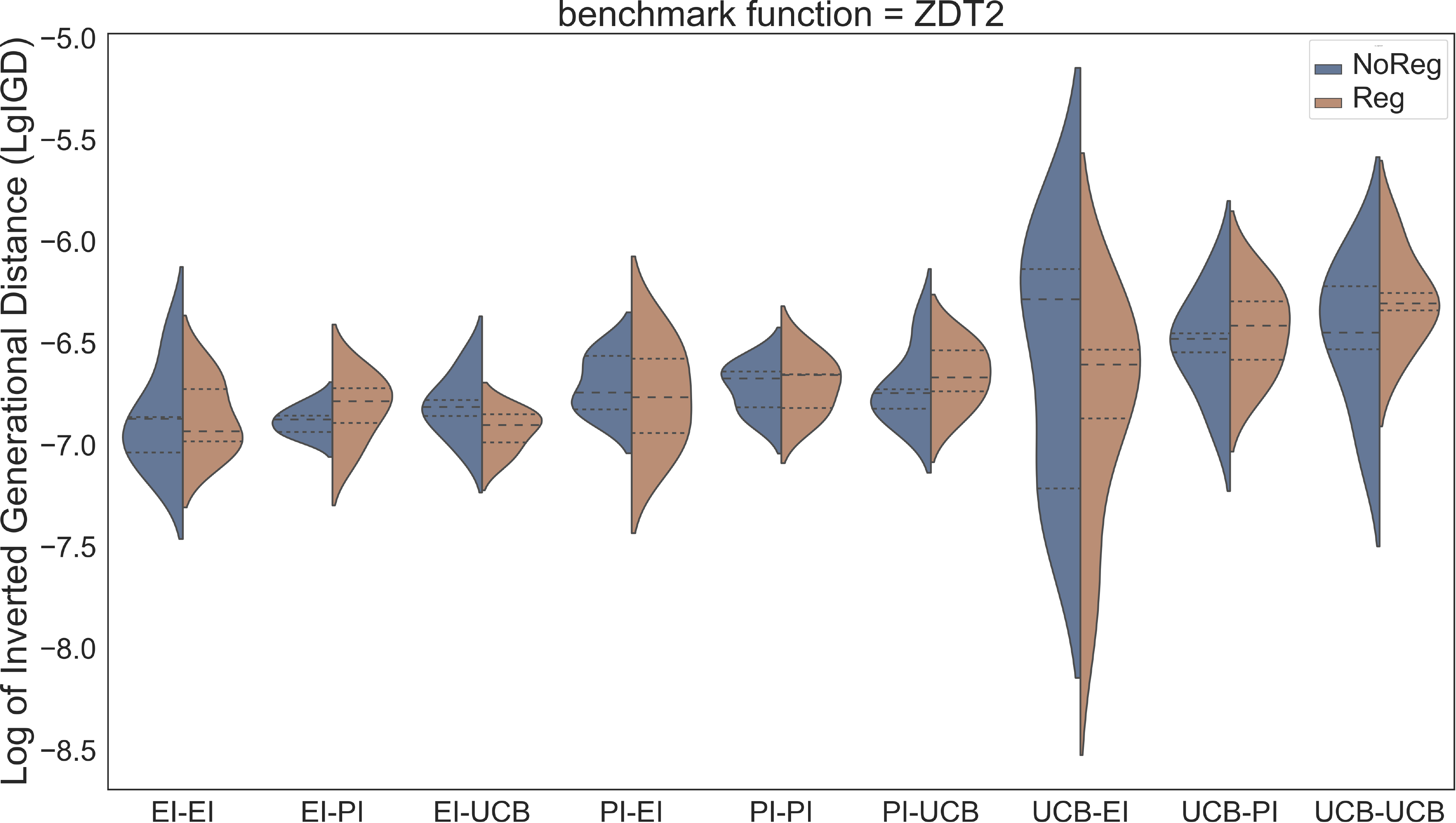}
        \end{subfigure}
        \hfill
        \begin{subfigure}[b]{0.33\textwidth}
        \centering
        \includegraphics[width=1.0\textwidth, keepaspectratio]{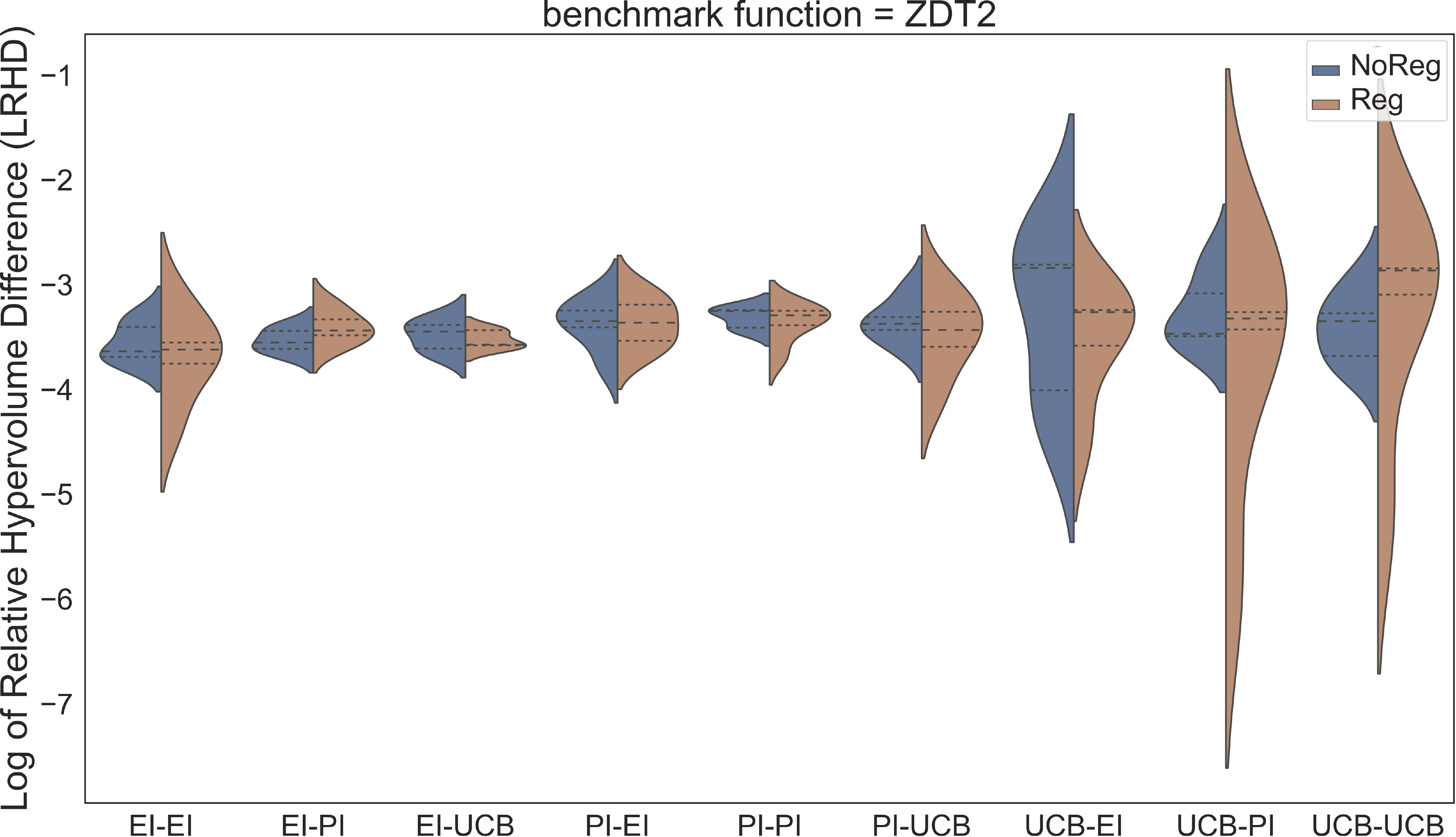}
        \end{subfigure}

        \vfill

        \begin{subfigure}[b]{0.33\textwidth}
        \centering
        \includegraphics[width=1.0\textwidth, keepaspectratio]{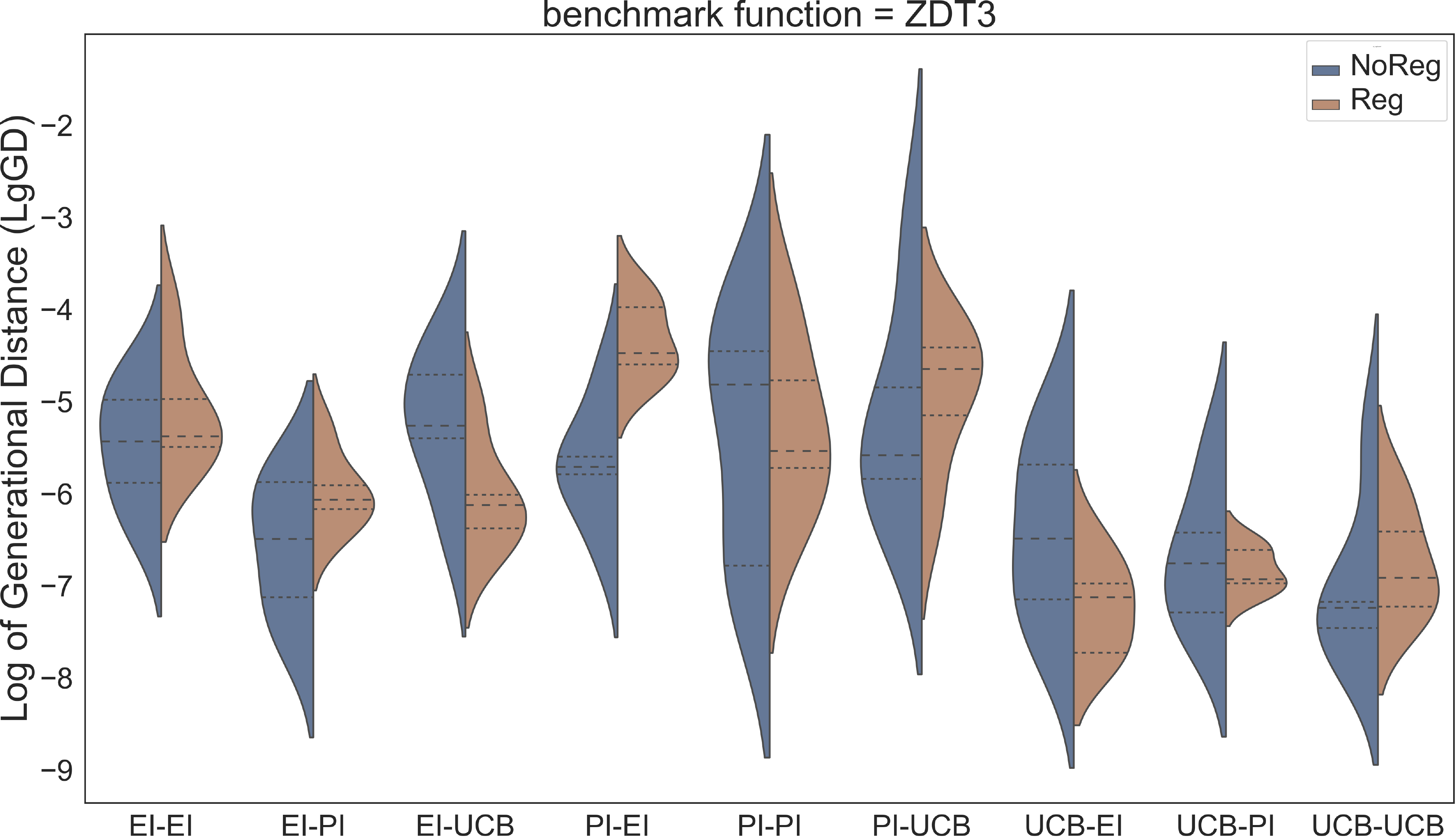}
        \end{subfigure}
        \hfill
        \begin{subfigure}[b]{0.33\textwidth}
        \centering
        \includegraphics[width=1.0\textwidth, keepaspectratio]{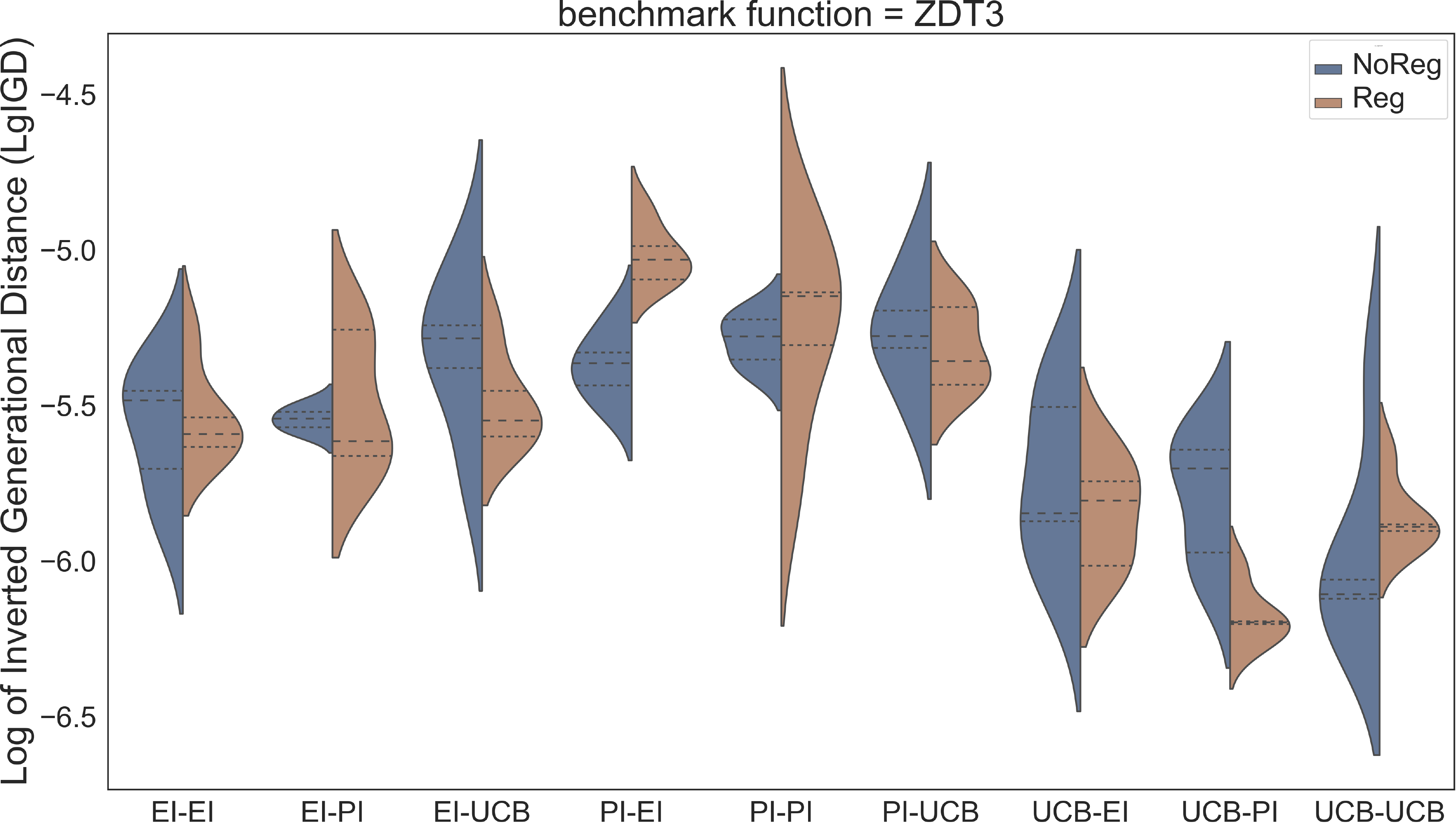}
        \end{subfigure}
        \hfill
        \begin{subfigure}[b]{0.33\textwidth}
        \centering
        \includegraphics[width=1.0\textwidth, keepaspectratio]{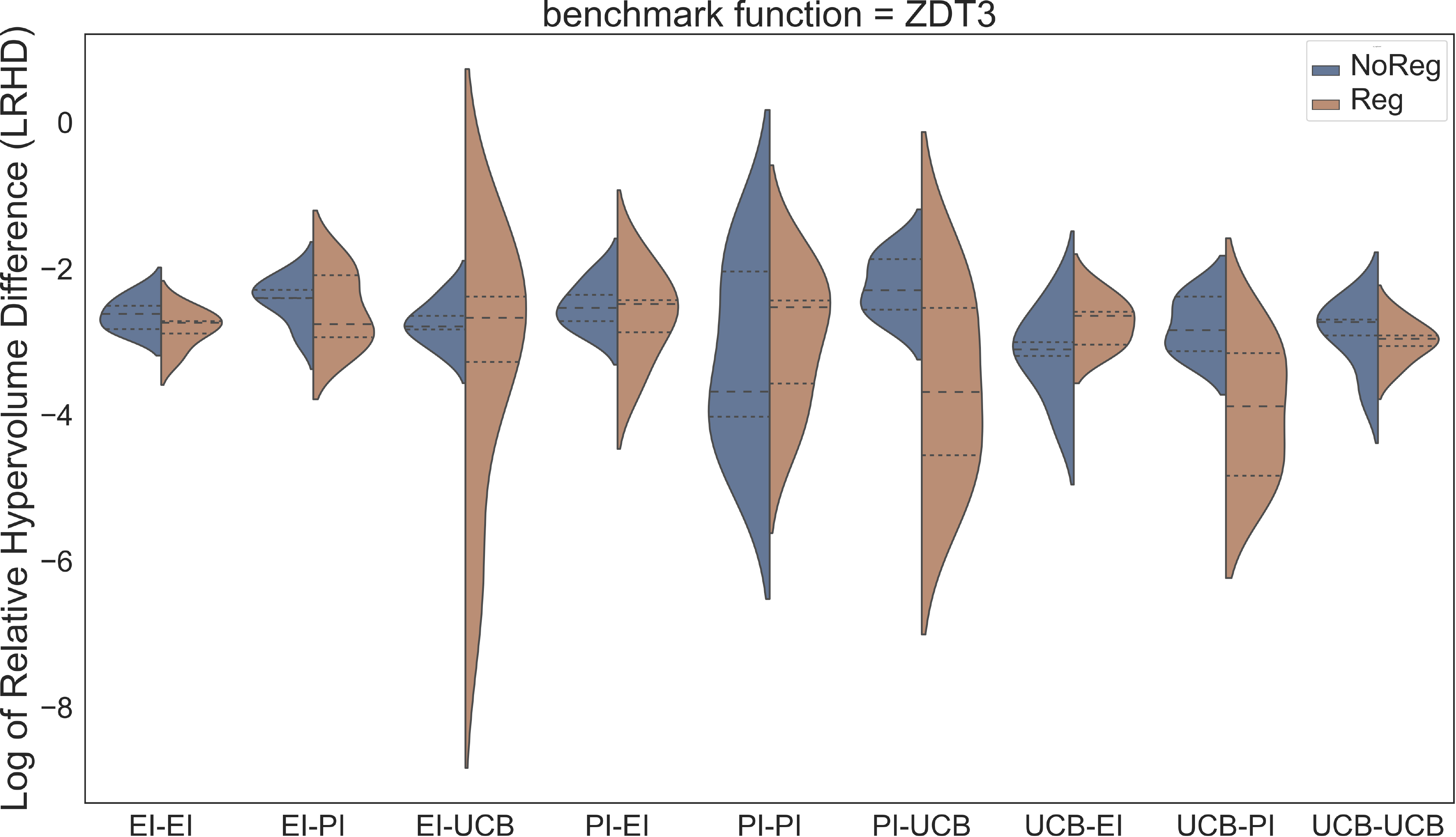}
        \end{subfigure}

        \vfill

        \begin{subfigure}[b]{0.33\textwidth}
        \centering
        \includegraphics[width=1.0\textwidth, keepaspectratio]{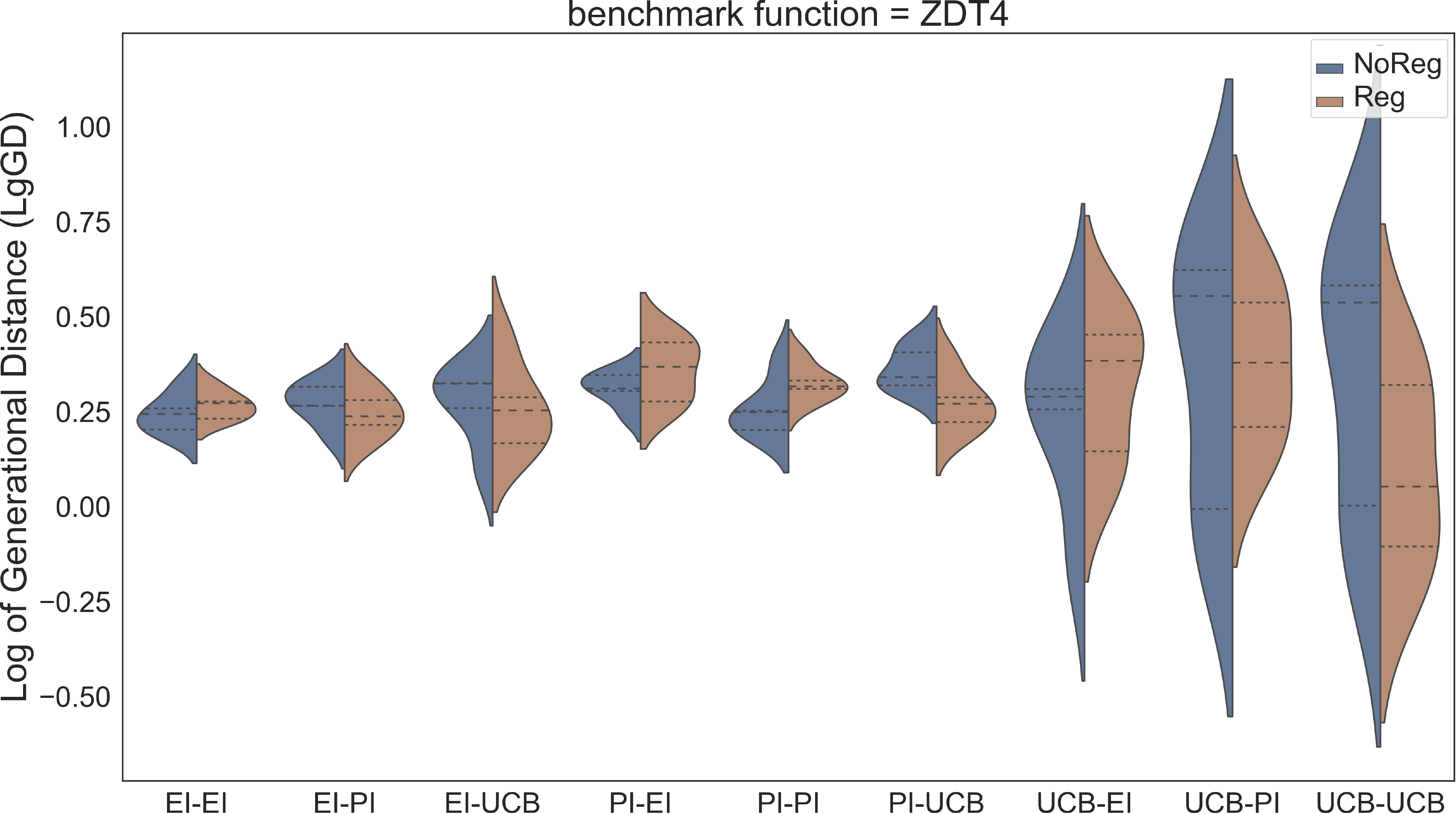}
        \end{subfigure}
        \hfill
        \begin{subfigure}[b]{0.33\textwidth}
        \centering
        \includegraphics[width=1.0\textwidth, keepaspectratio]{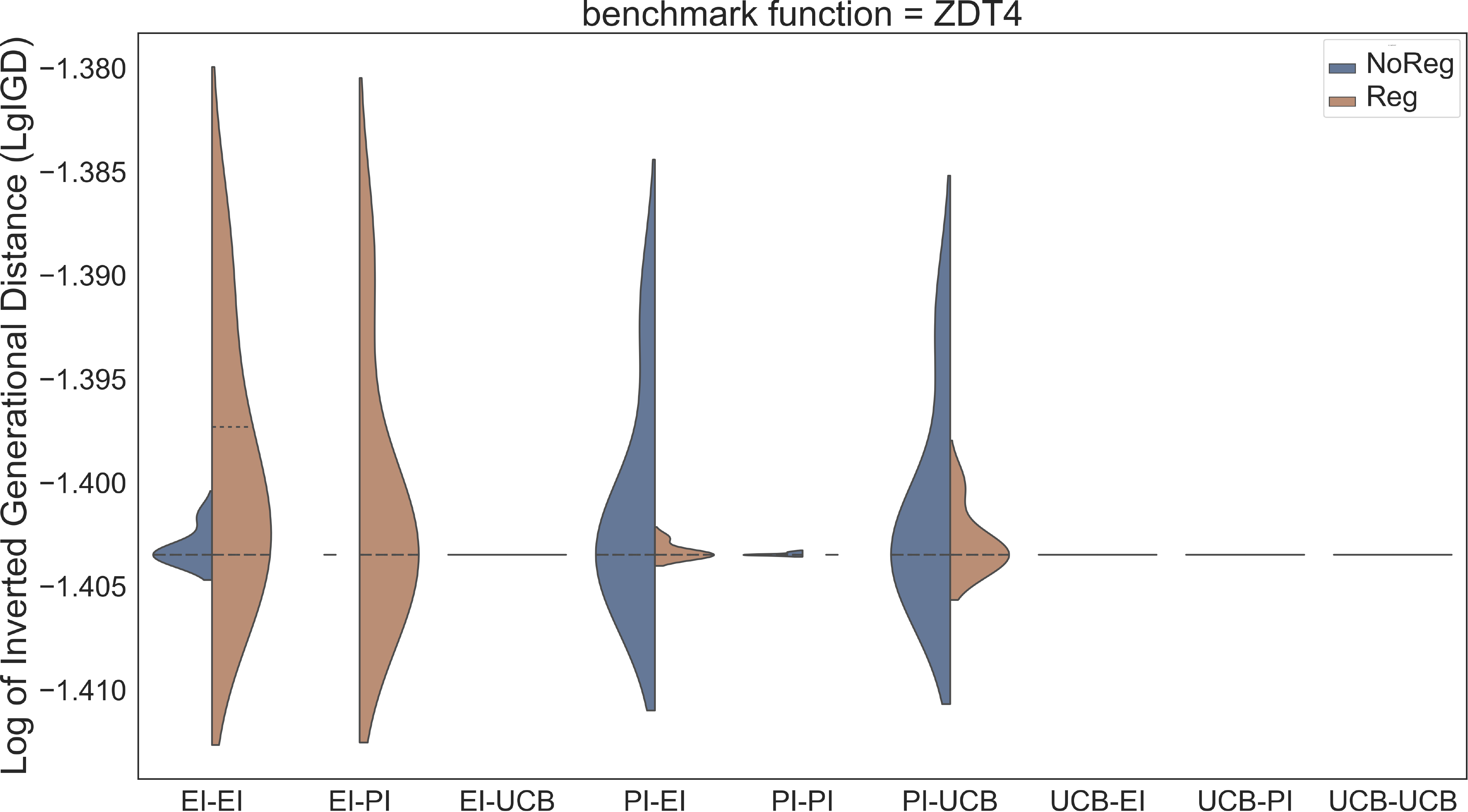}
        \end{subfigure}
        \hfill
        \begin{subfigure}[b]{0.33\textwidth}
        \centering
        \includegraphics[width=1.0\textwidth, keepaspectratio]{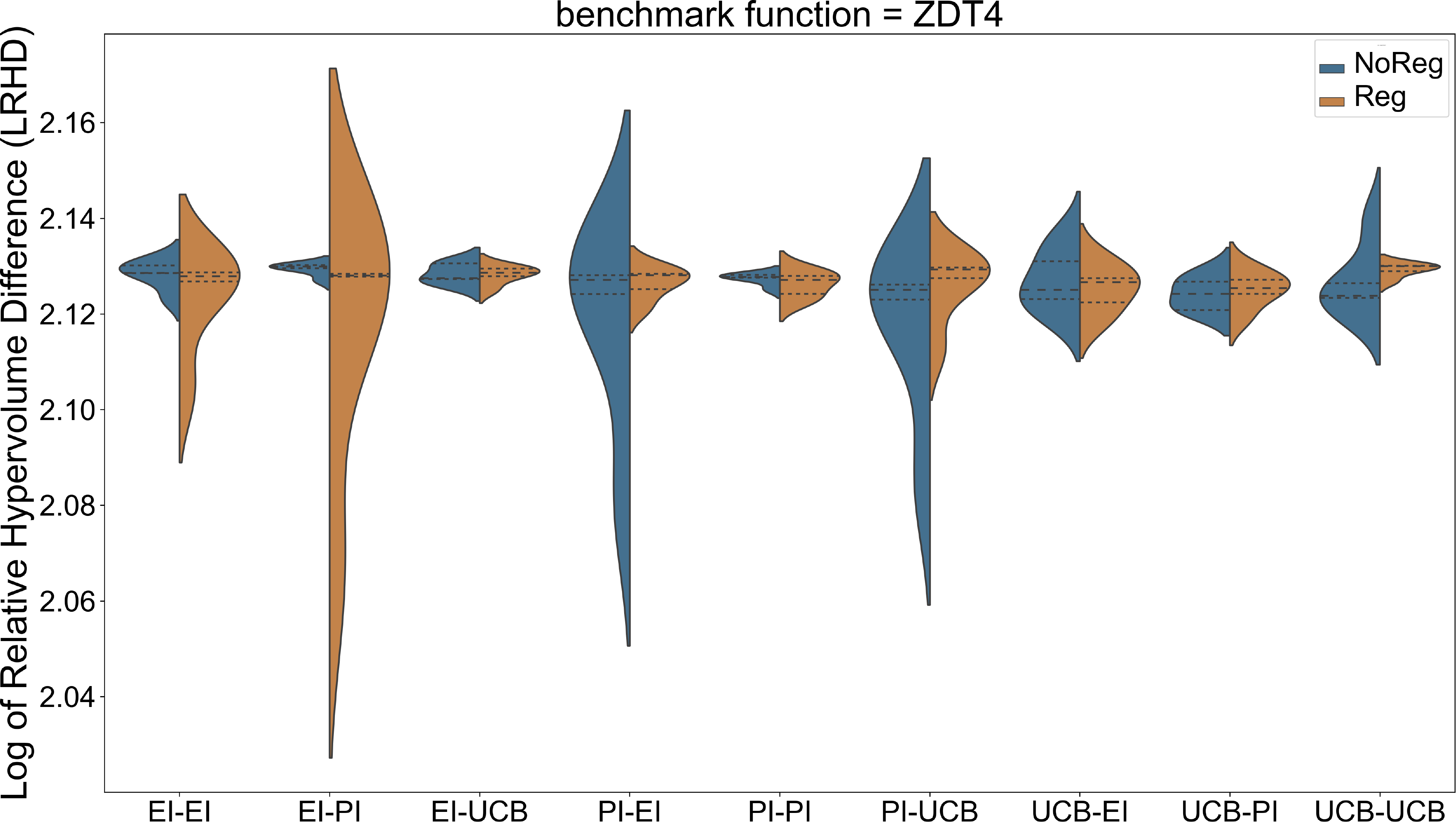}
        \end{subfigure}

        \vfill

        \begin{subfigure}[b]{0.33\textwidth}
        \centering
        \includegraphics[width=1.0\textwidth, keepaspectratio]{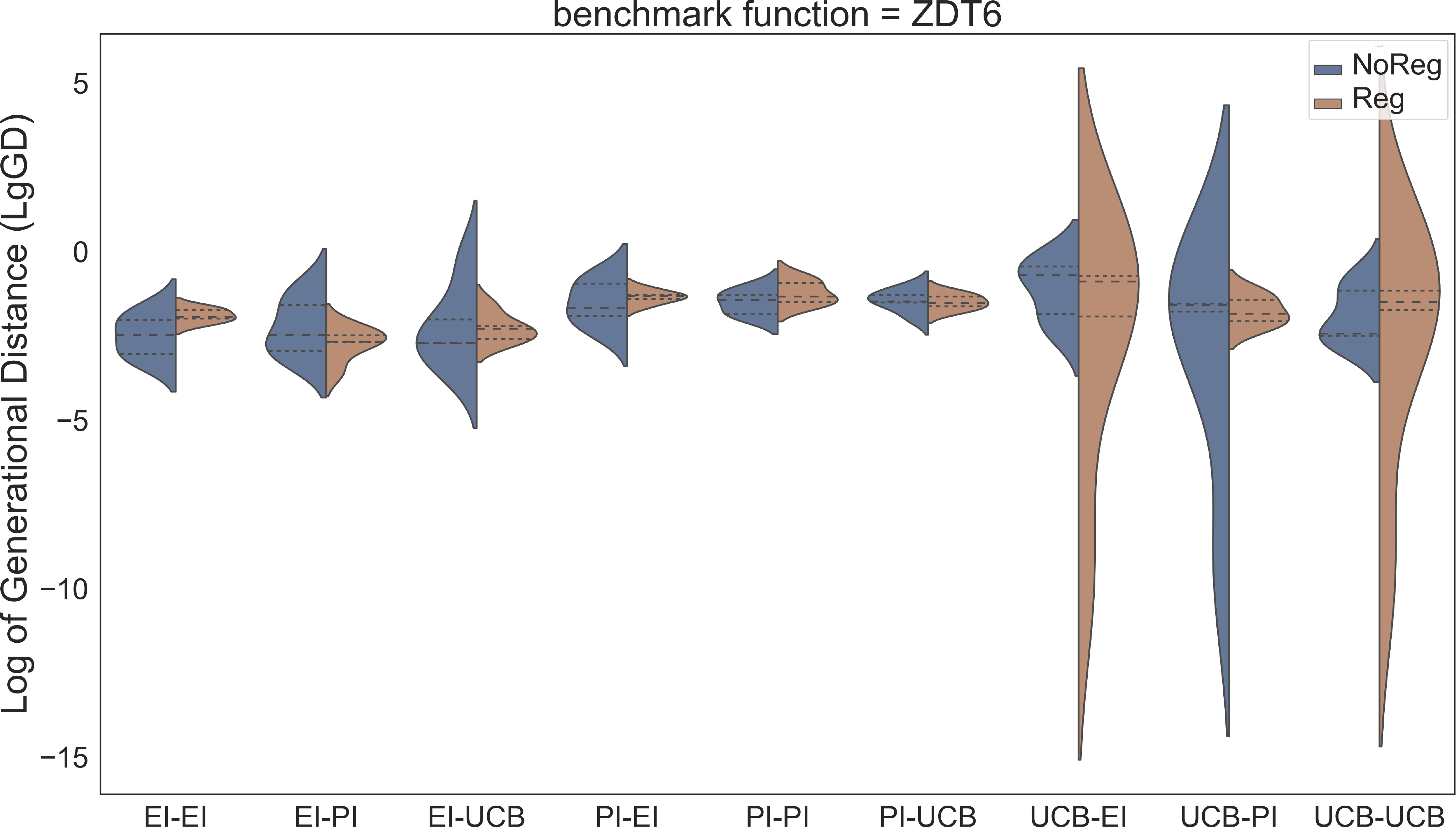}
        \end{subfigure}
        \hfill
        \begin{subfigure}[b]{0.33\textwidth}
        \centering
        \includegraphics[width=1.0\textwidth, keepaspectratio]{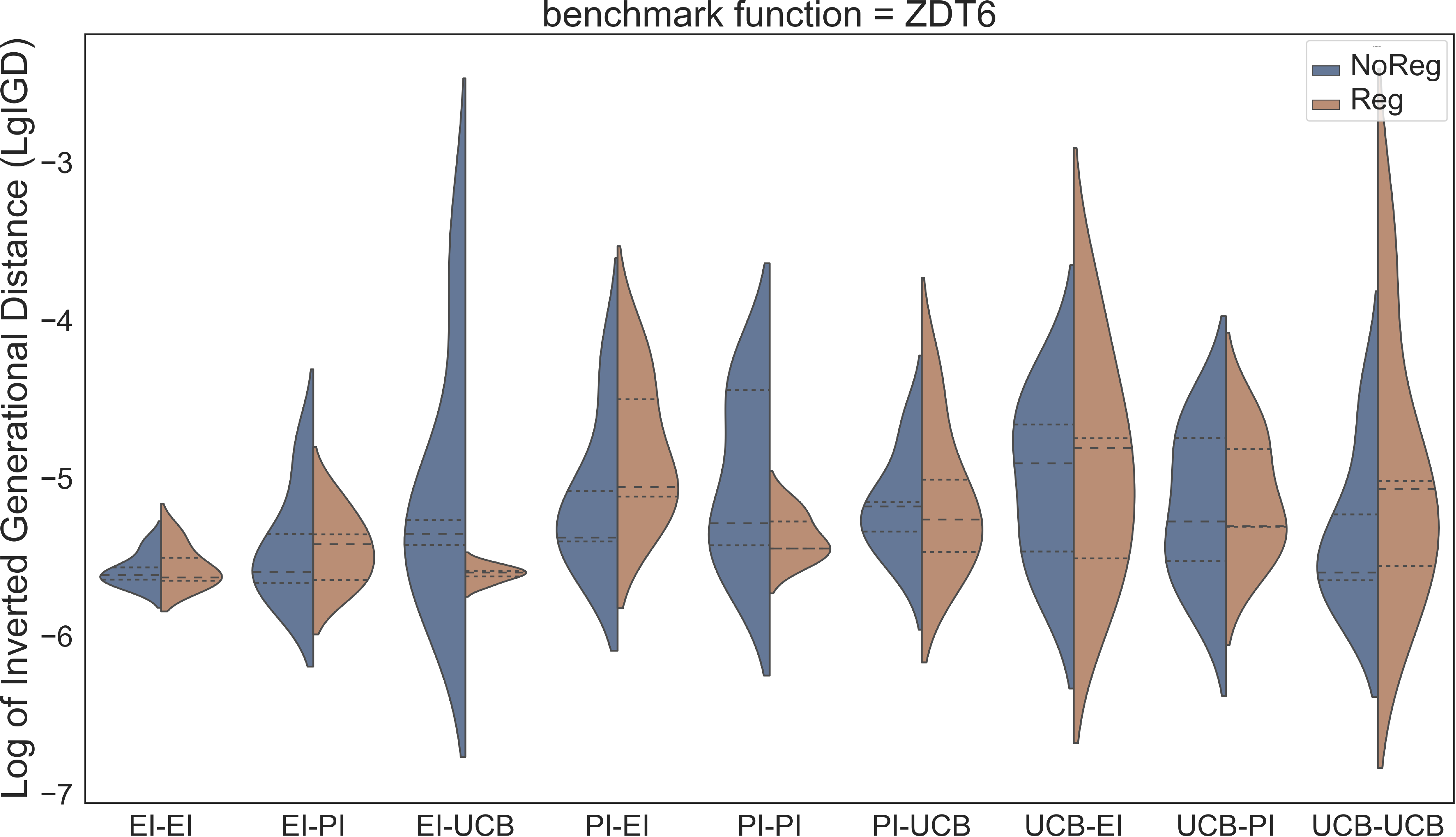}
        \end{subfigure}
        \hfill
        \begin{subfigure}[b]{0.33\textwidth}
        \centering
        \includegraphics[width=1.0\textwidth, keepaspectratio]{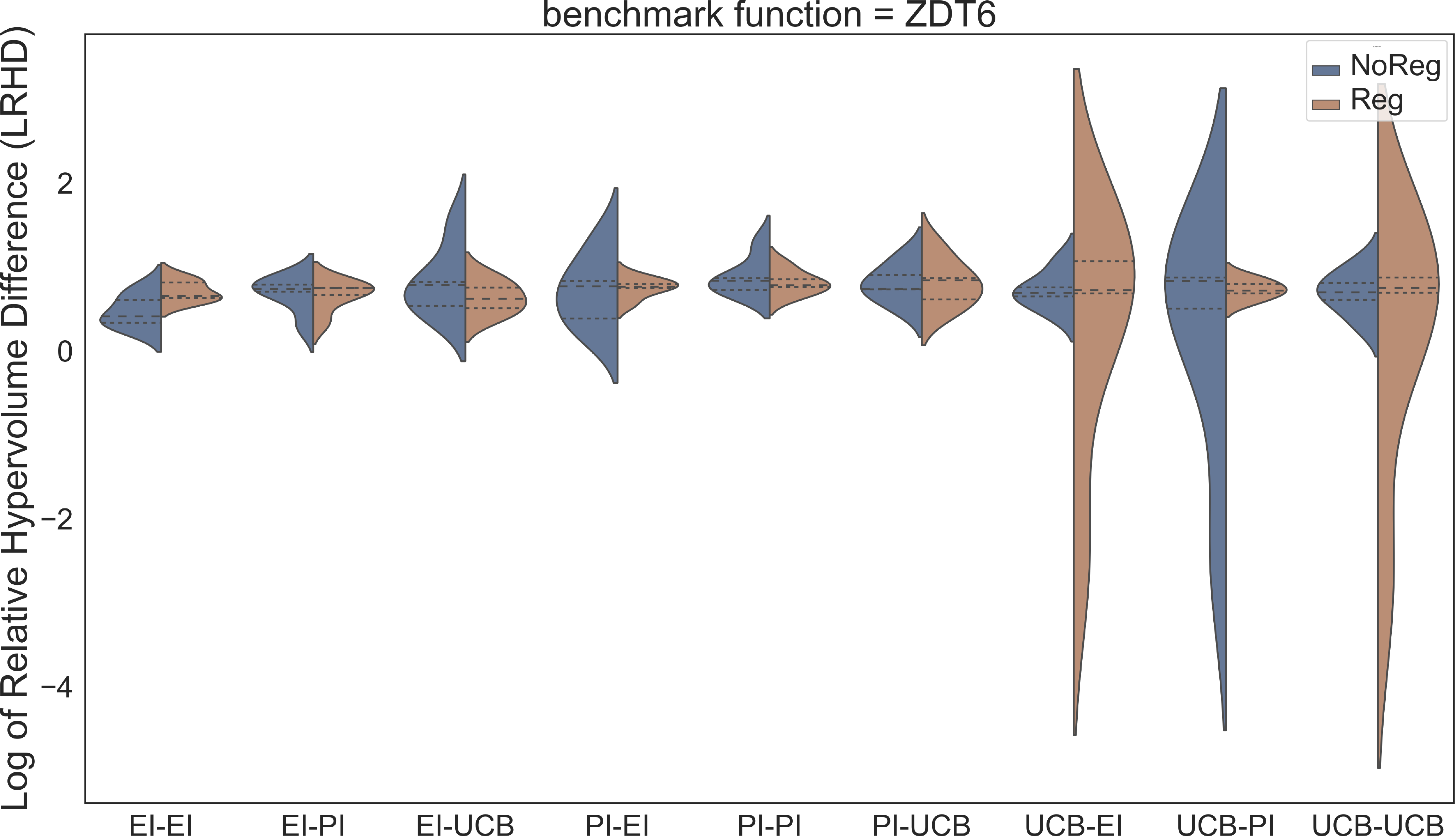}
        \end{subfigure}

\caption{ZDT benchmark results. Left: log of generational distance. Center: log of inverted generational distance. Right: Log of relative hypervolume difference. Labels of methods are denoted as acquisition function of objective function - acquisition function of Pareto frontier. Each row corresponds to a unique benchmark function in the ZDT benchmark test suite.}
\label{fig:zdtViolin}
\end{figure*}

\begin{figure*}[!t]

        \begin{subfigure}[b]{0.33\textwidth}
        \centering
        \includegraphics[width=1.0\textwidth, keepaspectratio]{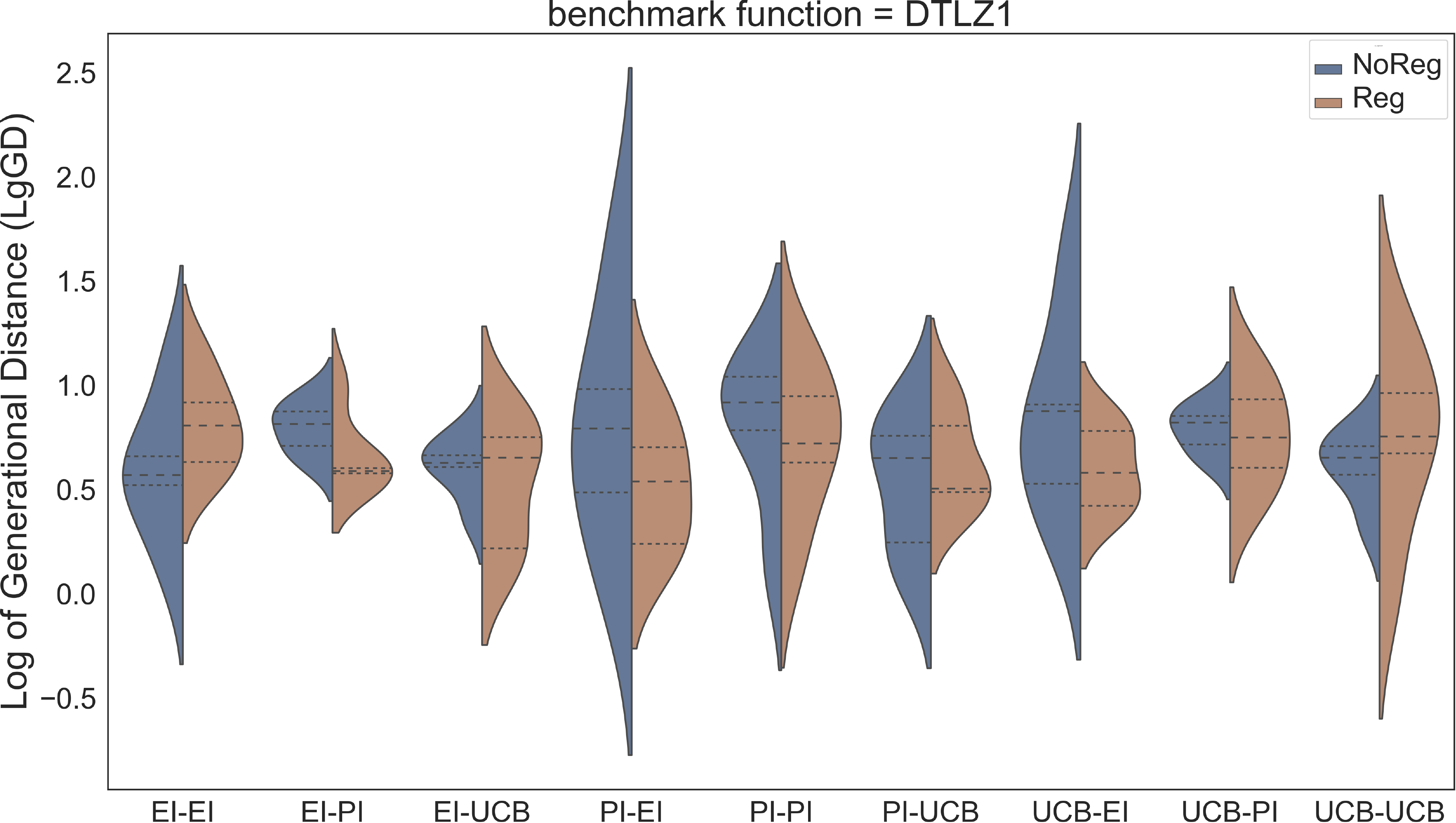}
        \end{subfigure}
        \hfill
        \begin{subfigure}[b]{0.33\textwidth}
        \centering
        \includegraphics[width=1.0\textwidth, keepaspectratio]{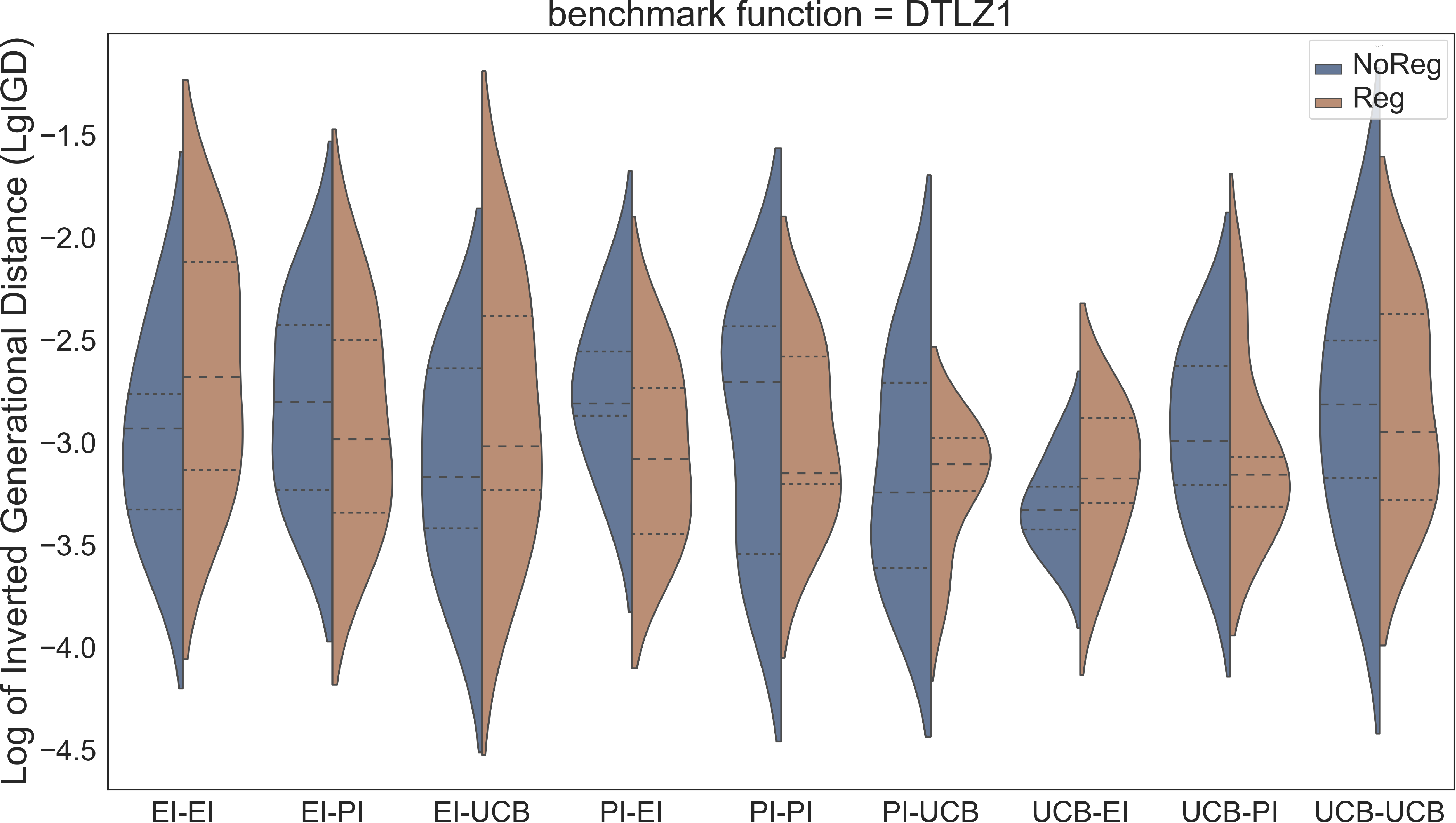}
        \end{subfigure}
        \hfill
        \begin{subfigure}[b]{0.33\textwidth}
        \centering
        \includegraphics[width=1.0\textwidth, keepaspectratio]{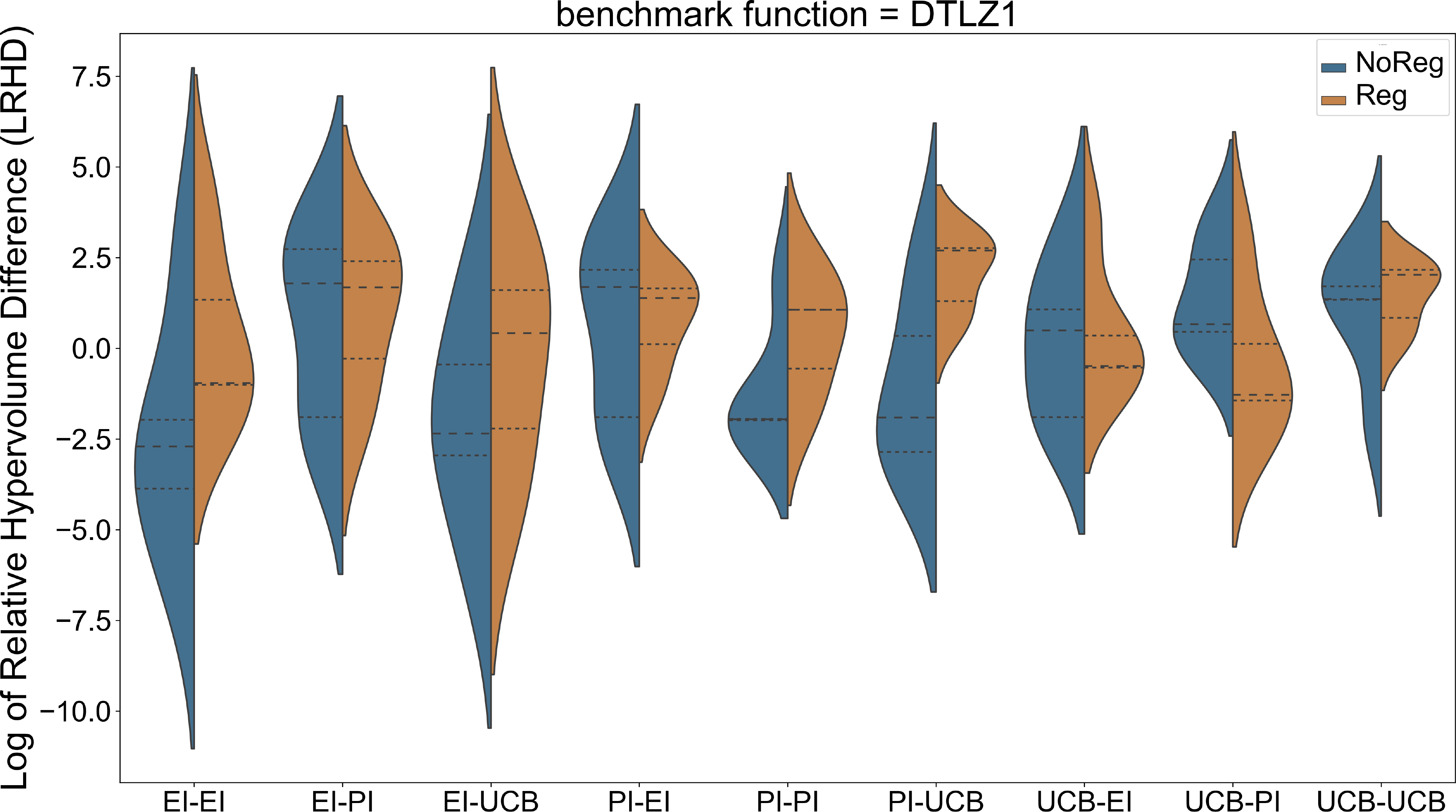}
        \end{subfigure}

        \vfill

        \begin{subfigure}[b]{0.33\textwidth}
        \centering
        \includegraphics[width=1.0\textwidth, keepaspectratio]{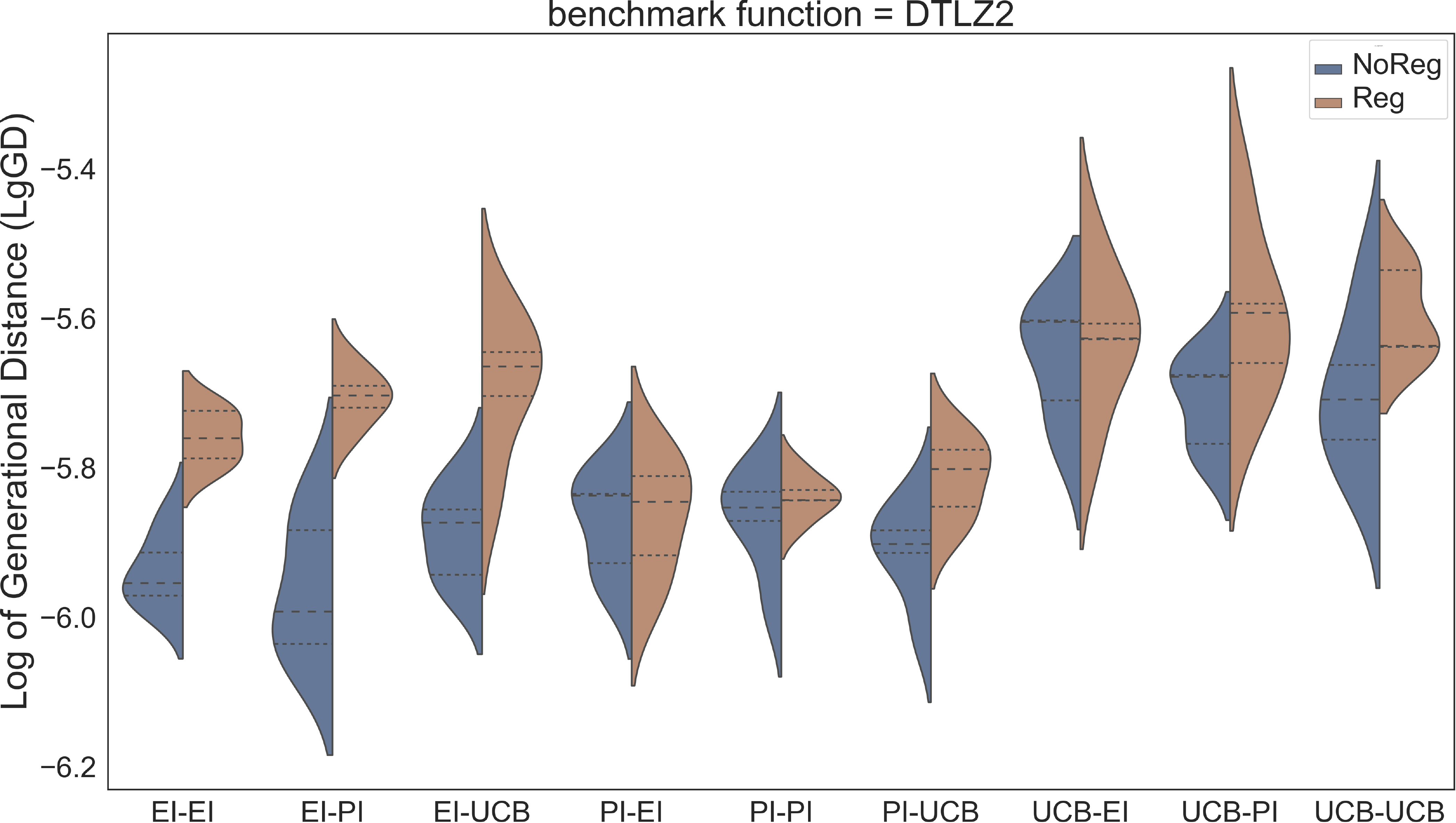}
        \end{subfigure}
        \hfill
        \begin{subfigure}[b]{0.33\textwidth}
        \centering
        \includegraphics[width=1.0\textwidth, keepaspectratio]{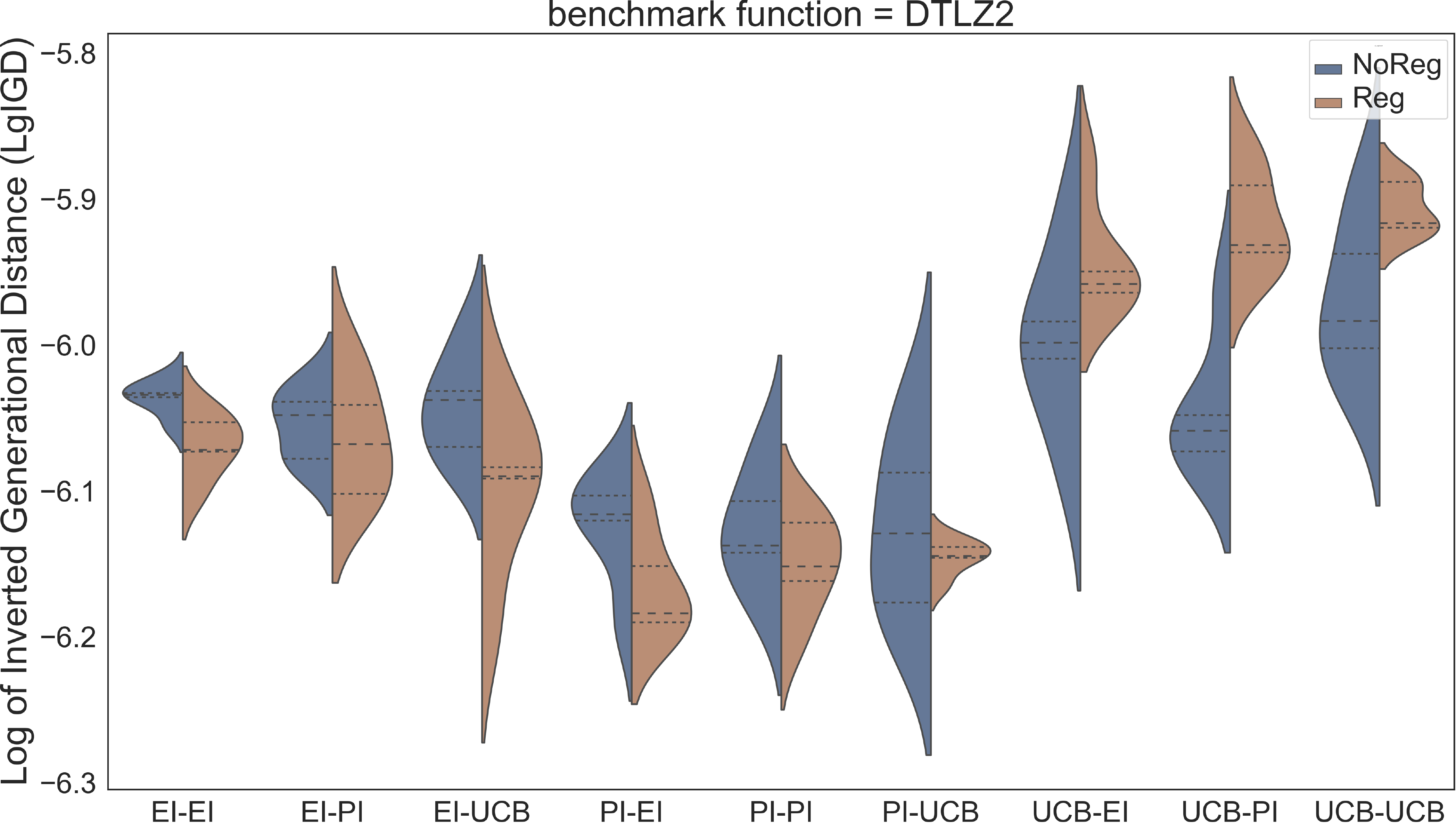}
        \end{subfigure}
        \hfill
        \begin{subfigure}[b]{0.33\textwidth}
        \centering
        \includegraphics[width=1.0\textwidth, keepaspectratio]{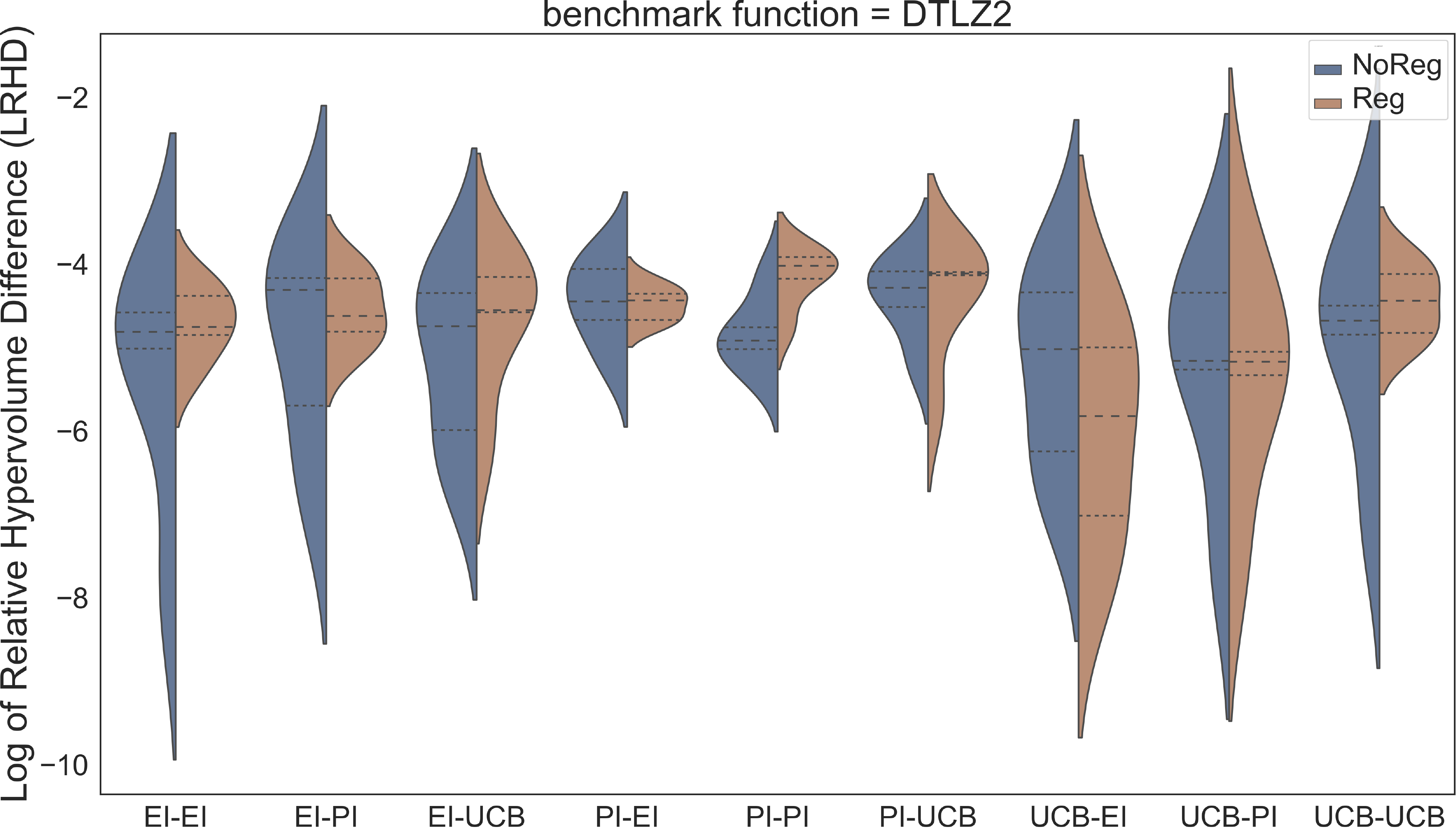}
        \end{subfigure}

        \vfill

        \begin{subfigure}[b]{0.33\textwidth}
        \centering
        \includegraphics[width=1.0\textwidth, keepaspectratio]{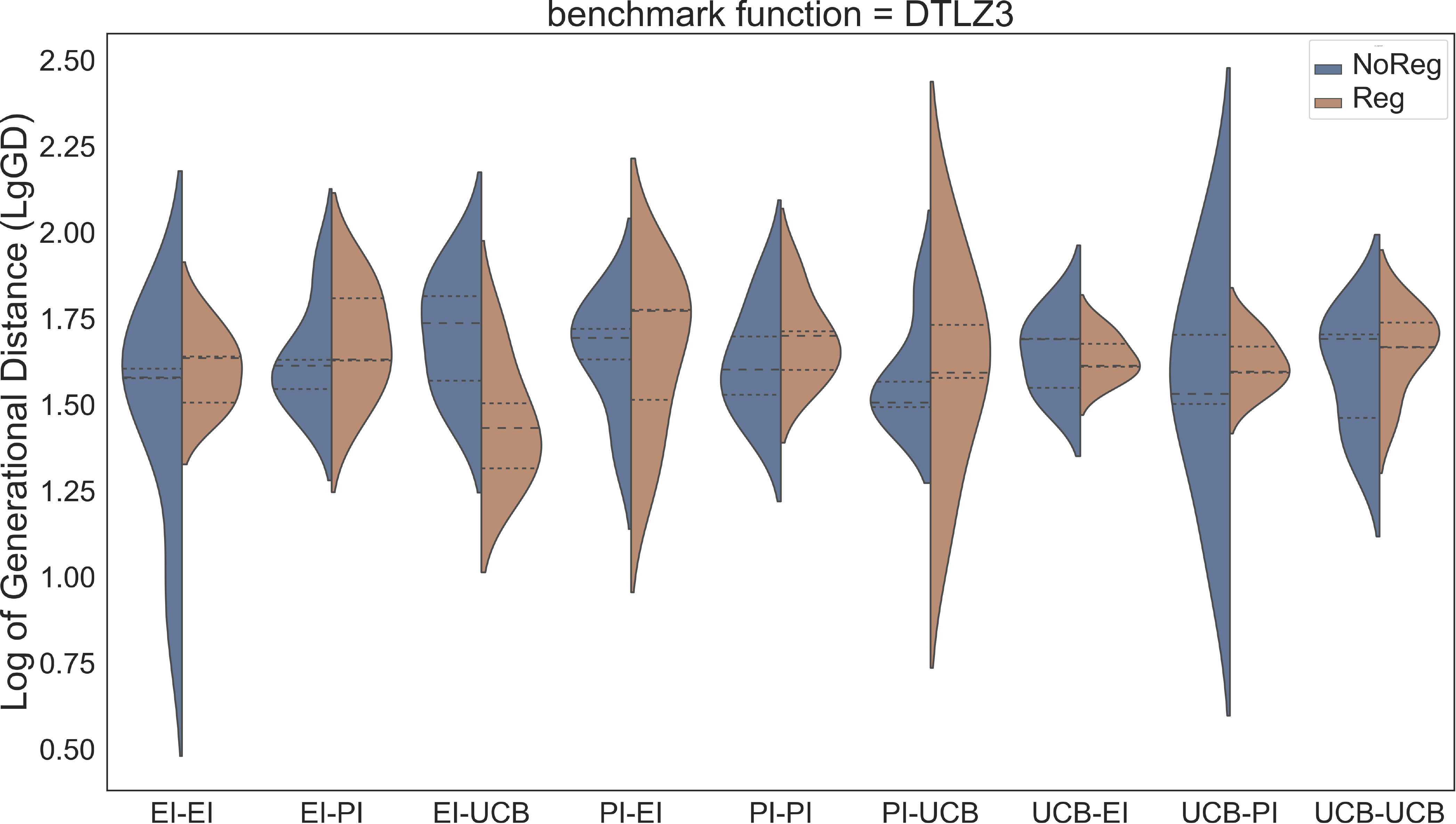}
        \end{subfigure}
        \hfill
        \begin{subfigure}[b]{0.33\textwidth}
        \centering
        \includegraphics[width=1.0\textwidth, keepaspectratio]{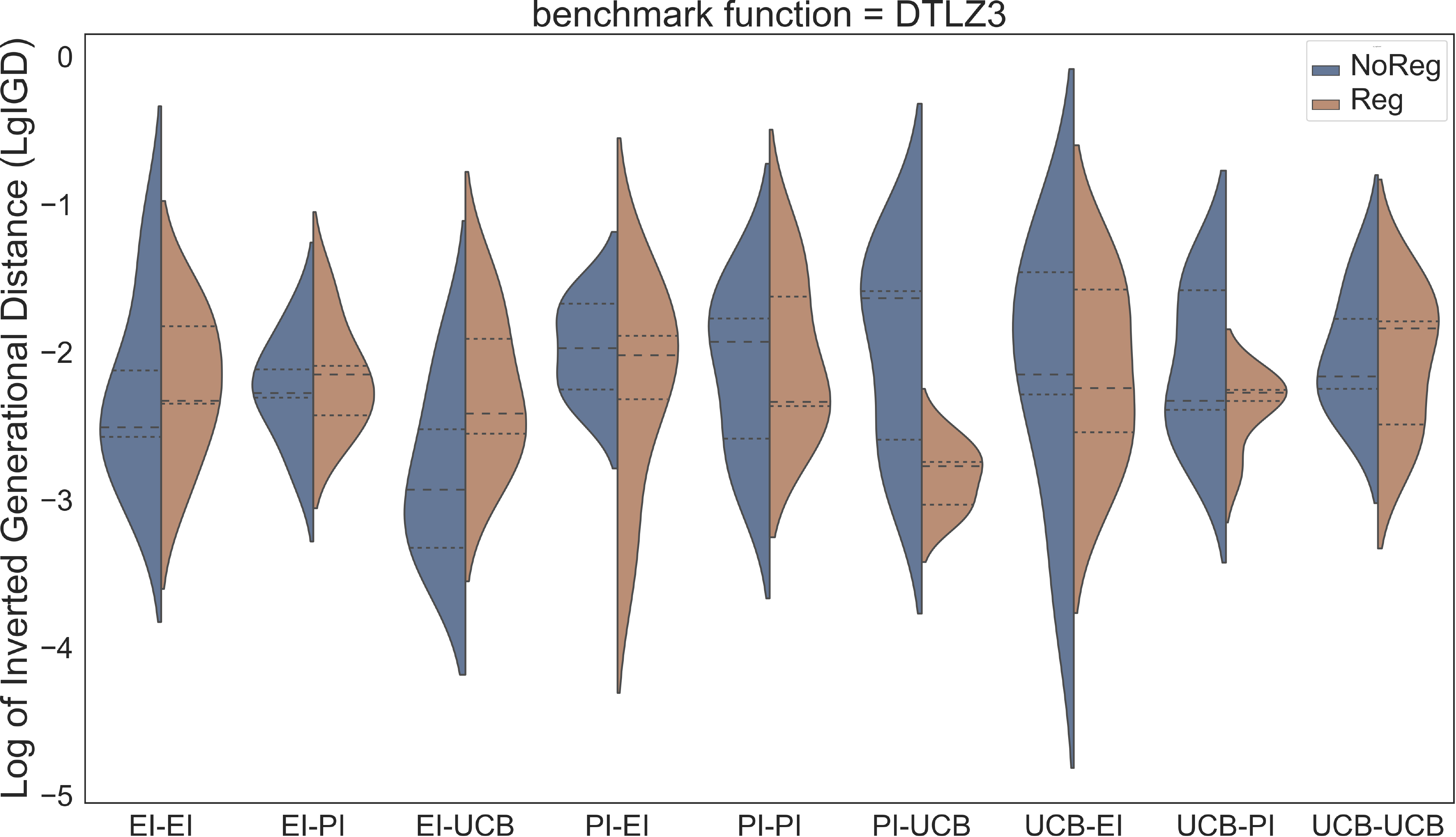}
        \end{subfigure}
        \hfill
        \begin{subfigure}[b]{0.33\textwidth}
        \centering
        \includegraphics[width=1.0\textwidth, keepaspectratio]{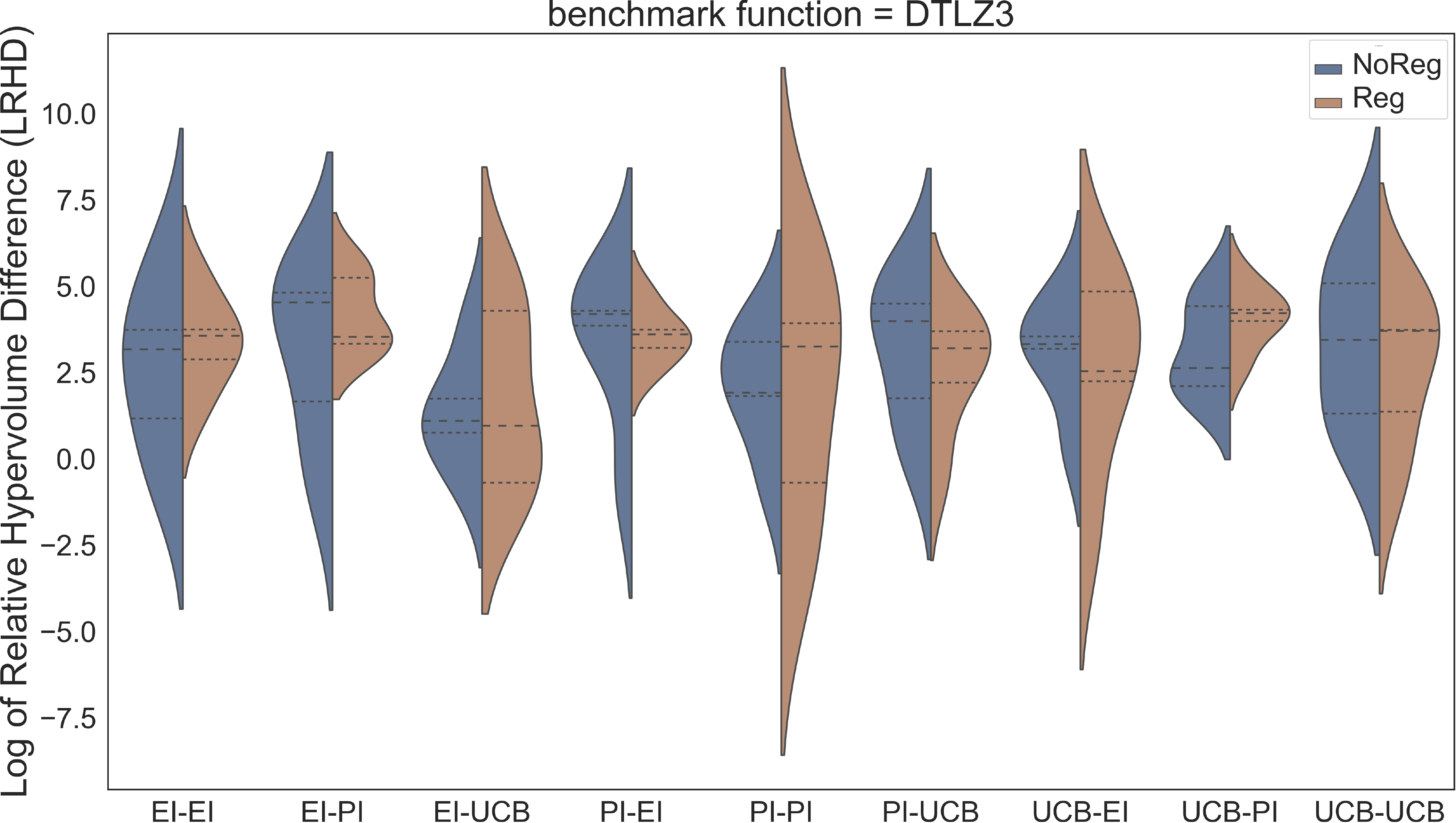}
        \end{subfigure}

        \vfill

        \begin{subfigure}[b]{0.33\textwidth}
        \centering
        \includegraphics[width=1.0\textwidth, keepaspectratio]{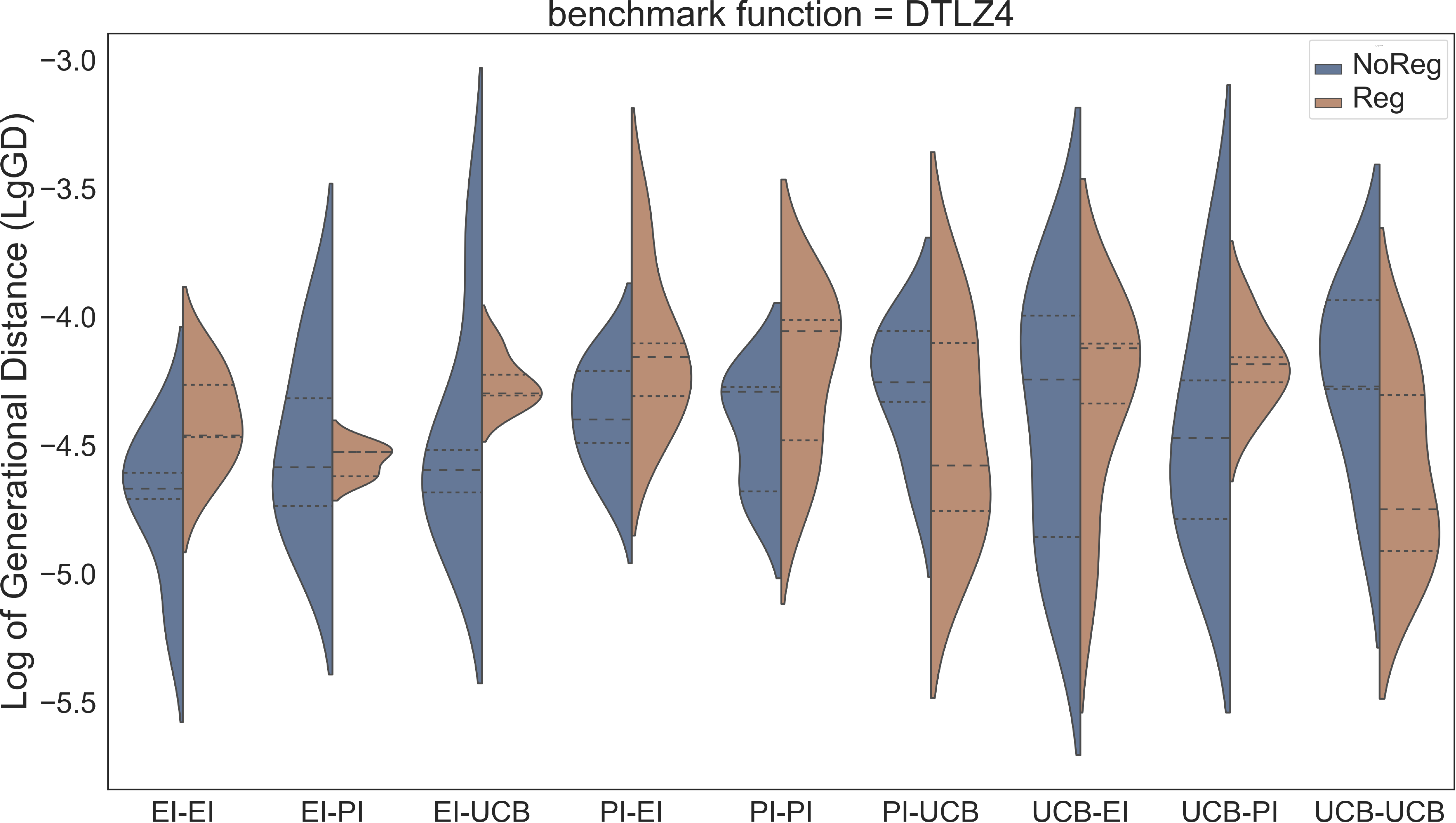}
        \end{subfigure}
        \hfill
        \begin{subfigure}[b]{0.33\textwidth}
        \centering
        \includegraphics[width=1.0\textwidth, keepaspectratio]{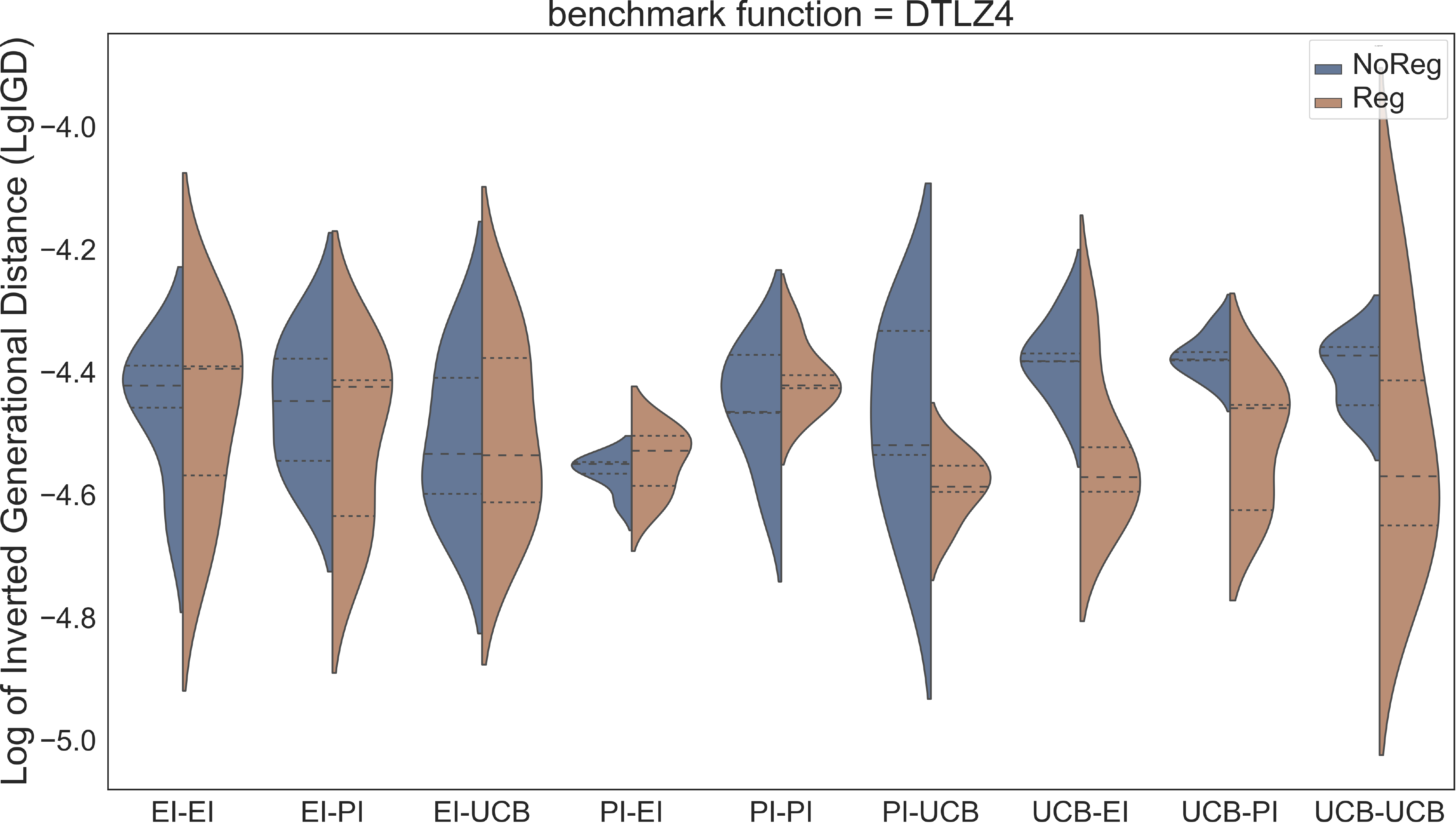}
        \end{subfigure}
        \hfill
        \begin{subfigure}[b]{0.33\textwidth}
        \centering
        \includegraphics[width=1.0\textwidth, keepaspectratio]{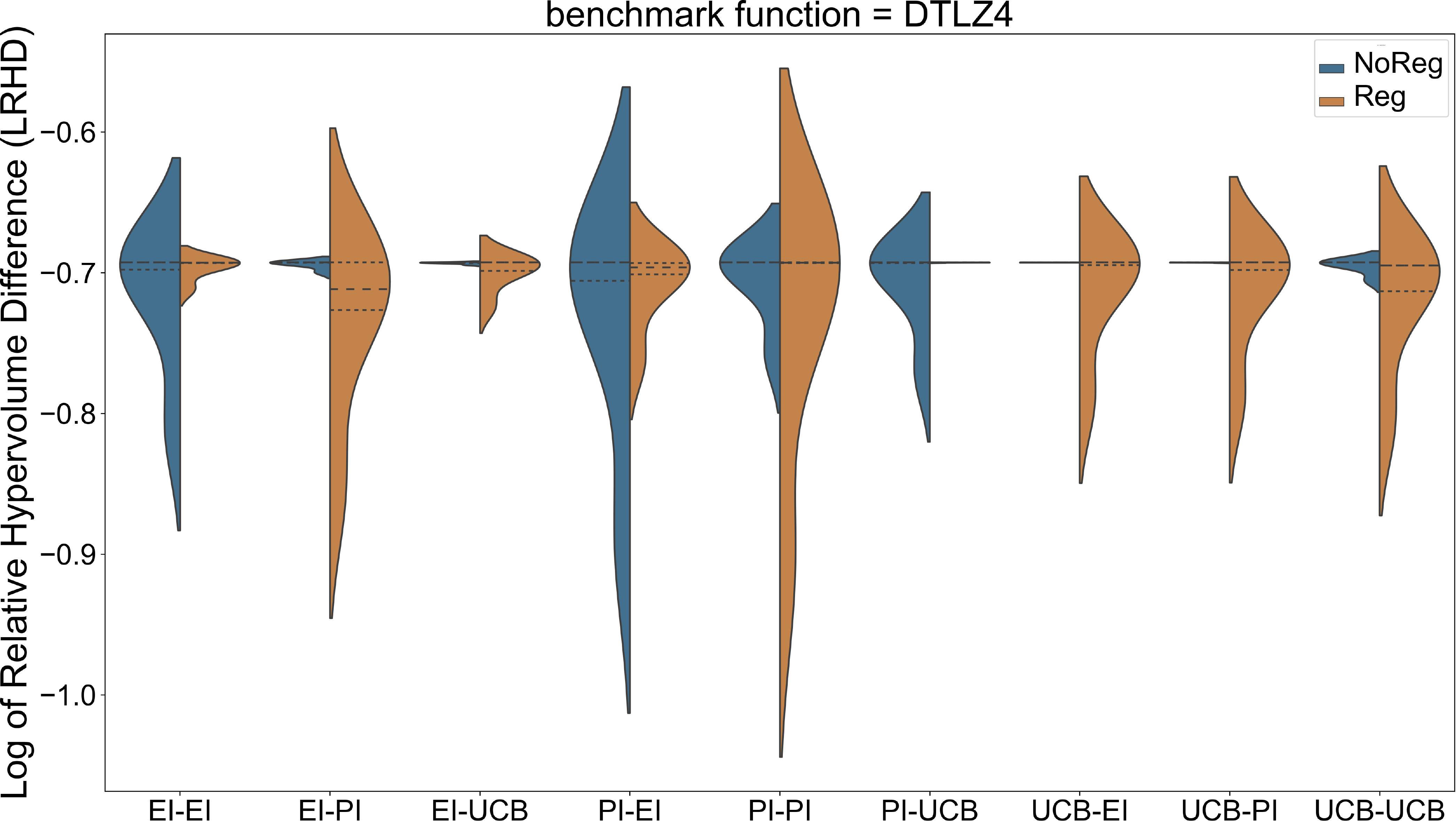}
        \end{subfigure}

        \vfill

        \begin{subfigure}[b]{0.33\textwidth}
        \centering
        \includegraphics[width=1.0\textwidth, keepaspectratio]{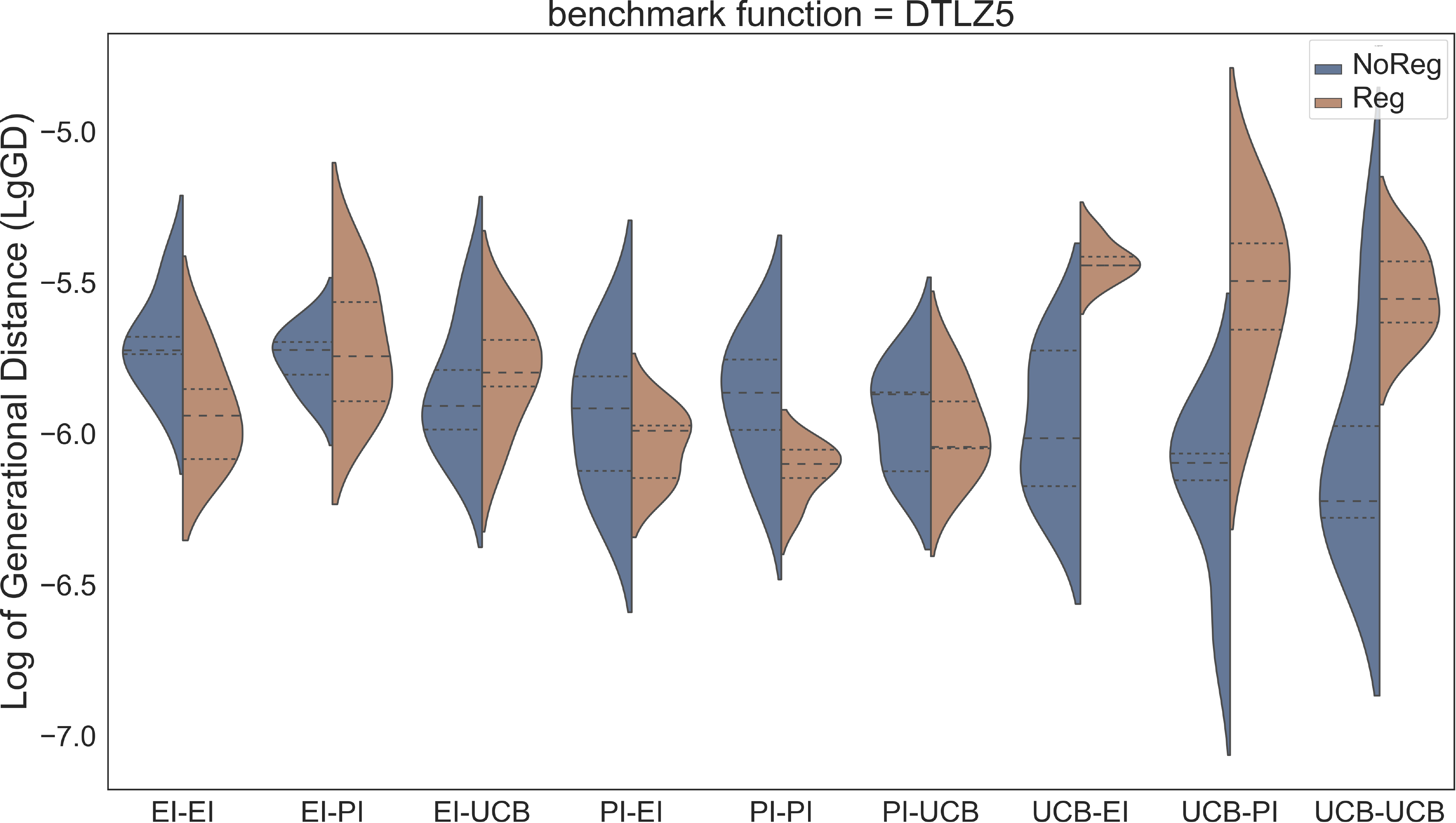}
        \end{subfigure}
        \hfill
        \begin{subfigure}[b]{0.33\textwidth}
        \centering
        \includegraphics[width=1.0\textwidth, keepaspectratio]{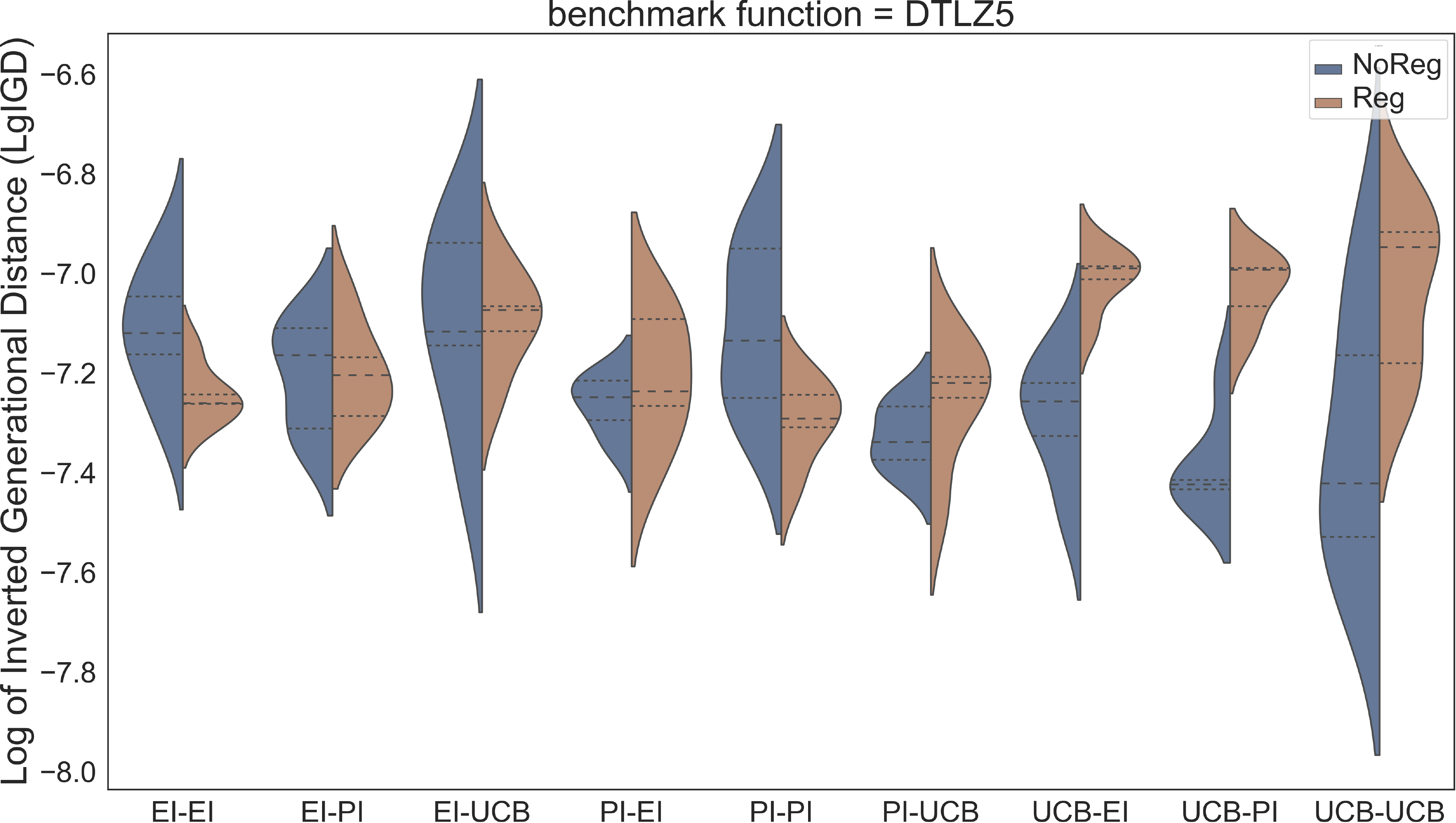}
        \end{subfigure}
        \hfill
        \begin{subfigure}[b]{0.33\textwidth}
        \centering
        \includegraphics[width=1.0\textwidth, keepaspectratio]{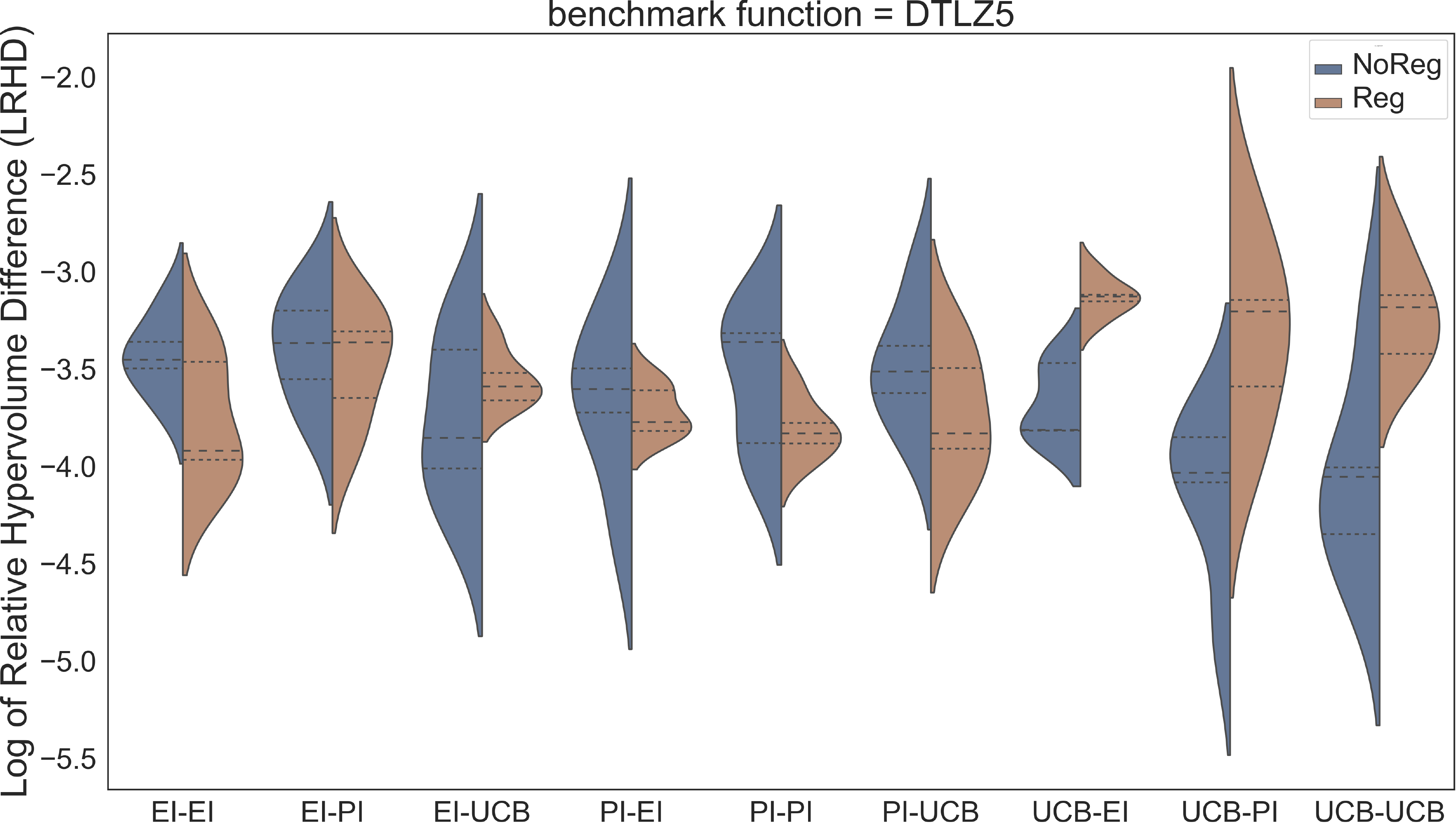}
        \end{subfigure}

        \vfill

        \begin{subfigure}[b]{0.33\textwidth}
        \centering
        \includegraphics[width=1.0\textwidth, keepaspectratio]{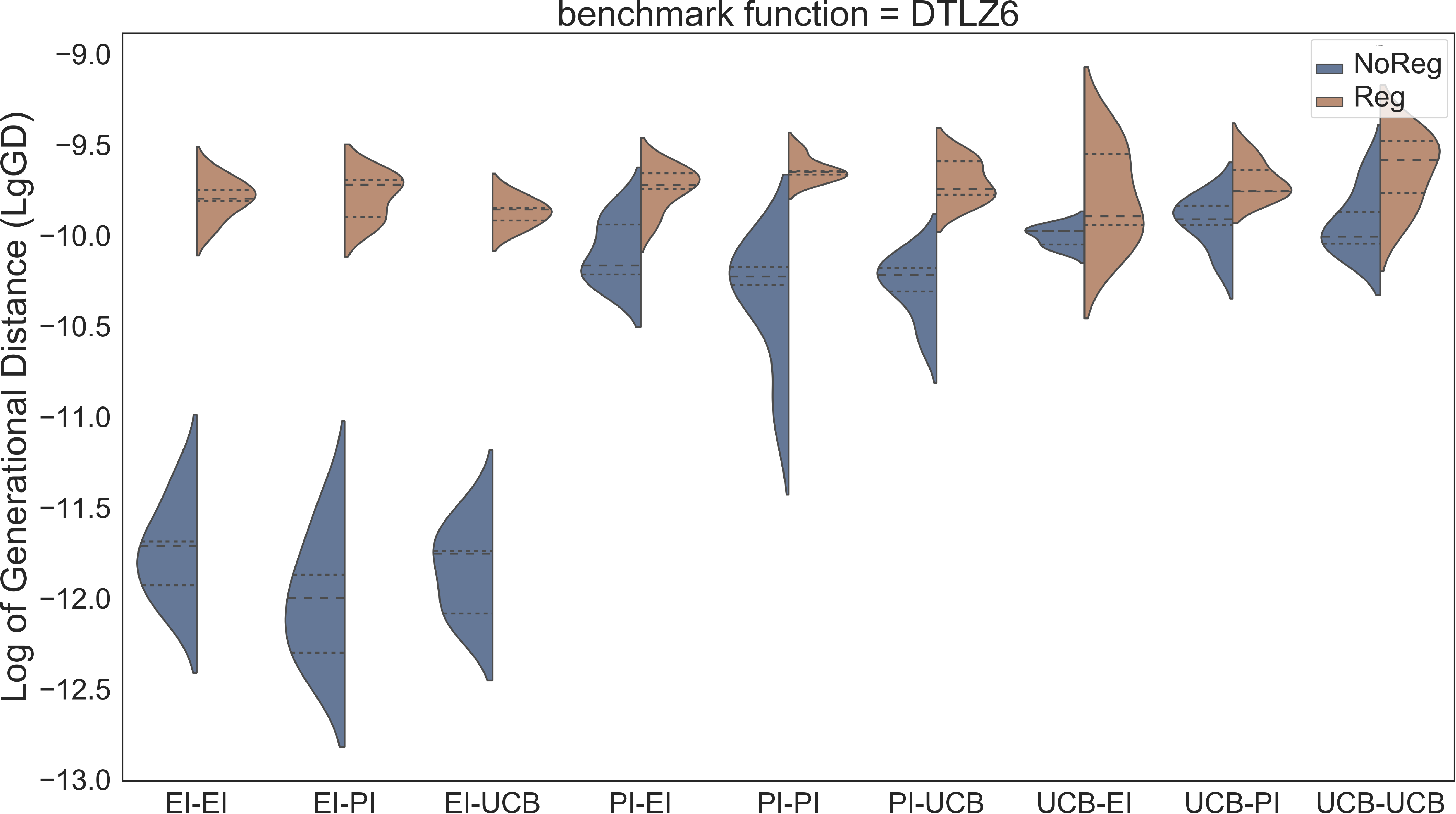}
        \end{subfigure}
        \hfill
        \begin{subfigure}[b]{0.33\textwidth}
        \centering
        \includegraphics[width=1.0\textwidth, keepaspectratio]{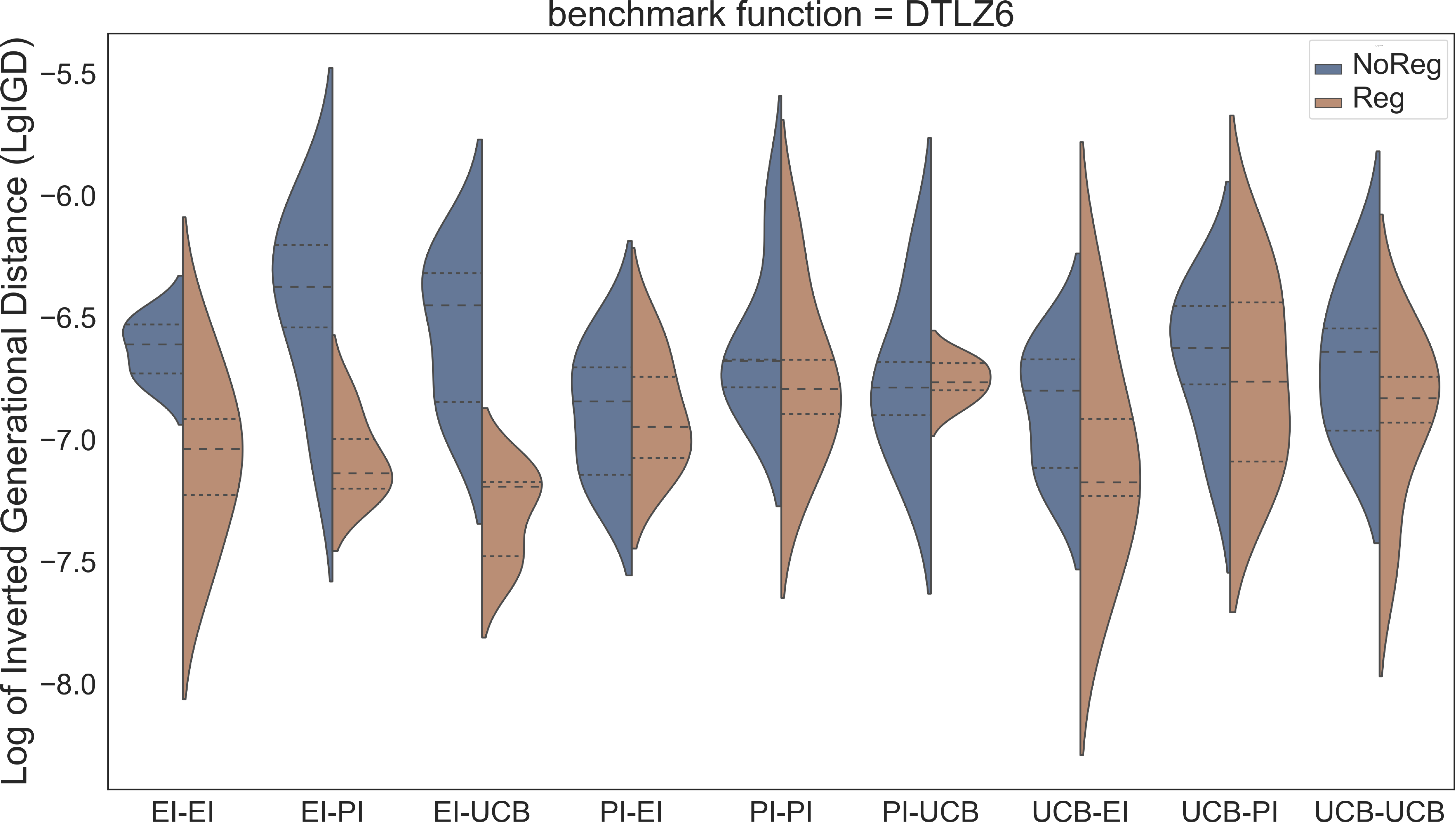}
        \end{subfigure}
        \hfill
        \begin{subfigure}[b]{0.33\textwidth}
        \centering
        \includegraphics[width=1.0\textwidth, keepaspectratio]{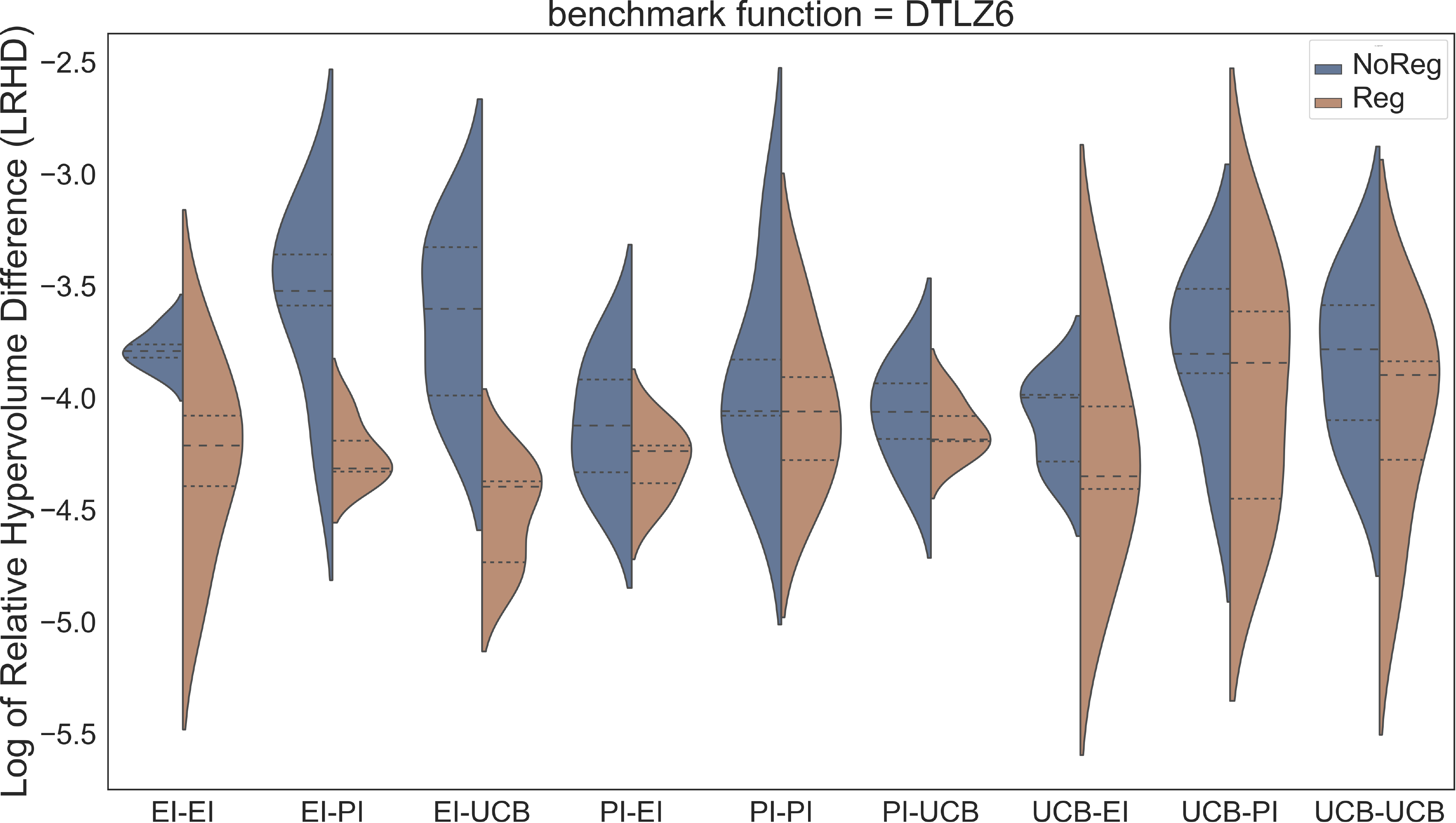}
        \end{subfigure}

\caption{DTLZ benchmark results. Left: log of generational distance. Center: log of inverted generational distance. Right: Log of relative hypervolume difference. Labels of methods are denoted as acquisition function of objective function - acquisition function of Pareto frontier. Each row corresponds to a unique benchmark function in the DTLZ benchmark test suite.}
\label{fig:dtlzViolin}
\end{figure*}

In Figures \ref{fig:zdtViolin} and \ref{fig:dtlzViolin}, the label of the $x$-axis is noted by the objective acquisition function, followed by the Pareto acquisition function. For example, UCB-PI means UCB is used as the objective acquisition function for the scalarized Tchebycheff, and PI is used to search for the Pareto frontier. 
Seaborn package \cite{waskom2020seaborn} is used to visualize the performance metrics, measure the log of generational distance, log of inverted generational distance, and log of relative hypervolume difference. 

Figure \ref{fig:zdtViolin} shows the comparison between 18 variants of the srMO-BO-3GP methods in ZDT test suite, including regularized vs. non-regularized augmented Tchebycheff scalarized single-objective functions. 
Overall, regularized objective function tends to slightly outperform non-regularized objective function. 
For the Pareto frontier acquisition function, EI and PI acquisition functions performs better than UCB acquisition function. 
However, for the objective acquisition function, UCB seems to perform best, then to EI, then to PI. 

\begin{figure*}[!htbp]
\centering

        \begin{subfigure}[b]{0.33\textwidth}
        \centering
        \includegraphics[height=1.0\textwidth, keepaspectratio]{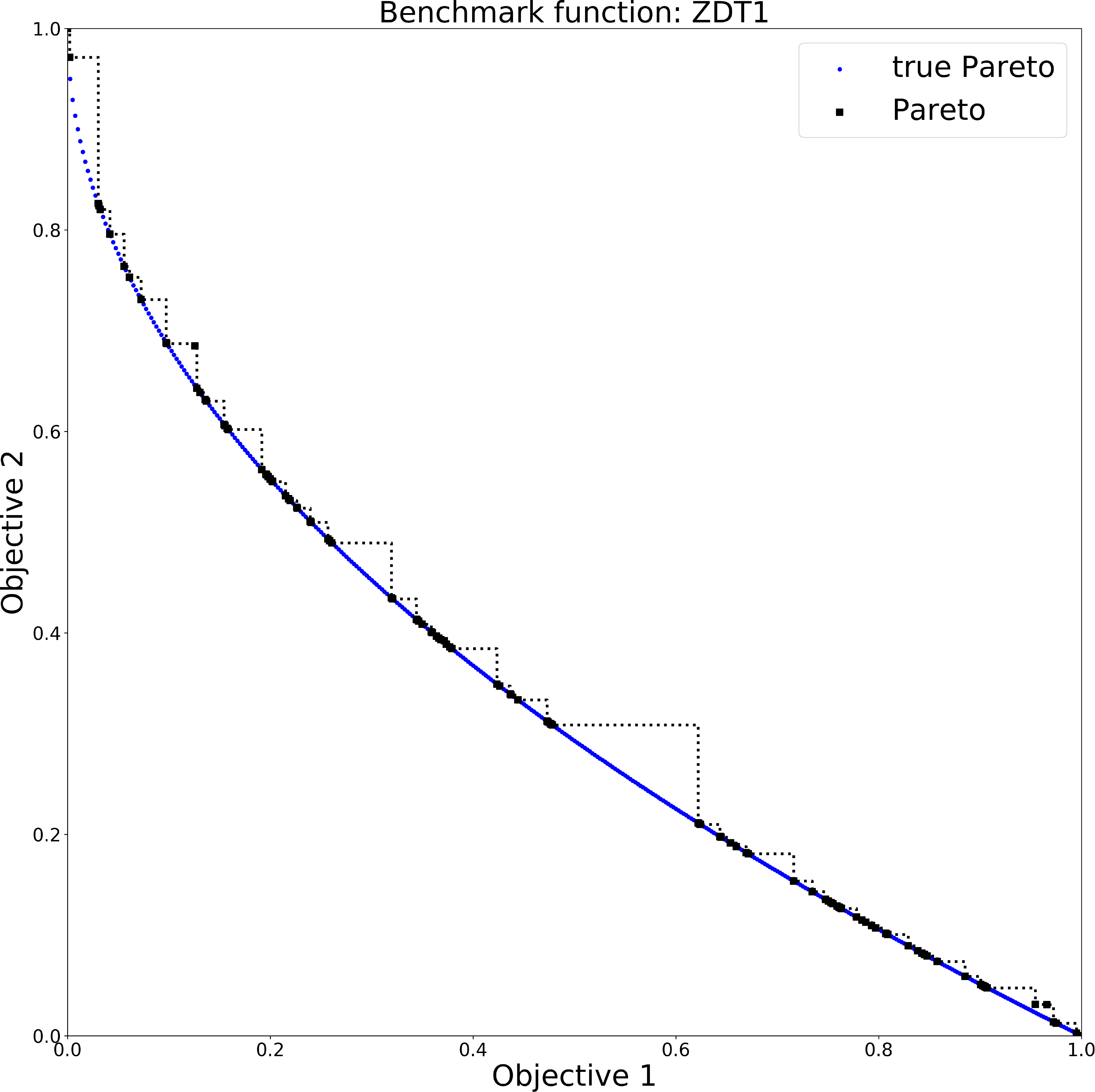}
        \caption{ZDT1}
        \end{subfigure}
        \hfill
        \begin{subfigure}[b]{0.33\textwidth}
        \centering
        \includegraphics[height=1.0\textwidth, keepaspectratio]{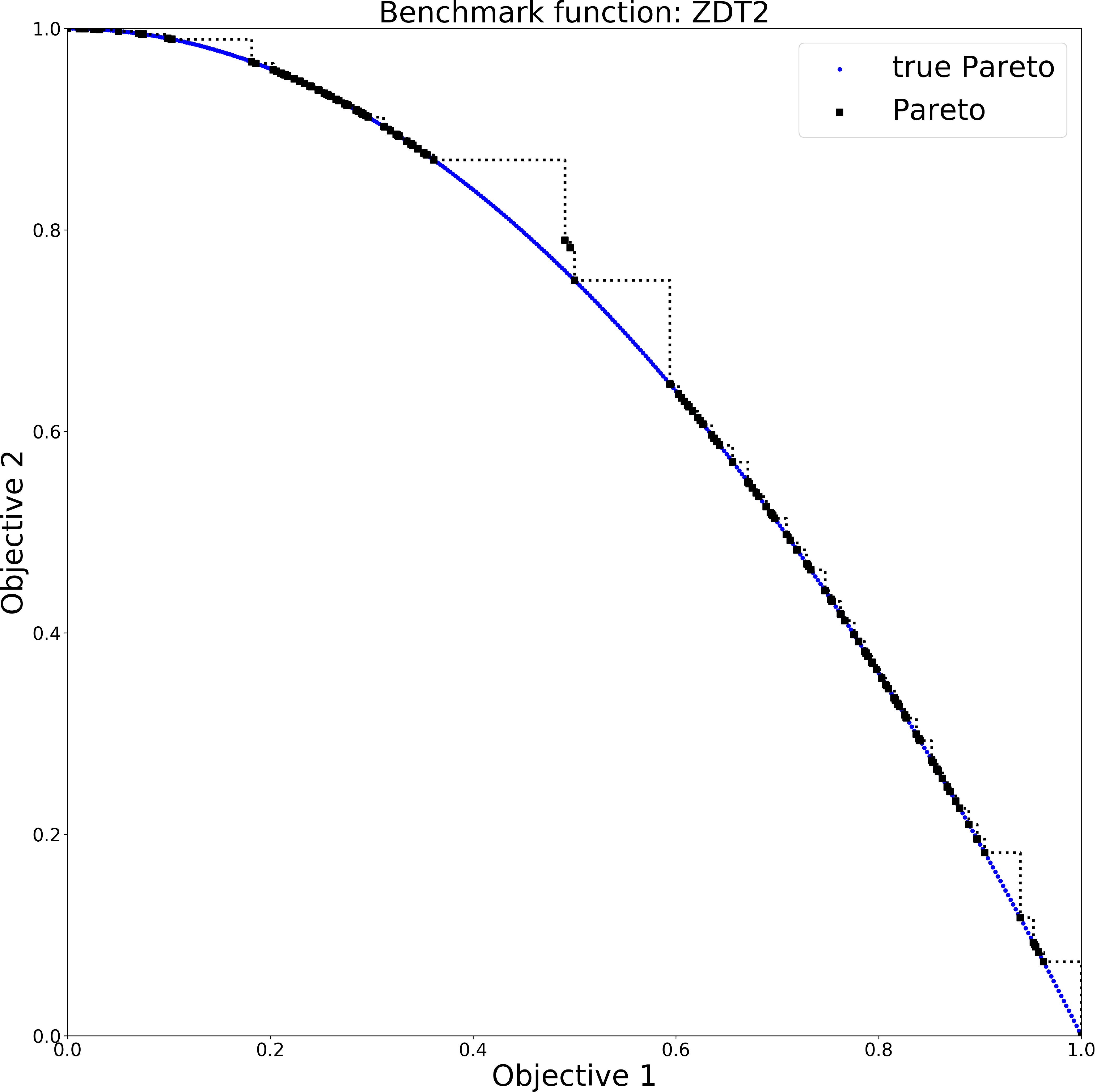}
        \caption{ZDT2}
        \end{subfigure}
        \hfill
        \begin{subfigure}[b]{0.33\textwidth}
        \centering
        \includegraphics[height=1.0\textwidth, keepaspectratio]{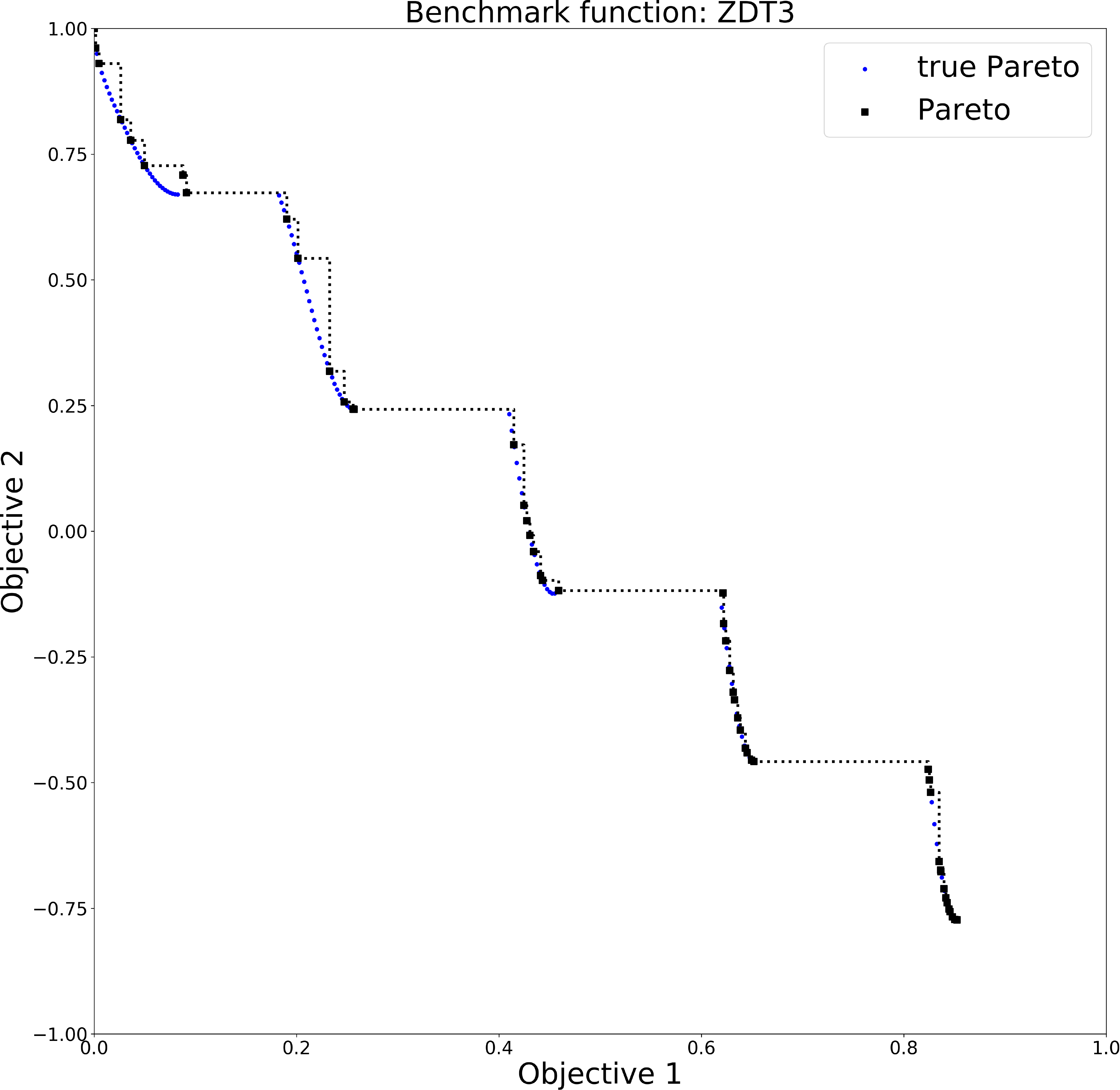}
        \caption{ZDT3}
        \end{subfigure}

        \vfill

        \begin{subfigure}[b]{0.33\textwidth}
        \centering
        \includegraphics[height=1.0\textwidth, keepaspectratio]{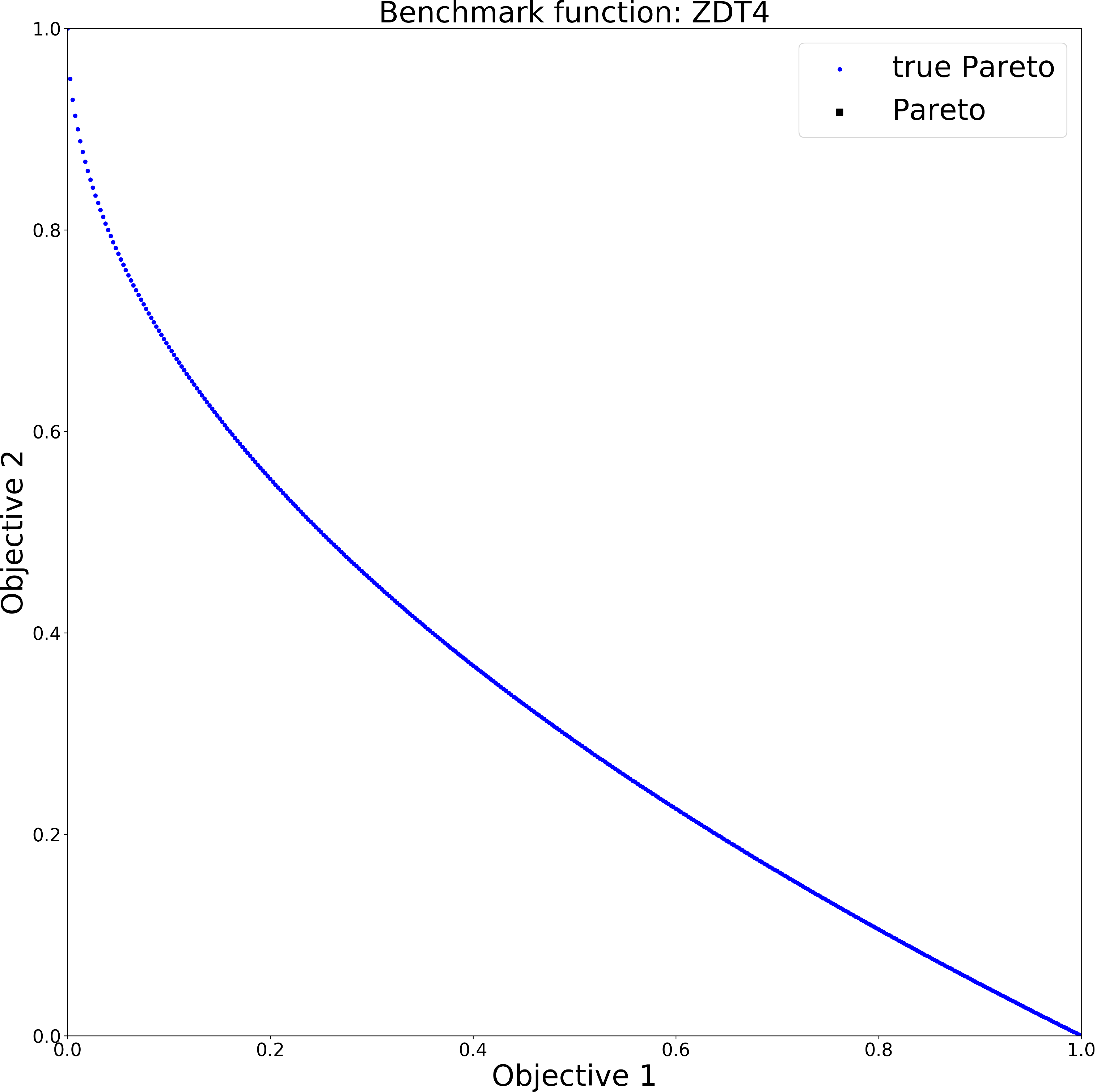}
        \caption{ZDT4}
        \end{subfigure}
        \hfill
        \begin{subfigure}[b]{0.33\textwidth}
        \centering
        \includegraphics[height=1.0\textwidth, keepaspectratio]{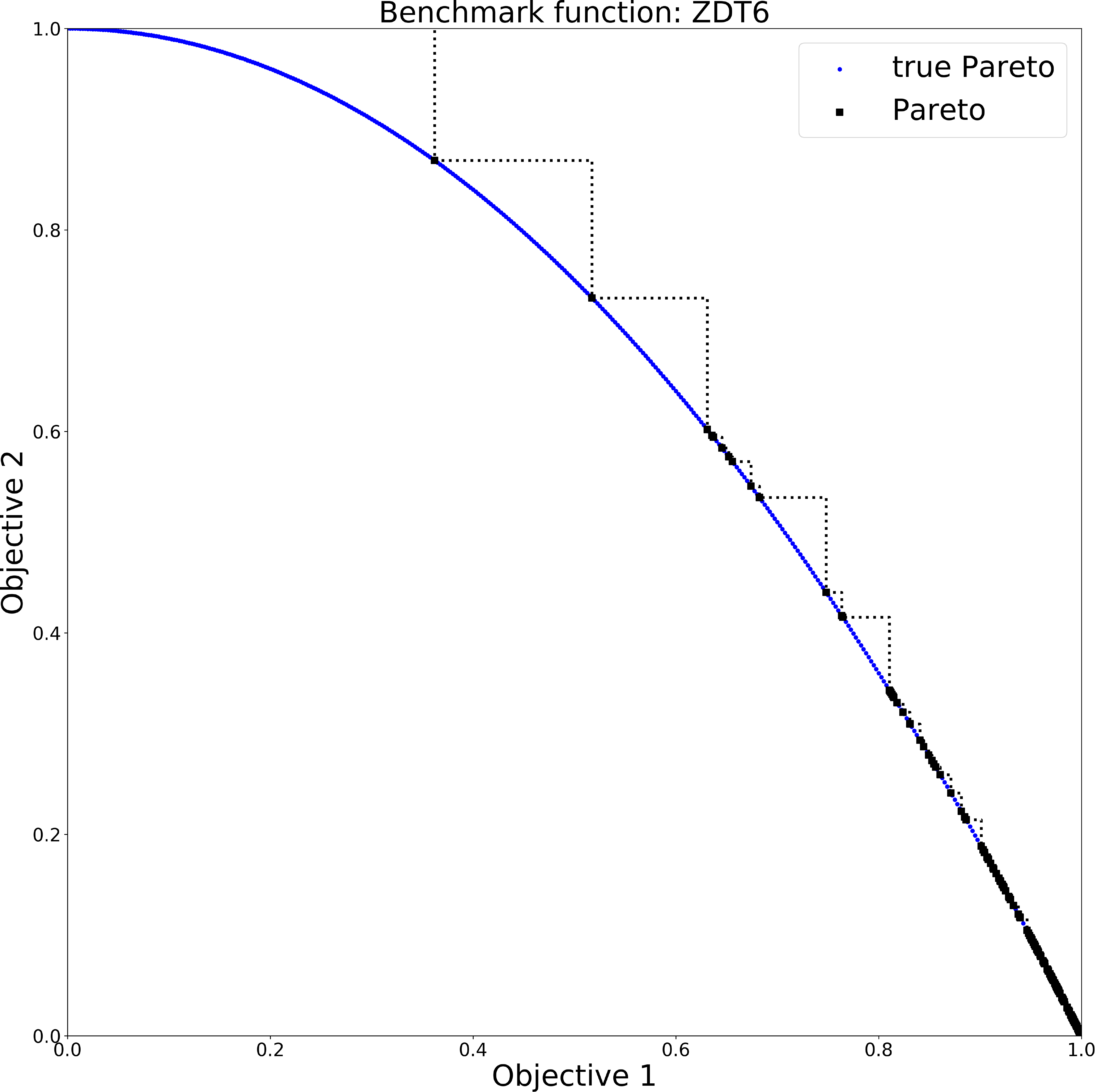}
        \caption{ZDT6}
        \end{subfigure}

        \begin{subfigure}[b]{0.33\textwidth}
        \centering
        \includegraphics[height=1.0\textwidth, keepaspectratio]{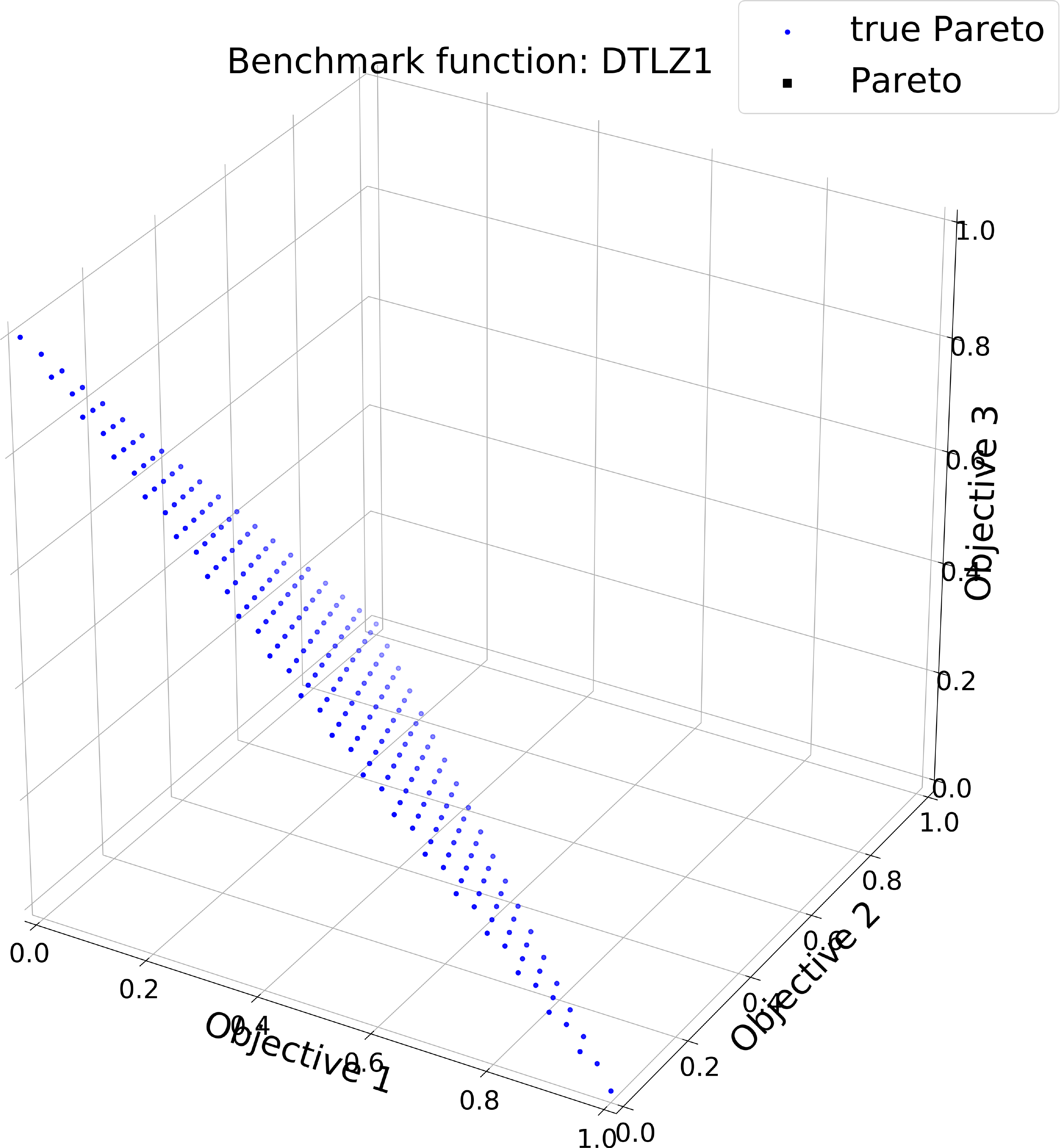}
        \caption{DTLZ1}
        \end{subfigure}
        \hfill
        \begin{subfigure}[b]{0.33\textwidth}
        \centering
        \includegraphics[height=1.0\textwidth, keepaspectratio]{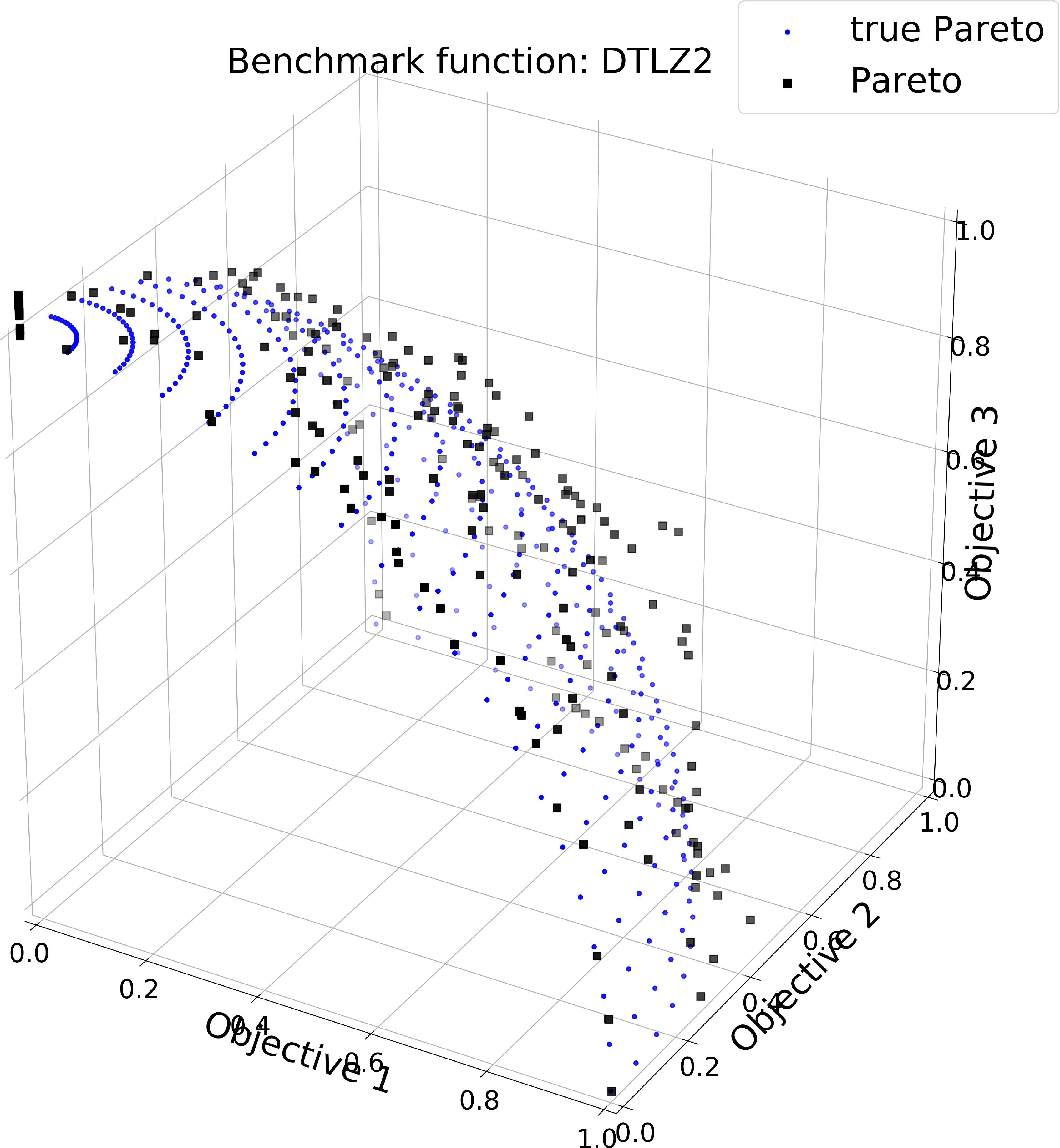}
        \caption{DTLZ2}
        \end{subfigure}
        \hfill
        \begin{subfigure}[b]{0.33\textwidth}
        \centering
        \includegraphics[height=1.0\textwidth, keepaspectratio]{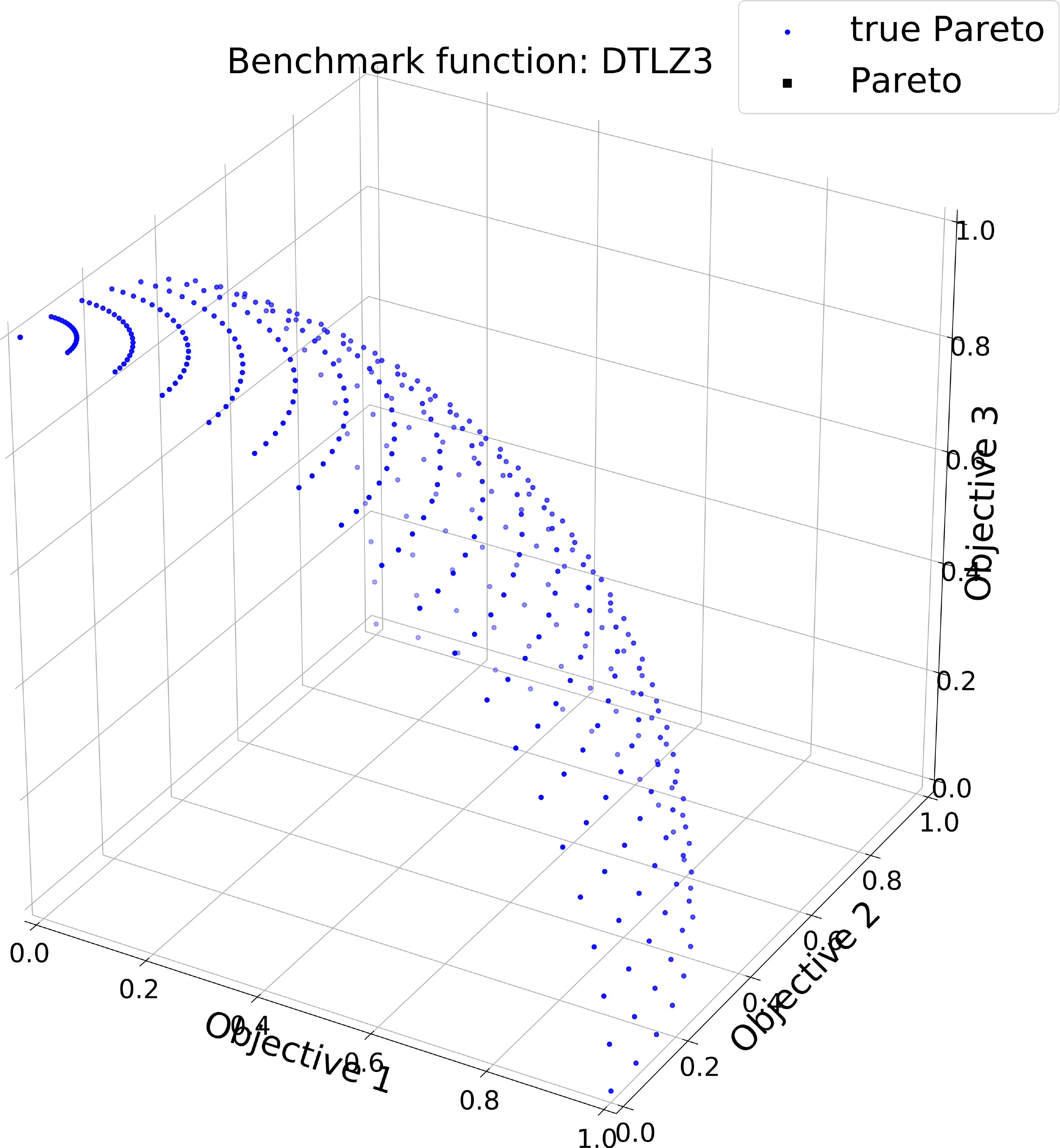}
        \caption{DTLZ3}
        \end{subfigure}

        \vfill

        \begin{subfigure}[b]{0.33\textwidth}
        \centering
        \includegraphics[height=1.0\textwidth, keepaspectratio]{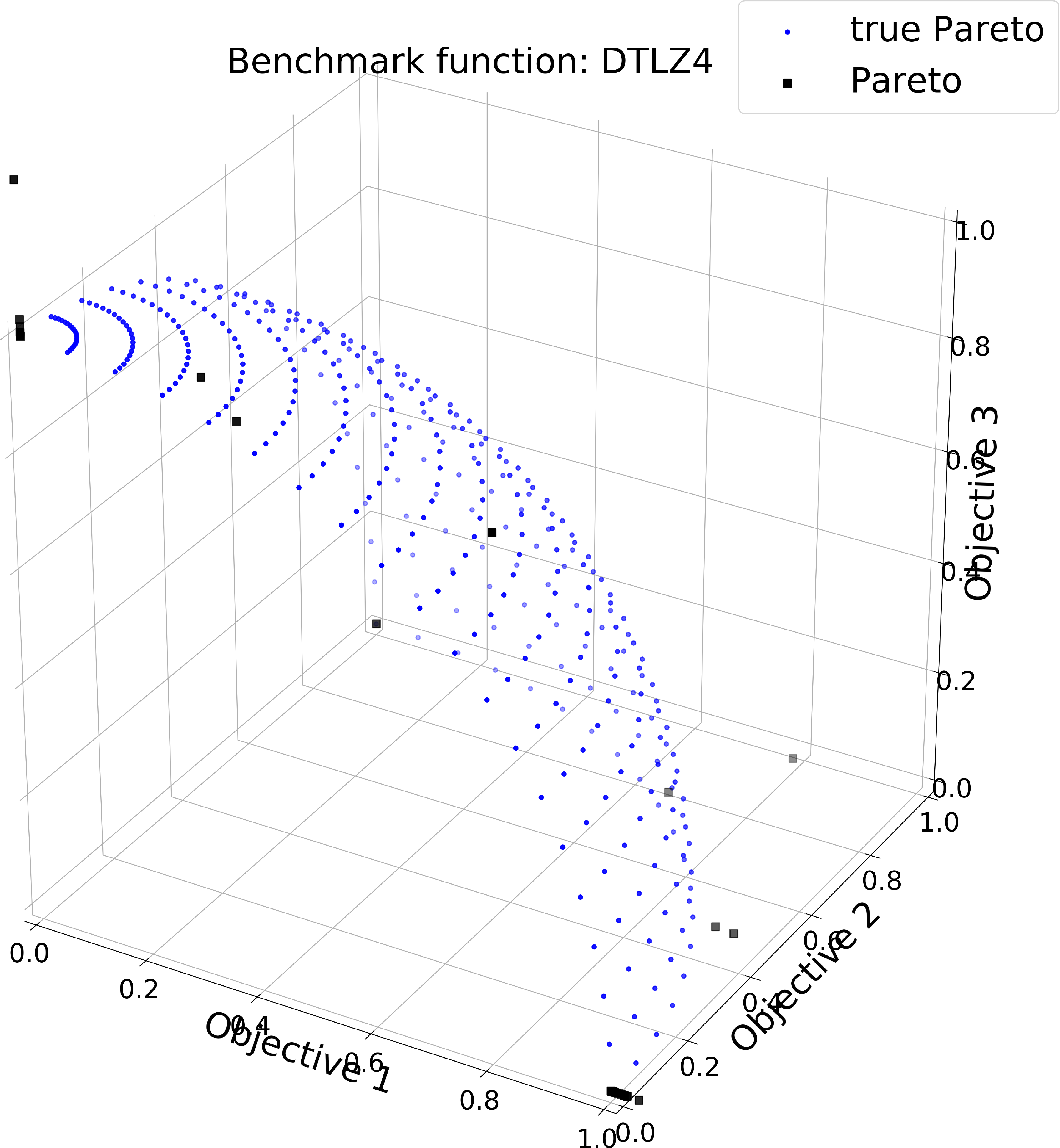}
        \caption{DTLZ4}
        \end{subfigure}
        \hfill
        \begin{subfigure}[b]{0.33\textwidth}
        \centering
        \includegraphics[height=1.0\textwidth, keepaspectratio]{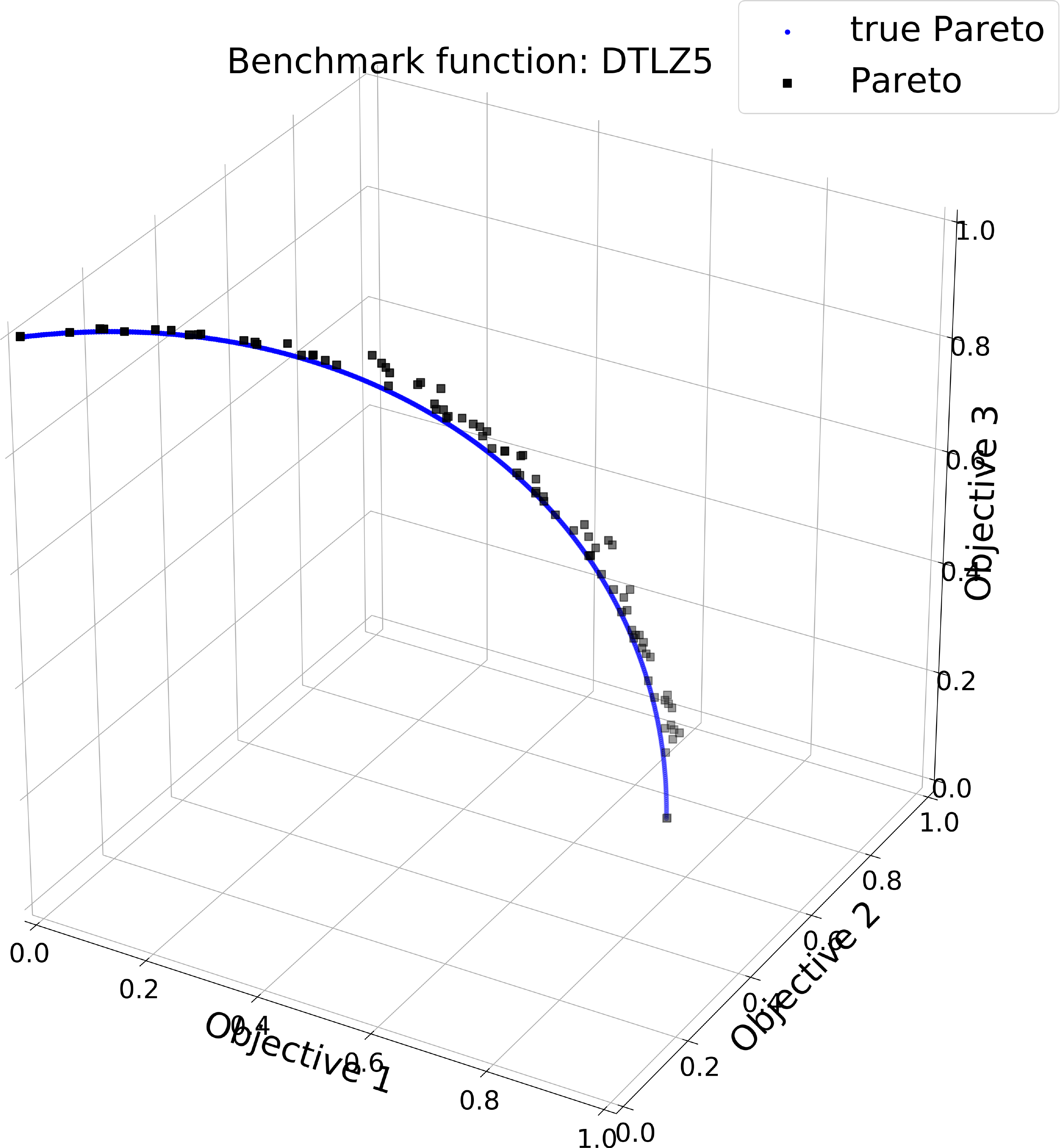}
        \caption{DTLZ5}
        \end{subfigure}
        \hfill
        \begin{subfigure}[b]{0.33\textwidth}
        \centering
        \includegraphics[height=1.0\textwidth, keepaspectratio]{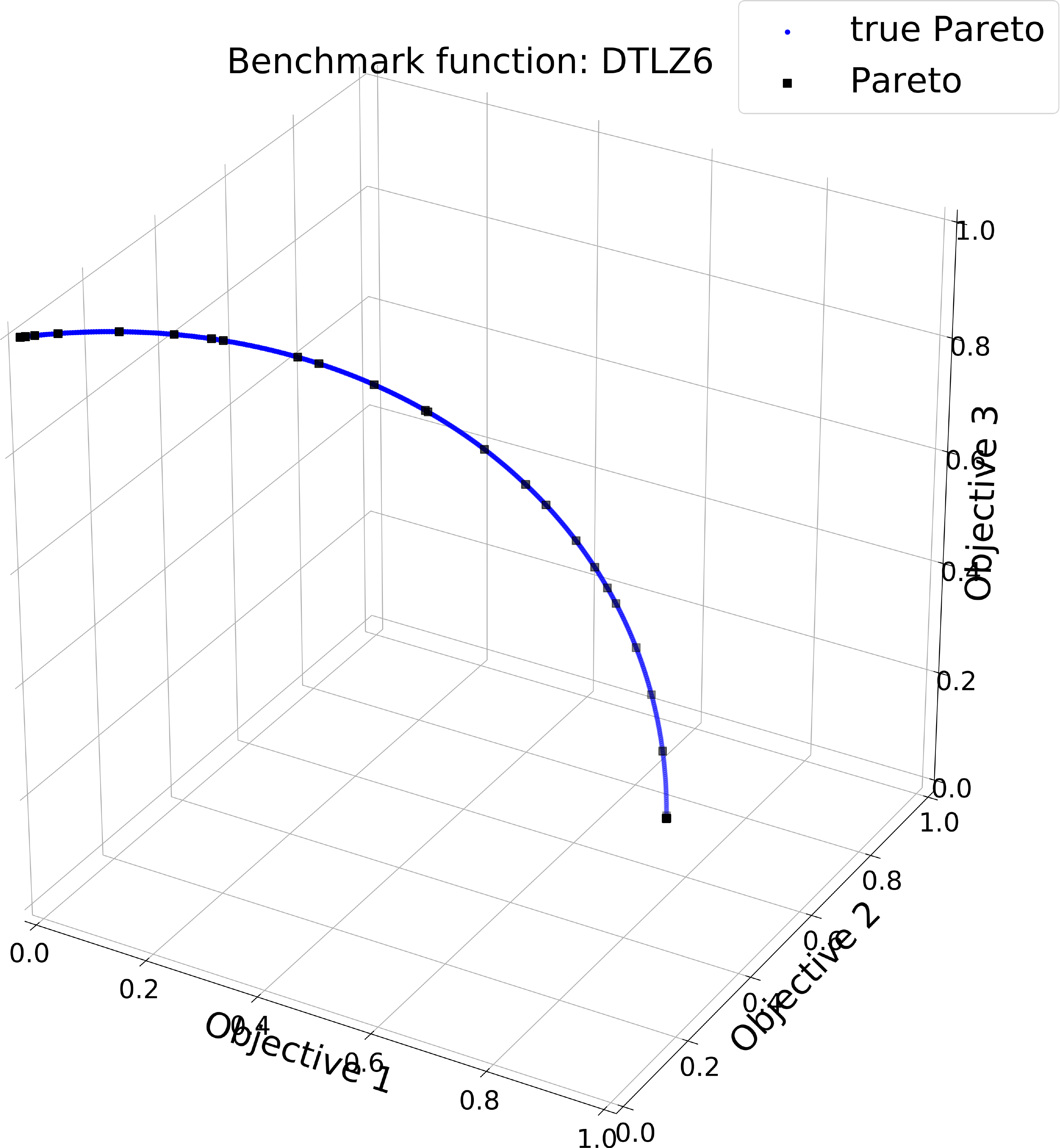}
        \caption{DTLZ6}
        \end{subfigure}

        \caption{Pareto frontier comparison for ZDT/DTLZ test suites.} 
        \label{fig:ParetoVisualization}
\end{figure*}

Figure \ref{fig:dtlzViolin} shows the comparison between 18 variants of the srMO-BO-3GP methods in DTLZ test suite, including regularized vs. non-regularized augmented Tchebycheff scalarized single-objective functions. 
Again, regularization seems to slightly improve the performance in terms of hypervolume. 
For the Pareto acquisition function, all UCB, EI, and PI acquisition functions seem to perform on par with each other. 
For the objective acquisition function, UCB outperforms both EI and PI, and EI and PI are of the same performance. 
\black{Figure \ref{fig:ParetoVisualization} compares Pareto frontiers obtained by the proposed algorithm against the true Pareto frontiers, showing an acceptable agreement in ZDT and DTLZ benchmarks. Overall, the proposed algorithm performs relatively well on most benchmarks, except for those highly oscillatory with multiple modes, but the optimum' objectives are reasonably close with the global optimum' objectives.}

\section{Engineering applications: Thermomechanical finite element for flip-chip package design optimization}
\label{sec:Applications}

\begin{figure}[!t]
\centering
\includegraphics[width=0.50\textwidth, keepaspectratio]{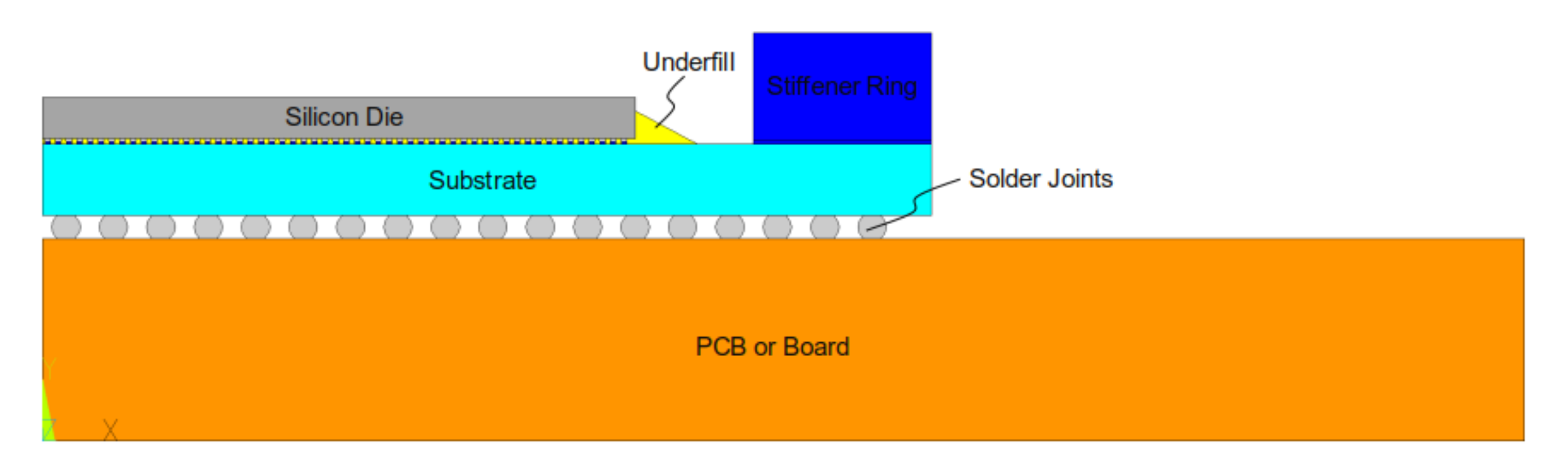}
\caption{Finite element model geometry.}
\label{fig:feModelGeom}
\end{figure}

We demonstrate the applicability of our proposed framework to a thermomechanical FEM model for flip-chip package design, where five objectives are considered. 
Figure~\ref{fig:feModelGeom} shows the geometric model of the thermomechanical problem, where the mesh density varies for different levels of fidelity. 
Two design variables are associated with the die, three are associated with the substrate, three more are associated with the stiffener ring, two are with the underfill, and the last one is with the PCB board. 
Only two levels of fidelity are considered in this example. 
Table~\ref{tab:ansysFCBGAdesignVar} show the design variables, the physical meaning of the design variables, as well as their lower and upper bounds in this case study. 

\begin{table}[!htbp]
\centering
\scriptsize
\caption{Design variables for the FCBGA design optimization.}
\label{tab:ansysFCBGAdesignVar}
\begin{tabular}{|l|l|l|l|l|} \hline
\textbf{Variable}   & \textbf{Design part}            & \textbf{Lower bound}         & \textbf{Upper bound}            &  \textbf{Optimal value}           \\ \hline
$x_1$      & die                    & 20000               & 30000                  &  20702                   \\ 
$x_2$      & die                    & 300                 & 750                    &  320                     \\ 
$x_3$      & substrate              & 30000               & 40000                  &  35539                   \\ 
$x_4$      & substrate              & 100                 & 1800                   &  1614                    \\ 
$x_5$      & substrate              & $10\cdot 10^{-6}$   & $17\cdot 10^{-6}$      &  $17\cdot 10^{-6}$       \\ 
$x_6$      & stiffener ring         & 2000                & 6000                   &  4126                    \\ 
$x_7$      & stiffener ring         & 100                 & 2500                   &  1646                    \\ 
$x_8$      & stiffener ring         & $8\cdot 10^{-6}$    & $25 \cdot 10^{-6}$     &  $8.94\cdot 10^{-6}$     \\ 
$x_9$      & underfill              & 1.0                 & 3.0                    &  1.52                    \\ 
$x_{10}$   & underfill              & 0.5                 & 1.0                    &  0.804                   \\ 
$x_{11}$   & PCB board              & $12.0\cdot 10^{-6}$ & $16.7\cdot 10^{-6}$    &  $16.7\cdot 10^{-6}$     \\ \hline
\end{tabular}
\normalsize
\end{table}

After the numerical solution is obtained, the component warpage at 20$^{\circ}$C, 200$^{\circ}$C, and the strain energy density of the furthest solder joint are calculated. 
The five objectives are listed as follows,
\begin{enumerate}[{Objective}-1:]
\item warpage at $20^{\circ}$C,
\item warpage at $200^{\circ}$C,
\item damage metric for BGA lifetime prediction,
\item damage metric for C4 interconnection lifetime prediction,
\item first principal stress at C4 corner, where cracking and delamination frequently occurs,
\end{enumerate}
where the goal is to minimize all these objectives.

\begin{figure}[!t]
\centering
        \begin{subfigure}[b]{0.475\textwidth}
        \centering
        \includegraphics[height=175px, keepaspectratio]{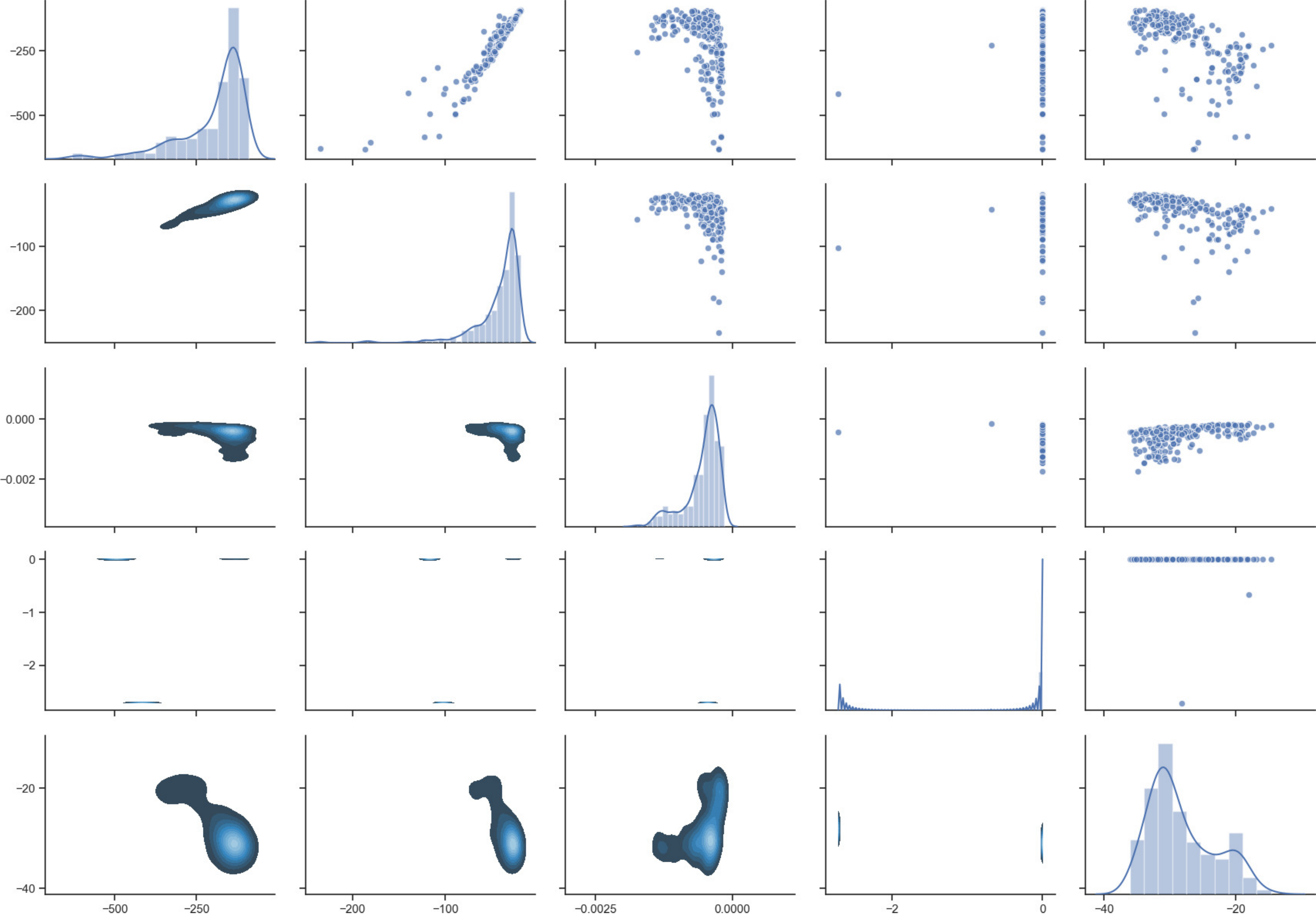}
        \caption{Correlation between objectives and their joint densities.}
        \label{fig:objCorr}
        \end{subfigure}
        \hfill
        \begin{subfigure}[b]{0.475\textwidth}
        \centering
        \includegraphics[height=175px, keepaspectratio]{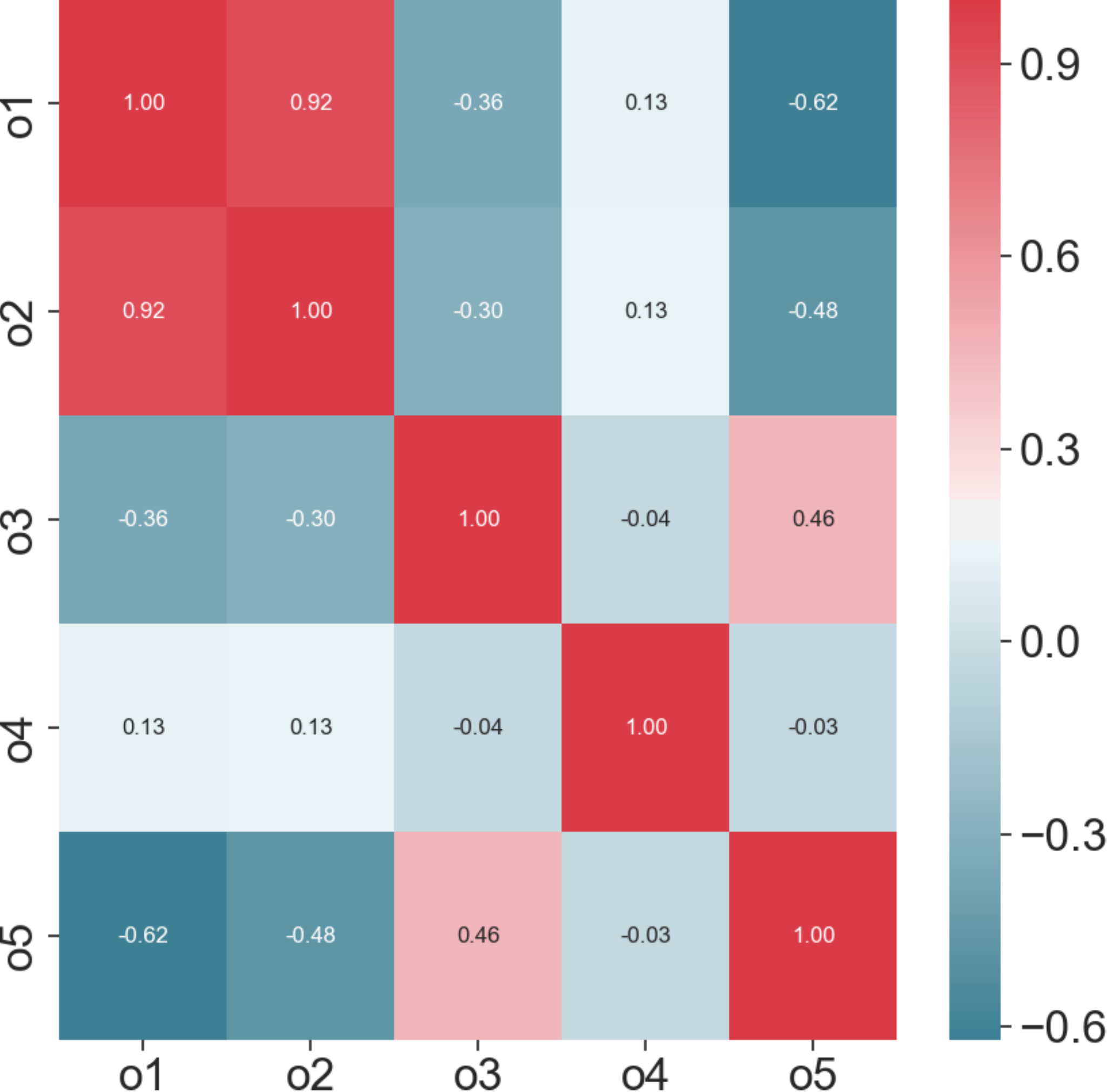}
        \caption{Correlation heatmap between objectives.}
        \label{fig:objCorrHeatmap}
        \end{subfigure}
        \caption{Correlation between five objectives in the FEM example and their joint densities.}
\end{figure}

Figure~\ref{fig:objCorr} shows the scatter plot matrix between the five objectives and their joint densities, whereas Figure~\ref{fig:objCorrHeatmap} presents the Pearson correlation, which is bounded between (-1) and (+1). 
851 simulations are performed, in which only 256 simulations are feasible. 
Out of 256 feasible simulations, 84 of those represent the current Pareto frontier of this numerical study. 
The objective 1 is very positively correlated with the objective 2, as both of them corresponds to the warpages at different temperatures. 
This implies that minimizing the object 1 would also minimize the objective 2, thus no trade-off is found. 
The same argument applies for objectives 3 and 5, where positive correlation is found. 
The objective 4 is poorly correlated with other objectives. 
Trade-offs are found between members of the group of objectives 1 and 2 and those of the group of objective 3 and 5. 
Overall, it is challenging to plot the Pareto front on high-dimensional space for a practical visualization. 
However, we have demonstrated that our methodology is applicable to problems with many objective functions. 
In practice, the number of objectives are often reduced to minimum, before the optimization study is conducted.

\section{Discussion}
\label{sec:Discussion}


BO is a powerful and flexible framework which allows for many useful extensions. 
For example, the local GP approach~\cite{tran2018efficient,bostanabad2019globally} could be used to improve the scalability of GP. 
An extended version of local GP has also been developed~\cite{tran2019constrained} to solve the mixed-integer optimization problems, where the number of discrete/categorical variables is relatively small. 
Accelerated BO methods by exploiting computational resource on high-performance computing platform have been proposed~\cite{tran2019pbo,tran2020aphbo} to reduce the amount of physical waiting time. 
Multi-fidelity BO approaches~\cite{tran2019sbfbo2cogp,tran2020smfbo2cogp} have been developed to couple information between different levels of fidelity by exploiting the correlation between low- and high-fidelity models to reduce the computational efforts. 
Its success has been demonstrated, at least in the field of bioengineering~\cite{travaglino2020computational} and computational fluid dynamics~\cite{tran2018weargp,tran2020weargp}.

For multi-objective problems, hypervolume-based approaches remain more popular than the scalarization approaches. Furthermore, there are limitations in combining multiple objectives into single objective \black{as this approach relies on the weights, which are randomly sampled and may lead to convergence and diversity issues}. 
Perhaps, combining the Pareto classifier with a hypervolume-based approach would result in a better numerical \black{performance}. \black{This topic remains as an interesting direction in the future.}

An advantage of the proposed method is that the inclusion of the uncertain Pareto classifier makes the effort of locating the Pareto frontier easier in the input space $\mathcal{X}$, regardless of whether the Pareto frontier is discontinuous, convex, or concave, because this uncertain Pareto classifier only considers the input space. 
As shown in this paper, the proposed method performs relatively well for multiple objectives in both ZDT and DTLZ test suites. 

\black{As usual, the predictive performance of GP is highly dependent on the dimensionality of the problem, i.e. the number of design variables, and the characteristics of the underlying function, which is typically assumed to be smooth. It is worthy to note that the proposed algorithm does not depend on the number of objective functions, as the Pareto classifier is constructed directly on the input space $\mathcal{X}$ and the weighted Tchebycheff scheme combines multiple objectives to a single objective. Therefore, the proposed srMO-BO-3GP method has a competitive advantage for problems with high number of objective functions. Furthermore, it does not depend on the characteristics of the Pareto frontiers, i.e. whether they are convex, concave, or discontinuous. The known constraints do not have any effect on the optimization problem, as we assume they can be cheaply evaluated and the functional evaluation $f(\cdot)$ would not be invoked if one of the known constraints $g_k$ is violated. The number of unknown constraints also does not have any effect on the optimization problems, as we are primarily concerned with the input space $\mathcal{X}$, and the probability of satisfying all unknown constraints is modeled by another GP, which can be considered as a black-box function. In summary, the main factor that may have the most impact on the numerical performance of srMO-BO-3GP is the dimensionality of the optimization problem.}

\black{The drawback of Tchebycheff decomposition for MO problems is that it can only identify a unique Pareto solution point for a fixed weight vector at best (cf. Miettinen~\cite{miettinen2012nonlinear}, Section 2.1). The diversity of the Pareto frontier solely depends on the weight vectors, which is uniformly distributed in this case. Furthermore, because the Tchebycheff scheme does not consider the current Pareto frontiers, it does not promote diversity the Pareto frontiers, even though one might argue that the convergence is preserved. The second term in the composite acquisition function (Equation \ref{eq:FullAcquisition}), which corresponds to the Pareto classifier, is inserted to promote the diversity in Pareto frontiers and bypass the weight adjustment problem.}

\black{As theoretically formulated in the BO framework, the acquisition function strikes for the balance of exploration and exploitation, which would probabilistically converge to the global optimum, regardless of the number and locations of initial sampling points, as described in Bull ~\cite{bull2011convergence}. Strong theoretical guarantees are often provided in the case of single objective function as BO is well known as a no-regret optimization algorithm. Along with the Tchebycheff argument in Miettinen~\cite{miettinen2012nonlinear}, where optimizing with a fixed weight vector would lead to a unique Pareto point, the proposed srMO-BO-3GP method does not depend on the initial sampling points, at least from a theoretical point of view.}

The proposed method does not cope well with high-dimensional problems, as they are the usual challenges for GP. 
While some solutions have been proposed, the scalability problem remains as an active research area and is posed as the future extension to this work.

\section{Conclusion}
\label{sec:Conclusion}

In this paper, we propose a multi-objective Bayesian optimization framework, called srMO-BO-3GP in a sequential setting with applications to engineering-based simulations. 
In this framework, three distinct GPs are coupled together. 
The first GP is used to approximate a single-objective function, which is converted from the original multi-objective functions using a regularization-augmented Tchebycheff function. 
The second GP is used to learn the unknown constraints, which are evaluated simultaneously with the set of objective functions, supporting the general case where some performance metrics reflect design aspirations (objectives) while others must satisfy prescribed requirements (constraints). 
The third GP is used as an uncertain binary classifier to learn the Pareto frontier, where its own acquisition function is embedded in the main acquisition function. 
The srMO-BO-3GP framework is demonstrated using two numerical benchmarking test suites, ZDT and DTLZ, and an engineering thermomechanical FEM application.

\section*{Acknowledgment}
The views expressed in the article do not necessarily represent the views of the U.S. Department of Energy or the United States Government. Sandia National Laboratories is a multimission laboratory managed and operated by National Technology and Engineering Solutions of Sandia, LLC., a wholly owned subsidiary of Honeywell International, Inc., for the U.S. Department of Energy's National Nuclear Security Administration under contract DE-NA-0003525. 

%
%

\bibliographystyle{asmems4}
\bibliography{lib}   

\end{document}